\documentclass[lettersize,journal]{IEEEtran}
\usepackage{amsmath,amsfonts}
\usepackage{amssymb}
\usepackage{mathrsfs}
\usepackage{booktabs}
\usepackage{multirow}
\usepackage{xcolor}
\usepackage{subfigure}
\usepackage{bm}
\usepackage{dsfont}
\usepackage{algorithmic}
\usepackage{algorithm}
\usepackage{array}
\usepackage{textcomp}
\usepackage{stfloats}
\usepackage{url}
\usepackage{verbatim}
\usepackage{graphicx}
\usepackage{cite}
\usepackage{makecell}
\begin{document}

\title{Distributional Drift Adaptation with Temporal Conditional Variational Autoencoder for Multivariate Time Series Forecasting}

\author{Hui~He,~\IEEEmembership{}
        Qi~Zhang,~\IEEEmembership{}
        Kun~Yi,~\IEEEmembership{}
        Kaize~Shi,~\IEEEmembership{}
        Zhendong~Niu~\IEEEmembership{}
        and~Longbing~Cao,~\IEEEmembership{Senior Member, IEEE}
\thanks{Hui He is with the School of Medical Technology, Beijing Institute of Technology, Beijing 100081, China (e-mail: hehui617@bit.edu.cn).}
\thanks{Qi Zhang is with the School of Computer Science and Technology, Beijing Institute of Technology, Beijing 100081, China, and the Department of Computer Science, Tongji University, Shanghai 201804, China (e-mail: zhangqi\_cs@tongji.edu.cn).}
\thanks{Kun Yi is with the School of Computer Science and Technology, Beijing Institute of Technology, Beijing 100081, China (e-mail: yikun@bit.edu.cn).}
\thanks{Kaize Shi is with the Data Science and Machine Intelligence Lab, University of Technology Sydney, NSW 2007, Australia (e-mail: Kaize.Shi@uts.edu.au).}
\thanks{Zhendong Niu is with the School of Computer Science and Technology, Beijing Institute of Technology, Beijing 100081, China, and the Engineering Research Center of Integration and Application of Digital Learning Technology, Ministry of Education, China (e-mail: zniu@bit.edu.cn).}
\thanks{Longbing Cao is with the DataX Research Centre, School of Computing, Macquarie University, Sydney, NSW 2109, Australia (e-mail: longbing.cao@mq.edu.au).}
\thanks{Manuscript received April 19, 2021; revised August 16, 2021. \\
(Corresponding authors: Qi Zhang and Zhendong Niu.)}}

\markboth{Journal of \LaTeX\ Class Files,~Vol.~14, No.~8, August~2021}%
{Shell \MakeLowercase{\textit{et al.}}: A Sample Article Using IEEEtran.cls for IEEE Journals}

\IEEEpubid{0000--0000/00\$00.00~\copyright~2021 IEEE}

\maketitle

\begin{abstract}
Due to the non-stationary nature, the distribution of real-world multivariate time series (MTS) changes over time, which is known as \textit{distribution drift}. Most existing MTS forecasting models greatly suffer from distribution drift and degrade the forecasting performance over time. Existing methods address distribution drift via adapting to the latest arrived data or self-correcting per the meta knowledge derived from future data. Despite their great success in MTS forecasting, these methods hardly capture the intrinsic distribution changes, especially from a distributional perspective. Accordingly, we propose a novel framework \textit{temporal conditional variational autoencoder} (TCVAE) to model the dynamic distributional dependencies over time between historical observations and future data in MTSs and infer the dependencies as a temporal conditional distribution to leverage latent variables. Specifically, a novel temporal Hawkes attention mechanism represents temporal factors subsequently fed into feed-forward networks to estimate the prior Gaussian distribution of latent variables. The representation of temporal factors further dynamically adjusts the structures of Transformer-based encoder and decoder to distribution changes by leveraging a gated attention mechanism. Moreover, we introduce conditional continuous normalization flow to transform the prior Gaussian to a complex and form-free distribution to facilitate flexible inference of the temporal conditional distribution. Extensive experiments conducted on six real-world MTS datasets demonstrate the TCVAE's superior robustness and effectiveness over the state-of-the-art MTS forecasting baselines. We further illustrate the TCVAE applicability through multifaceted case studies and visualization in real-world scenarios.
\end{abstract}


\section{Introduction}
\IEEEPARstart{M}{ultivariate} time series (MTS) is a continuous and intermittent unfolding of time-stamped variables over time. MTS forecasting predicts the future values of time-stamped variables based on their extremely substantial historical observations. MTS forecasting is widely applicable for various applications, such as traffic management~\cite{DBLP:conf/nips/0001YL0020, DBLP:conf/icde/CirsteaYGKP22}, financial trading~\cite{DBLP:conf/kdd/ZhangAQ17, DBLP:journals/eswa/LeeK20, DBLP:journals/tnn/TranIKG19}, energy optimization~\cite{DBLP:conf/aaai/XuCZSNYLC020}, user modeling~\cite{DBLP:journals/air/TarusNM18,DBLP:journals/tnn/ZhangCSN22} and epidemic propagation studies~\cite{DBLP:journals/pami/SpadonHBMRS22, DBLP:conf/nips/CaoWDZZHTXBTZ20}. A recent typical example is to predict the case trends of COVID-19 infections, which can provide hints or evidence for intervention policies and prompt actions to contain the virus spread or resurgence and large-scale irreparable socioeconomic impact~\cite{Cao-covid21}.

\IEEEpubidadjcol

\begin{figure}[!t]
\centering
\subfigure[Traffic flows of sensor $\#$3, $\#$17 and $\#$33]{
    \label{figure 1a}
    \includegraphics[width=1.0\linewidth]{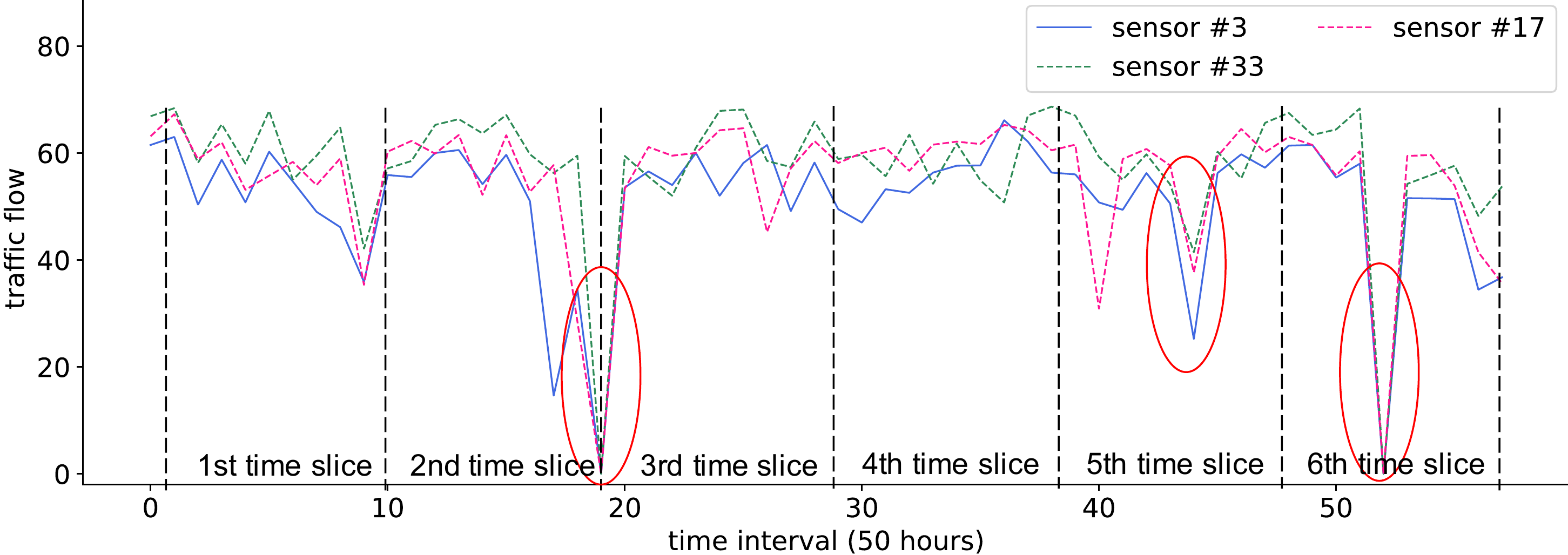}
    }
\subfigure[Distribution of six successive time slices]{
    \label{figure 1b}
    \includegraphics[width=0.466\linewidth]{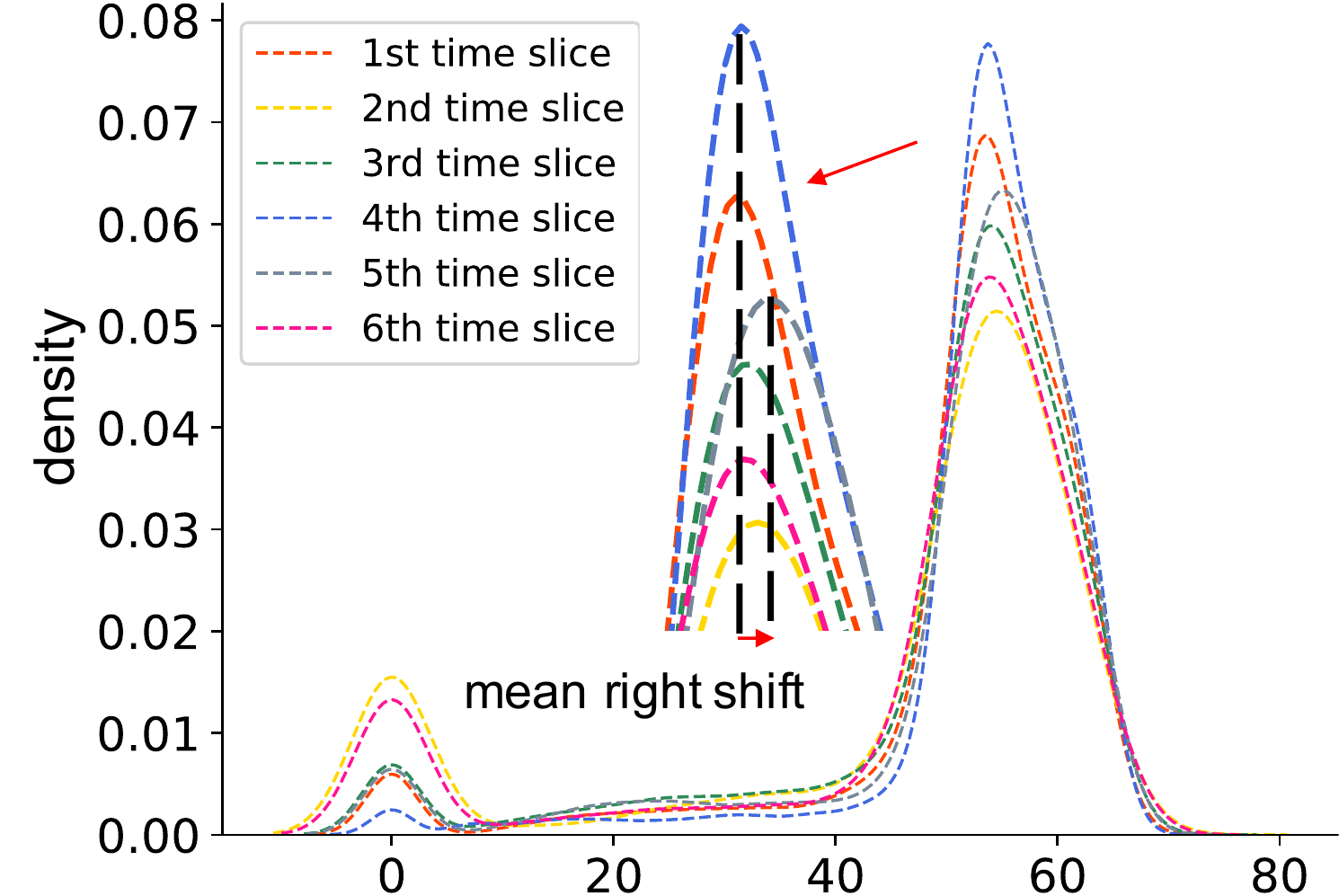}
    }
\subfigure[Distribution of six cumulative time slices]{
    \label{figure 1c}
    \includegraphics[width=0.466\linewidth]{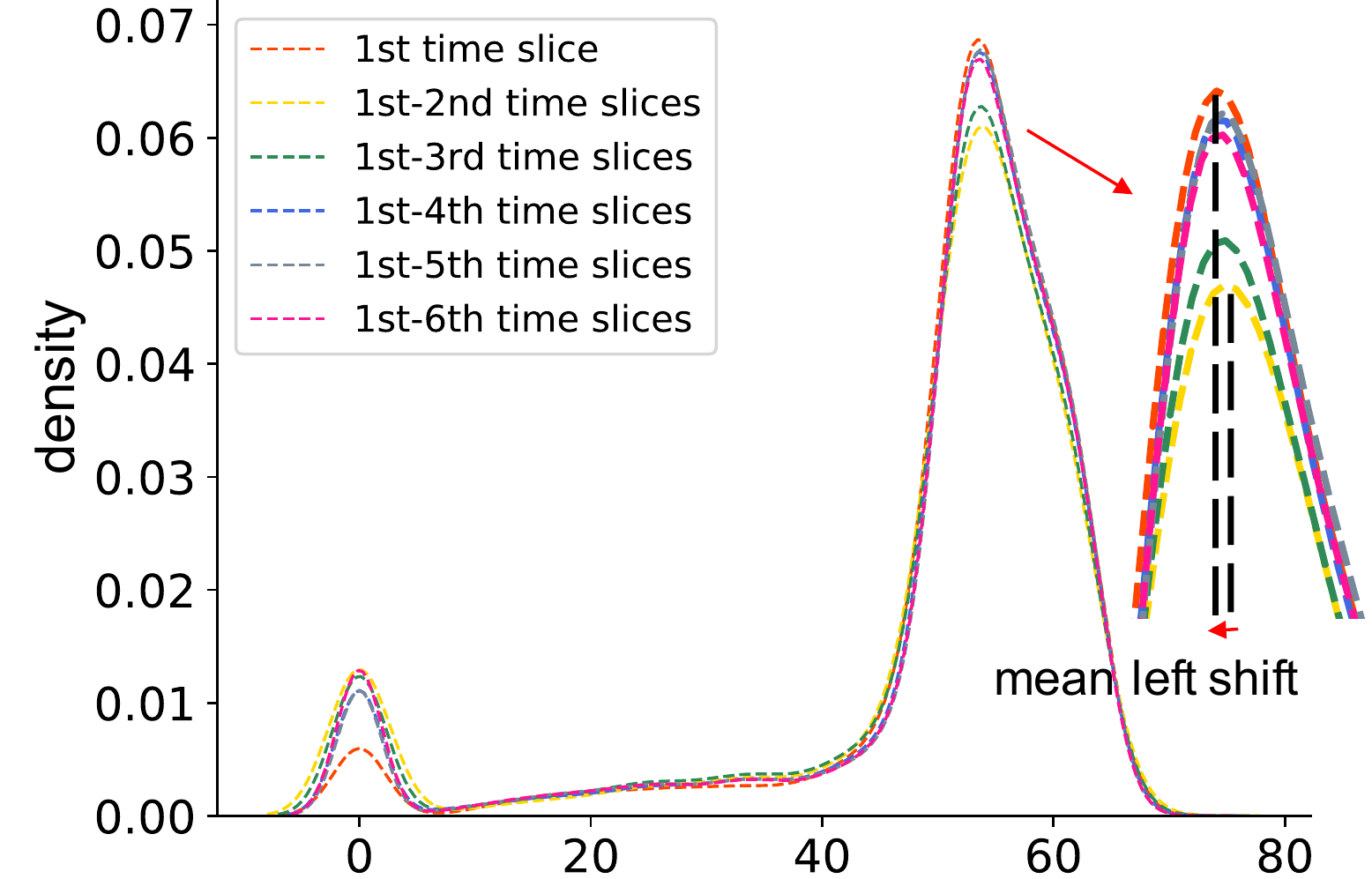}
    }
\caption{An example of MTS traffic flow derived from METR-LA.}
\label{figure 1}
\end{figure}

Traditional models such as vector autoregression (VAR)~\cite{Vector1993} and probabilistic models~\cite{SeegerSF16,9380704} show appealing interpretability and theoretical guarantees for MTS forecasting. Recent advanced models such as LSTM~\cite{DBLP:conf/sigir/LaiCYL18,DBLP:journals/ml/ShihSL19,DBLP:journals/tnn/BandaraBH21,9669023,10197239}, TCN~\cite{DBLP:conf/nips/SenYD19} and Transformer~\cite{DBLP:conf/aaai/ZhouZPZLXZ21,DBLP:conf/iclr/LiuYLLLLD22,DBLP:conf/nips/WuXWL21,DBLP:conf/icml/ZhouMWW0022,DBLP:conf/iclr/ZhangY23,10308867} are more competent in capturing temporal dependencies. These models, despite their successful applications in various domains, generally assume that the distribution of time series keeps stationary through time, requiring constant conditional dependencies between future predictions and historical observations. The above assumption oversimplifies the highly non-stationary and non-IID (or non-i.i.d., non-independent and identically distributed) nature of real-world MTSs \cite{cj_Cao14}, inapplicable to distribution-evolving MTSs and causing forecasting performance downgrade over time. Such a phenomenon forms \textit{distribution drift}~\cite{DBLP:conf/aaai/LiYLX022, 9783029}, where the statistical properties of MTSs change over time. As illustrated in Fig.~\ref{figure 1}, the traffic flows recorded by three sensors $\#$3, $\#$17 and $\#$33 from 9/1/2018 to 12/28/2018 are shown in the upper part, where the flows are split into six successive time slices. The flows sharply descend and ascend, forming the turning points between the 2$nd$ and 3$rd$ slices and within the 5$th$  and 6$th$ slices (marked in red ellipse). In addition, we take out the six slices of sensor $\#$3 and show its statistic evolution in terms of different time slices and cumulative slices in the lower left and right parts, respectively. We can see distribution drifts occur between any successive time series slices. For clarity, we only highlight the most intense shift, i.e., a mean right shift from the 4$th$ to the 5$th$ slice, as shown in the left part of Fig.~\ref{figure 1}, and a mean left shift from the 1$st$-2$nd$ time slices to the 1$st$-6$th$ time slices, as shown in the right part of Fig.~\ref{figure 1}. Obviously, the distribution of successive time series slices changes over time, significantly challenging even well-trained forecasting models. Therefore, addressing MTS distribution drift becomes a critical but challenging issue of MTS forecasting.

Few studies focus on capturing the distribution drift for MTS forecasting to date. Previous methods~\cite{DBLP:conf/aaai/ZhouLLYZJ19, DBLP:journals/corr/abs-2204-05101,DBLP:journals/corr/abs-2205-14415,DBLP:journals/tnn/PassalisTKGI20,9509335,DBLP:journals/corr/abs-2302-14829} address the distribution drift via adapting to the latest arrived data and applying the adapted models for future data prediction. However, such methods which hardly capture the intrinsic distribution drift suffer catastrophic forgetting~\cite{1611835114} and are unsuitable for scenarios with a sudden distributional change. Subsequently, recent adaptation methods~\cite{DBLP:conf/aaai/LiYLX022,DBLP:conf/aaai/YueWDYHTX22,DBLP:conf/cikm/Du0FPQXW21,DBLP:conf/cikm/YouZDFH21,DBLP:journals/tnn/SongLLZ23,10.1145/3580305.3599315} focus on future data and bridge the gaps between historical observations and future data by generating training samples/features following future data distribution or calculating their gradient difference. These methods learn to correlate the distribution drift with meta knowledge derived from future data and correct the bias caused by their distribution drift in a discriminative meta-learning framework. Despite their great success in MTS forecasting, these methods do not capture significant changes of distributional dependencies.

Different from previous adaptation-based methods neutralizing the bias caused by the distribution drift, we aim to model the dynamic distributional dependencies over time between historical observations ($\mathcal{X}$) and future data ($\mathcal{Y}$). Inspired by the superiority of generative models, e.g., variational autoencoder (VAE), in approximating data distribution in various fields~\cite{DBLP:conf/acl/ZhaoZE17, DBLP:conf/iclr/GuCHK19, DBLP:conf/ijcai/YangYDSZZ21}, we infer the distributional dependencies as a conditional distribution $p(\mathcal{Y}|\mathcal{X})$ by leveraging latent variables $\mathcal{Z}$ in a generative modeling framework. To achieve the dynamics of the conditional distribution over time, we further regularize the latent variables sampled from a prior Gaussian distribution conditional on temporal factors $\mathcal{C}$ (statistic features of $\mathcal{X}$ such as mean and start values), i.e., $p(\mathcal{Z}|\mathcal{C})$. As the distributional dependence $p(\mathcal{Y}|\mathcal{X})$ is approximated by the latent variables, it accordingly evolves to a temporal conditional distribution $p(\mathcal{Y}|\mathcal{X},\mathcal{C})$, where the dynamic dependencies between historical observations and future data vary with temporal factors and competently depict the distribution drift. 

In light of the above discussion, we propose a novel framework called temporal conditional variational autoencoder (TCVAE) to infer the distributional drift in non-stationary MTSs. TCVAE aims to address three major challenges: (a) accurately approximating the temporal Gaussian distribution $p(\mathcal{Z}|\mathcal{C})$ of the latent variables; (b) dynamically adapting the encoder and decoder structures to distributional changes; (c) flexibly inferring the temporal conditional distribution $p(\mathcal{Y}|\mathcal{X},\mathcal{C})$ from the latent variables. Regarding (a), TCVAE introduces a Hawkes process-based attention mechanism to represent temporal factors and feed-forward networks to learn the mean and variance of the Gaussian distribution $p(\mathcal{Z}|\mathcal{C})$. Subsequently, the representation of temporal factors is utilized in a gated attention mechanism that adjusts the structures of Transformer-based encoder and decoder in a meta-learning manner to handle (b). Moreover, conditional continuous normalizing flow (CCNF) is exploited to invertibly transform the Gaussian distribution into a complex distribution that is more flexible to infer the temporal conditional distribution to address (c). Our contributions are summarized as follows:
\begin{itemize}
    \item A temporal conditional variational autoencoder (TCVAE) adapts to distribution drift in MTSs. To our knowledge, TCVAE is the first attempt to depict distribution drift in a generative framework for MTS forecasting.
    \item A temporal Hawkes attention mechanism represents temporal factors that estimate the temporal Gaussian and a gated attention mechanism dynamically adapts the network structure of the encoder and decoder to distributional changes.
    \item CCNF is used to transform the temporal Gaussian to a flexible and form-free distribution for effectively inferring the temporal conditional distribution.
\end{itemize}

Extensive experimental results on six public MTS datasets show the superior robustness and effectiveness of TCVAE over state-of-the-art baselines. Case studies on traffic and pandemic data further interpret the applicability of TCVAE in real-world environments.

\section{RELATED WORK}
\subsection{Multivariate Time Series Forecasting}
In time series forecasting, traditional statistical models, such as vector autoregression (VAR)~\cite{DBLP:journals/corr/abs-1910-11800} and Gaussian process (GP)~\cite{DBLP:conf/nips/SalinasBCMG19}, are simple but interpretable for time series forecasting. However, they make strong assumptions about stationary processes and fall short in capturing non-stationary to non-IID scenarios~\cite{cj_Cao14,WangRXC21}. Recently, various deep learning models have become prevalent for MTS forecasting. RNN-based~\cite{ DBLP:conf/sigir/LaiCYL18,DBLP:journals/ml/ShihSL19,DBLP:journals/tnn/MohajerinW19}, CNN-based~\cite{DBLP:conf/cikm/HuangWWT19,DBLP:conf/kdd/DengCJST21}, Transformer-based~\cite{DBLP:conf/aaai/ZhouZPZLXZ21,DBLP:conf/iclr/LiuYLLLLD22,DBLP:conf/nips/WuXWL21,DBLP:conf/icml/ZhouMWW0022} and MLP-based~\cite{DBLP:conf/nips/YiZFWWHALCN23} models are free from stationary assumptions and are capable of modeling nonlinear long- and short-term temporal  patterns. These models pay more attention to capturing the intra-series temporal patterns, but generally neglect the inter-series couplings between multiple variables~\cite{WangRXC21}, which weakens their forecasting capacity.  
Further, the recent DNN models~\cite{DBLP:conf/ijcai/WuPLJZ19, DBLP:conf/ijcai/YuYZ18} organize MTSs by a graph with variables as nodes to portray the correlations between variables and graph neural networks (GNNs) to learn the inter-series correlations. GNNs highly depend on a predefined topology structure of inter-series correlations, hence, are hardly applicable to MTSs with evolving dependencies and distributional drift.
Subsequently, MTGNN~\cite{DBLP:conf/kdd/WuPL0CZ20}, AGCRN~\cite{DBLP:conf/nips/0001YL0020}, StemGNN~\cite{DBLP:conf/nips/CaoWDZZHTXBTZ20} and FourierGNN~\cite{DBLP:conf/nips/YiZFHHWACN23} extract the uni-directed relations among variables and capture shared patterns by graph learning rather than predefined priors. Similarly, CATN~\cite{DBLP:conf/aaai/HeZBYN22} introduces a tree (i.e., an ordered graph) to structure multivariable time series with a clear hierarchy and proposes a tree-aware network to learn the hierarchical and grouped correlations among multiple variables. 
Due to the 
complex structures of MTSs, it is difficult to learn general graphs, especially for high-dimensional data~\cite{DBLP:conf/icpr/LyuL0GSZ20}. 
More recent deep MTS models capture interactions across multiple MTSs~\cite{DBLP:conf/iclr/ZhangY23,DBLP:conf/aaai/FangRSSL00023,Yangdsaa23,10.1145/3653447} and high-dimensional dependencies~\cite{10197239}.

Due to the non-stationarity of the real-world environment, distributional drift has been an essential issue in time series data. 
However, few existing methods focus on the distributional drift of time series. Some methods~\cite{DBLP:conf/aaai/ZhouLLYZJ19, DBLP:journals/corr/abs-2204-05101,DBLP:journals/corr/abs-2205-14415,9783029, DBLP:journals/tnn/PassalisTKGI20, 9509335, DBLP:journals/corr/abs-2302-14829} strategically cater to future data via adapting their models with the newly arrived data or historical significant fluctuations. For instance, the studies~\cite{DBLP:journals/tnn/PassalisTKGI20, DBLP:journals/corr/abs-2205-14415, DBLP:journals/corr/abs-2302-14829} use static or learnable statistics to adaptively normalize the newly arrived input for stationarization and then denormalize the output for better predictability.
However, they hardly learn the intrinsic distributional drift, limiting generalization ability at the data level. 
In this light, adaptation methods~\cite{DBLP:conf/aaai/LiYLX022,DBLP:conf/aaai/YueWDYHTX22,DBLP:conf/cikm/Du0FPQXW21,DBLP:conf/cikm/YouZDFH21, DBLP:journals/tnn/SongLLZ23,10.1145/3580305.3599315} are designed to generate new training samples/features following future data distribution or calculating gradient differences between historical observations and future data. For example, DDG-DA~\cite{DBLP:conf/aaai/LiYLX022} predicts the evolution of data distribution from a meta-learning perspective through training a predictor to estimate future data distribution and generating corresponding samples for training. 
AdaRNN~\cite{DBLP:conf/cikm/Du0FPQXW21} characterizes and matches temporal distribution to generate common knowledge shared among training samples, which can be generalized well on unseen future data.
LLF~\cite{DBLP:conf/cikm/YouZDFH21} mitigates the impact of distributional drift via computing the difference between the gradients on historical data and future data.
More recently, DoubleAdapt~\cite{10.1145/3580305.3599315} proposes two meta-learners aimed at adapting data to a locally stationary distribution, while also equipping the model with task-specific parameters under mitigated distribution drifts. 
Different from the above methods neutralizing the bias caused by distributional drift, we aim to model the dynamic distributional dependencies between historical observations and future data over time.


\subsection{Variational Autoencoder}
Variational autoencoders (VAEs) as deep generative models are capable of approximating data distribution and have enjoyed great success in various real-world applications, including time series analysis~\cite{DBLP:conf/aaai/NguyenQ21, DBLP:conf/iclr/FortuinHLSR19, DBLP:journals/corr/abs-2111-08095}, dialog generation~\cite{DBLP:conf/acl/ZhaoZE17,DBLP:conf/iclr/GuCHK19}, text-to-speech~\cite{DBLP:conf/icml/KimKS21}, recommendation~\cite{DBLP:conf/sigir/AskariSS21} and image processing~\cite{DBLP:conf/iclr/PrakashKJ21}. In time series analysis, Nguyen et al.~\cite{DBLP:conf/aaai/NguyenQ21} introduced a temporal regularization and noise mechanism into a temporal latent autoencoder network to model the predictive distribution of time series implicitly. To well learn the discrete representations of time series, Fortuin et al.~\cite{DBLP:conf/iclr/FortuinHLSR19} overcame the non-differentiability of discrete representations and used a gradient-based self-organizing map algorithm. Dasai et al.~\cite{DBLP:journals/corr/abs-2111-08095} relied on user-defined distribution, such as level, trend, and seasonality, to generate interpretable time series forecasting. More recently, He et al.~\cite{DBLP:journals/tnn/HeCLWYCLZ23} combined GNN, LSTM and VAE to model spatial and temporal dependency for anomaly detection of a complex cloud system. Transparently, these models mainly focus on addressing challenges related to high-dimensionality, missing labels, and interpretability. In addition, based on the robust representation of VAE, conditional VAE regularizes distributional dependencies between historical observations and future data conditional on certain constraints. Inspired by controllable dialog generation~\cite{DBLP:conf/nips/MaCX18, DBLP:conf/iclr/GuCHK19}, we adopt temporal conditional VAE to depict the according dynamic distributional dependencies varying with temporal factors.

\subsection{Continuous Normalizing Flow}
Continuous normalizing flow (CNF) is a continuous version of normalizing flows, which replaces the layer-wise transformations of normalizing flows with ODEs. It is widely used in image/video generation~\cite{DBLP:journals/tog/AbdalZMW21, DBLP:conf/iclr/KumarBEFLDK20}, molecular graph generation~\cite{DBLP:conf/icml/LuoYJ21}, 3D face recognition~\cite{DBLP:conf/aaai/ZhangYXL21} and time series forecasting~\cite{DBLP:conf/iclr/RasulSSBV21}. 
Luo et al.~\cite{DBLP:conf/icml/LuoYJ21} proposed a discrete latent variable model GraphDF based on CNF methods for the molecular graph generation. Zhang et al.~\cite{DBLP:conf/aaai/ZhangYXL21} learned distributional representation and then transformed this distribution into a flexible form by CNF for low-quality 3D face recognition. These models focus on estimating distribution with a flexible form but conditional constraints need to be considered for capturing complex relationships among variables in reality. For example, given a set of attributes, Abdal et al.~\cite{DBLP:journals/tog/AbdalZMW21} devised a CNF-based technique to conditionally resample images with high quality from the GAN latent space. Rasul et al.~\cite{DBLP:conf/iclr/RasulSSBV21} used a CNF to represent the data distribution in an autoregressive deep learning model. 
Inspired by these two types of time point modeling processes~\cite{DBLP:conf/aaai/ZhangYXL21, DBLP:journals/tog/AbdalZMW21}, we design a CCNF to invertibly transform Gaussian distribution into a complex distribution to flexibly represent temporal conditional distribution.

\begin{figure*}[!t]
\centering
\includegraphics[width=1.0\linewidth]{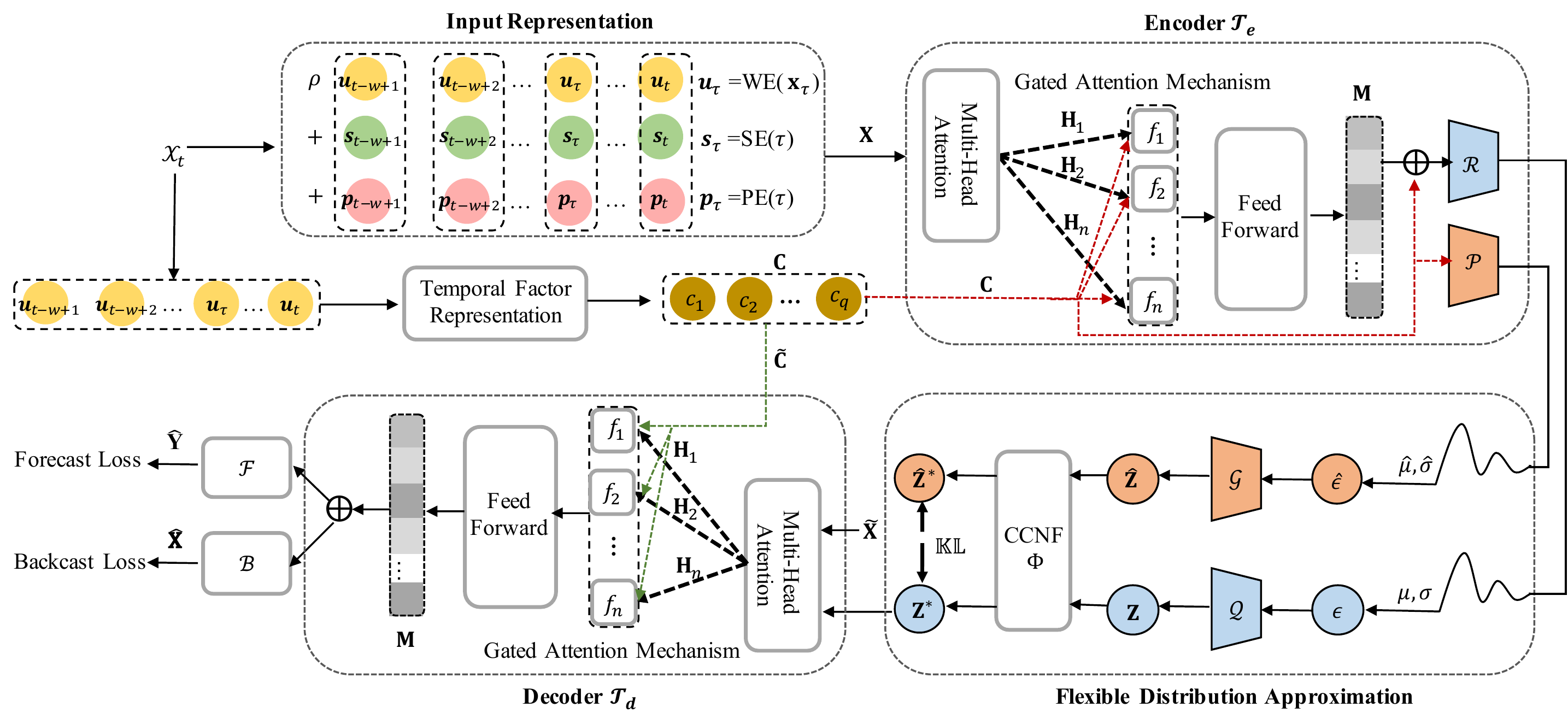}
\caption{The overall architecture of TCVAE for MTS forecasting. (1) The input $\mathcal{X}_t$ is first fed into the input representation module and $\{\bm{u}_{t-w+1},...,\bm{u}_{\tau},...,\bm{u}_t\}$ is extracted separately for temporal factor representation; (2) The encoder $\mathcal{T}_e$ takes $\bm{{\rm X}}$ as input and controls the information flow output from multiple heads by introducing temporal factors $\bm{{\rm C}}$ into gated attention. The corresponding details of the decoder $\mathcal{T}_d$ are similar to the encoder $\mathcal{T}_e$ with input $\tilde{\bm{{\rm X}}}$ and $\tilde{\bm{{\rm C}}}$; 
(3) The final loss is a combination of forecasting loss, backcasting loss, and $\mathbb{KL}$ divergence.}
\label{figure 2}
\end{figure*}

\section{PRELIMINARY}

A time series $\bm{{\rm x}}^{i}=\{x_{1}^i,x_{2}^i,...,x_{T}^i\} \in \mathbb{R}^T$ records the observed values of variable $i$ with $T$ timestamps. A multivariate time series is represented as $\mathcal{X}=\{\bm{{\rm x}}^1,\bm{{\rm x}}^2,...,\bm{{\rm x}}^i,...,\bm{{\rm x}}^{d_{x}}\} \in \mathbb{R}^{d_x \times T}$, where $d_{x}$ is the variable dimension (the number of univariate time series) and the sampling time interval between two adjacent observed values is constant for all time series. We reformulate the multivariate time series $\mathcal{X}=\{\bm{{\rm x}}_1,\bm{{\rm x}}_2,...,\bm{{\rm x}}_t,...,\bm{{\rm x}}_T\} \in \mathbb{R}^{T \times d_x}$ along the timeline, where $\bm{{\rm x}}_t$ denotes the multivariate values at timestamp $t$.

\textbf{Problem Statement}~Under the rolling forecasting setting, given a fixed input time window $w$ and time horizon $h$, we have the historical input of $w$ steps $\mathcal{X}_t = \{ \bm{{\rm x}}_{t-w+1}, \bm{{\rm x}}_{t-w+2},...,\bm{{\rm x}}_{\tau},...,\bm{{\rm x}}_t\} \in \mathbb{R}^{w \times d_x}$ at timestamp $t$, and our target is to forecast the future sequence of $h$ steps $\mathcal{Y}_t=\{\bm{{\rm y}}_{t+1}, \bm{{\rm y}}_{t+2},...,\bm{{\rm y}}_{\tau},...,\bm{{\rm y}}_{t+h}\} \in \mathbb{R}^{h \times d_y}$ successive to timestamp $t$, where $d_{y}$ is the variable dimension in forecasting, $\mathbf{x}_{\tau}$ and $\mathbf{y}_{\tau}$ denote the multivariate values at timestamp $\tau$ in the historical input and future sequence respectively. 

\textbf{Distributional Drift}~At timestamp $t$, assume $\mathcal{Y}$ is drawn from a conditional distribution $p(\mathcal{Y}|\mathcal{X})$. In the stationary case, $p(\mathcal{Y}_t|\mathcal{X}_t)$ known by a learning model at timestamp $t$ is applicable to future timestamps; however the time series data generally changes over time, a distributional drift may occur at timestamp $t$, indicating $p(\mathcal{Y}|\mathcal{X})\neq p(\mathcal{Y}_t|\mathcal{X}_t)$.



\textbf{VAE Architecture}~VAE gains creativity for its capability of accurately estimating meaningful latent distribution rather than a single data point as in traditional AE. The architecture of VAE consists of an encoder and a decoder. The encoder first encodes the input representation $\bm{{\rm X}}$ to a latent distribution and generates $\bm{{\rm Z}} \in \mathcal{Z}$ obeying this distribution, then the decoder decodes the original input representation $\bm{{\rm X}}$ from $\bm{{\rm Z}}$. The model distribution $p_{\theta}(\bm{{\rm x}})$ is formulated as:
\begin{equation} \label{eqn1}
    p_{\theta}({\bm{{\rm x}}})=\int_{\mathcal{Z}}p_{\theta}(\bm{{\rm x}}|\bm{{\rm z}})p_{\theta}(\bm{{\rm z}})d\mu(\bm{{\rm z}}), \forall \bm{{\rm x}} \in \bm{{\rm X}}, \bm{{\rm z}} \in \bm{{\rm Z}}
\end{equation}
where $p_{\theta}({\rm z})$ is a prior distribution over the latent $\bm{{\rm Z}}$ and $p_{\theta}(\bm{{\rm x}}|\bm{{\rm z}})$ is the estimated distribution. $\mu(\bm{{\rm z}})$ is the base measure of the latent space $\mathcal{Z}$. With the above architecture, while the encoder gives service to model the posterior distribution $q_{\phi}(\bm{{\rm z}}|\bm{{\rm x}})$, the decoder gives service to approximate the estimated distribution $p_{\theta}(\bm{{\rm x}}|\bm{{\rm z}})$.

\section{METHODOLOGY}
In this section, we propose TCVAE to address the distributional drift problem of MTS forecasting. The overview of the TCVAE architecture is shown in Fig.~\ref{figure 2}.

\subsection{Data Preprocessing}
To guarantee the robustness of TCVAE, the raw MTS data $\mathcal{X}=\{\bm{{\rm x}}_1,\bm{{\rm x}}_2,...,\bm{{\rm x}}_t,...,\bm{{\rm x}}_T\} \in \mathbb{R}^{T \times d_x}$ is normalized and then converted to time-series windows such as  $\mathcal{X}_t \in \mathbb{R}^{w \times d_x}$ at timestamp $t$, both for training and testing. Following~\cite{DBLP:conf/sigir/LaiCYL18,DBLP:journals/ml/ShihSL19}, we perform normalization on $\mathcal{X}$ as:
\begin{equation} \label{eqn2}
   \bm{{\rm x}}_t \leftarrow \frac{\bm{{\rm x}}_t-\min(\mathcal{X})}{\max(\mathcal{X}) -\min(\mathcal{X}) + {\epsilon}^{'}}
\end{equation}
where $\min(\cdot)$ and $\max(\cdot)$ are the minimum and maximum vectors respectively. To avoid zero-division, a constant offset ${\epsilon}^{'}=|\min(\mathcal{X})|$ is required. 

\subsection{Input Representation}
Transformer is used as the skeleton of the encoder ($\mathcal{T}_e$) and the decoder ($\mathcal{T}_d$) as shown in Fig.~\ref{figure 2}. Transformer-based models use a self-attention mechanism and timestamps served as global context. Intuitively, we regard the distributional drift problem as identifying the shifting context from global information like hierarchical timestamps (A.M. and P.M., day, week, month, year) and agnostic timestamps (holidays, special events). 
Assuming we have the multivariate values $\bm{{\rm x}}_{\tau} \in \mathcal{X}_t$ at timestamp $\tau$ and the corresponding global timestamp set $\mathcal{S}_{\tau} \in \mathbb{R}^{r}$ which contains $r=6$ types of global timestamps including month, day, week, hour, minute and A.M./P.M.. 
Following the input representation in~\cite{DBLP:conf/aaai/ZhouZPZLXZ21}, we embed global timestamps using trainable stamp embeddings ${\rm SE}(\tau)=\sum_{i=0}^{r-1}\mathbf{D}\mathcal{S}_{\tau}[i] \in \mathbb{R}^d$, where $d$ is the embedding dimension and $\mathbf{D} \in \mathbb{R}^{d \times 1}$ is an embedding matrix.
In addition, we use a fixed position embedding to preserve the local context, i.e., ${\rm PE}(pos,2j)=\sin{(pos/(2w)^{2j/d})}$, ${\rm PE}(pos,2j+1)=\cos{(pos/(2w)^{2j/d})}$, where $j \in \{1,...,\lfloor d/2 \rfloor\}$. 
Then, we adopt a token embedding, i.e., ${\rm WE}(\bm{{\rm x}}_{\tau})={\rm Conv1d}(\bm{{\rm x}}_{\tau})$ where ${\rm Conv1d}(\cdot)$ performs 1-D convolution (kernel width=3, stride=1), and projects $\bm{{\rm x}}_{\tau}$ into a $d$-dim vector. Thus, the final embedding output can be formulated as:
\begin{equation} \label{eqn3}
    \begin{split}
    \tilde{\textbf{x}}_{\tau}&=\rho{\rm WE}(\bm{{\rm x}}_{\tau})+{\rm PE}(\tau)+{\rm SE}(\tau)   \\
    \bm{{\rm X}}&=\bm{{\rm Concat}}(\{\tilde{\textbf{x}}_{\tau}\}_{\tau=t-w+1}^t)
    \end{split}
\end{equation}
where $\tilde{\textbf{x}}_{\tau}$ represents the embedding at timestamp $\tau$, $\bm{{\rm X}} \in \mathbb{R}^{w \times d}$ and $\rho$ is a learnable factor balancing the magnitude between the token embedding and position/stamp embeddings. The embeddings output from ${\rm WE}$, ${\rm PE}$ and ${\rm SE}$ are denoted as $\{\bm{u}_{t-w+1},...,\bm{u}_{\tau},...,\bm{u}_t\}$, $\{\bm{p}_{t-w+1},...,\bm{p}_{\tau},...,\bm{p}_t\}$ and $\{\bm{s}_{t-w+1},...,\bm{s}_{\tau},...,\bm{s}_t\}$ in a feature-wise manner respectively, where $\bm{u}_{\tau}={\rm WE}(\mathbf{x}_{\tau})\in \mathbb{R}^d$, $\bm{p}_{\tau}={\rm PE}(\tau) \in \mathbb{R}^d$ and $\bm{s}_{\tau}={\rm SE}(\tau)\in \mathbb{R}^d$. Let $\bm{{\rm X}}$ and $\bm{\tilde{{\rm X}}}=[\bm{{\rm X}}_t;\bm{{\rm X}}_0]$ denote the inputs of $\mathcal{T}_e$ and $\mathcal{T}_d$ respectively, where $\bm{{\rm X}}_t$ is the start token and $\bm{{\rm X}}_0$ is the target sequence that masks the variable values to zero.

\subsection{Temporal Factor Representation}
\textbf{Temporal Hawkes Attention}~
In a time slice, the values at each time have different impacts on future values. Therefore, when calculating temporal factor information, we should weigh the critical moments that affect future values~\cite{DBLP:conf/aaai/SawhneyAWDS21}. Assuming that the current time is $t$, we employ a temporal attention mechanism $\vartheta(\cdot)$ which learns to decide the critical time points (timestamps). This mechanism aggregates temporal features $\bm{{\rm U}}=[\bm{u}_{t-w+1},...,\bm{u}_{\tau},...,\bm{u}_t] \in \mathbb{R}^{d \times w}$ from different time points into an overall representation using learned attention weights $\alpha_{\tau}$ for each time point $t-w+1 \leq \tau \leq t$. We formulate this mechanism as:
\begin{equation} \label{eqn4}
    \alpha_\tau=\frac{exp(\bm{u}_\tau^{\mathrm{T}}\bm{W}\bm{{\rm U}})}{\sum_\tau exp(\bm{u}_\tau^{\mathrm{T}}\bm{W}\bm{{\rm U}})}
\end{equation}
\begin{equation} \label{eqn5}
    \zeta_\tau=\alpha_\tau \bm{u}_\tau
\end{equation}
\begin{equation} \label{eqn6}
    \vartheta(\bm{{\rm U}})=\sum_{\tau}\zeta_{\tau}
\end{equation}
where $\bm{W}$ is a learned linear transform, $\rm{T}$ denotes transpose and $\alpha_\tau$ denotes the attention weights used to aggregate all temporal features.

The Hawkes process is a self-excited time point process, where the discrete event sequence in continuous time is modeled. Each previous event excites this process to varying degrees. For example, the release of financial policies and crisis information will affect future prices corresponding to varying degrees in the stock market. Although $\vartheta(\bm{{\rm U}})$ has captured this complex excitement relationship, the hidden representation cannot well retain the original interpretable factors. Inspired by~\cite{DBLP:conf/kdd/ZhangAQ17}, we propose a temporal Hawkes attention (THA) mechanism to enhance the temporal attention mechanism $\vartheta(\cdot)$ with interpretability by using a Hawkes process. THA learns an excitation parameter $\epsilon$ corresponding to a time point $\tau$ and a decay parameter $\gamma$ to learn the decay rate of this induced excitement. We compute the temporal features $\bm{{\rm B}}$:
\begin{equation} \label{eqn7}
    \bm{{\rm B}}=\sum_{\tau=0,\Delta t_\tau \geq 0}(\zeta_\tau+\epsilon\max(\zeta_\tau,0)e^{-\gamma\Delta t_\tau})
\end{equation}
where $\Delta t_\tau$ is the time interval between the current and the past time point $\tau$. 

\textbf{Factor Extraction}~Given the temporal features $\bm{{\rm B}} \in \mathbb{R}^{w \times d}$, we manually construct temporal factors for each feature series $\bm{{\rm b}}_i \in \mathbb{R}^w$, $1 \leq i \leq d$. Specifically, we obtain the start value $sta$, end value $end$, maximum $max$, minimum $min$, med-value $med$, mean $\mu$ and standard deviation $std$ for each feature series and calculate the standard deviation of the means $\mu$ and the standard deviation $std$, i.e., $std({\mu})=std\left({\mu(\bm{{\rm b}}_i)}_{i=1}^d\right)$ and $std({std})=std\left({std(\bm{{\rm b}}_i)}_{i=1}^d\right)$ respectively. Note that we use replication operation on $std({\mu})$ and $std({std})$ to align dimensions. Accordingly, for each feature series, we construct a feature vector of $q$ temporal factors $\{sta, end, max, min, med, \mu, std, std({\mu}), std({std})\} \in \mathbb{R}^{q \times d}$, where $q=9$ is predefined. Stacking all the feature vectors, we obtain $\bm{{\rm C}} \in \mathbb{R}^{q \times d}$. The temporal factors of $\mathcal{T}_d$ are denoted as $\tilde{\bm{{\rm C}}}$, similar to those ($\bm{{\rm C}}$) of $\mathcal{T}_e$ as shown in Fig. \ref{figure 2}.



\subsection{Gated Attention Mechanism}
We believe that the encoding and decoding processes related to prediction should be adapted to distribution changes - as shown in the \textit{introduction}. Here, the traditional multi-head self-attention of encoder $\mathcal{T}_e$ and decoder $\mathcal{T}_d$ is briefly introduced. Given each training sample $\bm{{\rm X}} \in \mathbb{R}^{w \times d}$, we perform canonical scaled-dot product attention on the tuple input ($\bm{{\rm Q}}$ (query), $\bm{{\rm K}}$ (key), $\bm{{\rm V}}$ (value)) as $\mathcal{A}(\bm{{\rm Q}},\bm{{\rm K}},\bm{{\rm V}})={\rm Softmax}(\frac{\bm{{\rm Q}}\bm{{\rm K}}^T}{\sqrt{d}})\bm{{\rm V}}$, where $d$ is the input representation dimension. The ${\rm Softmax}$ operation shapes the convex combination weights for the values in $\bm{{\rm V}}$, allowing the matrix $\bm{{\rm V}}$ to be compressed into a smaller representative vector for simplified reasoning in the downstream neural network. $\mathcal{A}$ uses a $\sqrt{d}$ term to scale the weights to reduce the variance of the weights. Multi-head self-attention is applied in the form of passing $\bm{{\rm Q}}$, $\bm{{\rm K}}$, $\bm{{\rm V}}$ through $n$ heads and feed-forward layers. 

To control the information flow existing in multiple heads adapted to changing temporal factors, we propose a gated attention mechanism (GAM) that extends the above popular scalar attention mechanism by calculating a vector gate instead of a scalar value~\cite{DBLP:conf/emnlp/LaiTBK19}. The temporal factors are utilized in the vector gate with the soft or hard mode to adjust the structures of the encoder and decoder. Let $\bm{{\rm H}}_i$ stand for the $i$th head where $i \in \mathbb{Z}^n$ and $\mathcal{H}=\{\bm{{\rm H}}_1,\bm{{\rm H}}_2,...,\bm{{\rm H}}_n\}$. If we have the temporal factors $\bm{{\rm C}}$ in different time slices, the information flow passing through $\bm{{\rm H}}_i$ and guided by $\bm{{\rm C}}$ is controlled by the gate vector ${\bm g}_i$ with soft mode as ${\bm g}_i=\delta(f(\bm{{\rm C}}, \bm{{\rm H}}_i))$, where $\delta$ represents the element-wise Sigmoid function. However, the $\bm{g}_i$ of $\bm{{\rm H}}_i$ is also affected by other heads. Inspired by~\cite{DBLP:conf/emnlp/LaiTBK19, DBLP:conf/emnlp/DingL21}, we adapt the gate vector not only to temporal factors and a single head but also to temporal factors and the entire set of heads $\mathcal{H}$, see Fig.~\ref{figure 2}. To calculate the gate of head $\bm{{\rm H}}_i$, each head in the entire set of heads $\mathcal{H}$ and temporal factors will present an individual gate `vote'. Then we aggregate the votes to calculate gate ${\bm g}_i$ for $\bm{{\rm H}}_i$. The calculation process for ${\bm g}_i$ is as follows:
\begin{equation} \label{eqn8}
    \bm{v}^j=\bm{W}\bm{{\rm H}}_j+\bm{b}
\end{equation}
\begin{equation} \label{eqn9}
    \bm{v}^{\bm{{\rm C}}}=\bm{W}\bm{{\rm C}}+\bm{b}
\end{equation}
\begin{equation} \label{eqn10}
    s_i^j=\bm{{\rm H}}_i^T\bm{v}^j
\end{equation}
\begin{equation} \label{eqn11}
    s_i^{\bm{{\rm C}}}=\bm{{\rm H}}_i^T\bm{v}^{\bm{{\rm C}}}
\end{equation}
\begin{equation} \label{eqn12}
     \beta_i^j=\frac{\exp(s_i^j)}{\sum_{k \in [1,...,n]}\exp(s_i^k)+\exp(s_i^{\bm{{\rm C}}})}
\end{equation}
\begin{equation} \label{eqn13}
     \beta_i^{\bm{{\rm C}}}=\frac{\exp(s_i^{\bm{{\rm C}}})}{\sum_{k \in [1,...,n]}\exp(s_i^k)+\exp(s_i^{\bm{{\rm C}}})}
\end{equation}
\begin{equation} \label{eqn14}
     {\bm g}_i=f_i(\bm{{\rm C}},\bm{{\rm H}})=\delta(\sum_j(\beta_i^j\bm{{\rm H}}_j)+\beta_i^{\bm{{\rm C}}}\bm{{\rm C}})
\end{equation}
Here, $j \in \mathbb{Z}^n$, $\bm{W}$ and $\bm{b}$ are learnable weights and biases shared among functions $f_1,..f_n$. The parameterized function $f$ is more flexible in modeling the interaction between vectors $\bm{{\rm C}}$ and $\bm{{\rm H}}$. Vectors $\bm{v}^j$ and $\bm{v}^{\bm{{\rm C}}}$ are the outputs of $\bm{{\rm H}}_j$ and $\bm{{\rm C}}$ after linear transformation. $s_i^j$ is the unnormalized attention score that input $\bm{{\rm H}}_j$ put on $\bm{{\rm H}}_i$ and $\beta_i^j$ is the normalized score.
$s_i^{\bm{{\rm C}}}$ is the unnormalized attention score that temporal factor $\bm{{\rm C}}$ put on $\bm{{\rm H}}_i$ and $\beta_i^{\bm{{\rm C}}}$ is the normalized score.
The temporal factors are calculated together and the soft weights of $n$ heads are learned. Then we apply to calculate the multi-head self-attention as $\bm{{\rm M}}=[{\bm g}_1,...,{\bm g}_n]$, where $[\cdot,\cdot]$ denotes a concatenation operation.

\subsection{Flexible Distribution Approximation}
In MTS forecasting, the forecasting sequence and reconstruction sequence are similar generation processes~\cite{DBLP:conf/icml/KimKS21}. The standard VAE generation models assume that the latent variable $\bm{{\rm Z}}$ follows a simple prior distribution such as the Gaussian distribution $\mathcal{N}(\mu,\sigma^2I)$ with learnable parameters mean $\mu$ and variance $\sigma^2$. To make the training process differentiable, the re-parameterization trick can be adopted:
\begin{equation} \label{eqn15}
    \bm{{\rm Z}}=\mu+\varepsilon\sigma
\end{equation}
\begin{equation} \label{eqn16}
    \varepsilon \sim \mathcal{N}(0,I)
\end{equation}

However, the latent space is affected by underlying factors over time and becomes increasingly dynamic and complicated to estimate. Motivated by~\cite{DBLP:conf/iclr/GuCHK19},
we propose a flexible distribution approximation (FDA) module where temporal factors are involved in accurately approximating the prior and posterior distributions as shown in Fig.~\ref{figure 2}. We feed the temporal factors $\bm{{\rm C}}$ and the output $\bm{{\rm M}}$ of the GAM into the process of learning the latent variable. By transforming random noise $\epsilon$ using neural networks, we sample from the prior Gaussian distribution and posterior distribution over the latent variables. In particular, while the prior sample $\bm{{\hat{\rm Z}}} \sim p_{\theta}(\bm{{\hat{\rm Z}}}|\bm{{\rm C}})$ is produced by a generator $\mathcal{G}$ from temporal-dependent random noise $\hat{\epsilon}$, the approximated posterior sample $\bm{{\rm Z}} \sim q_{\phi}(\bm{{\rm Z}}|\bm{{\rm C}},\bm{{\rm M}})$ is produced by a generator $\mathcal{Q}$ from temporal-dependent random noise $\epsilon$. $\mathcal{G}$ and $\mathcal{Q}$ are feed-forward neural networks. Both $\hat{\epsilon}$ and $\epsilon$, whose mean covariance matrix are computed from $\bm{{\rm C}}$ with the prior network $\mathcal{P}$ and recognition network $\mathcal{R}$ respectively, are subject to a normal distribution:
\begin{equation} \label{eqn17}
    \left[\begin{array}{c} \hat{\mu} \\ \log\hat{\sigma}^2\end{array}\right]=\hat{\bm{W}}\mathcal{P}_{\theta}(\bm{{\rm C}})+\hat{\bm{b}}
\end{equation}
\begin{equation} \label{eqn18}
    \hat{\epsilon} \sim \mathcal{N}(\epsilon;\hat{\mu},\hat{\sigma}^2I)
\end{equation}
\begin{equation} \label{eqn19}
    \bm{{\hat{\rm Z}}}=\mathcal{G}_{\theta}(\hat{\epsilon})
\end{equation}
\begin{equation} \label{eqn20}
    \left[\begin{array}{c} \mu \\ \log{\sigma}^2\end{array}\right]=\bm{W}\mathcal{R}_{\phi}(\left[\begin{array}{c}\bm{{\rm M}} \\ \bm{{\rm C}} \end{array}\right])+\bm{b}
\end{equation}
\begin{equation} \label{eqn21}
    \epsilon \sim \mathcal{N}(\epsilon;\mu,\sigma^2I)
\end{equation}
\begin{equation} \label{eqn22}
    \bm{{\rm Z}}=\mathcal{Q}_{\phi}(\epsilon)
\end{equation}
where $\mathcal{G}_{\theta}(\cdot)$, $\mathcal{P}_{\theta}(\cdot)$, $\mathcal{Q}_{\phi}(\cdot)$ and $\mathcal{R}_{\phi}(\cdot)$ are feed-forward neural networks.

\begin{algorithm}[!t]
    \renewcommand{\algorithmicrequire}{\textbf{Require:}}
    \renewcommand{\algorithmicensure}{\textbf{}}
    \caption{The TCVAE training algorithm}
    \label{alg:1}
    \begin{algorithmic}[1]
        \REQUIRE
        window-sized input $\mathcal{X}_t$, encoder $\mathcal{T}_e$, decoder $\mathcal{T}_d$, feed-forward networks $\mathcal{G}, \mathcal{Q}, \mathcal{F}, \mathcal{B}$, iteration limit $N$, learning rate $\gamma$
        \STATE Initialize model parameters $\Theta$
        \STATE $n \leftarrow 0$
        \WHILE{$n \textless N$}
            \STATE $\bm{{\rm X}}, \bm{\tilde{{\rm X}}}, \bm{{\rm C}}, \bm{\tilde{{\rm C}}} \leftarrow \mathcal{X}_t$ by Equations (\ref{eqn2})-(\ref{eqn7})
            \STATE $ \mu,\sigma^2, \hat{\mu},\hat{\sigma}^2 \leftarrow \mathcal{T}_e(\bm{{\rm X}}, \bm{{\rm C}})$
            \STATE $\bm{\hat{{\rm Z}}} \leftarrow \mathcal{G}_{\theta}(\hat{\epsilon})$, $\hat{\epsilon} \leftarrow$ $\hat{\mu}, \hat{\sigma}^2, \epsilon$ by Equations (\ref{eqn18}), (\ref{eqn19})
            \STATE $\bm{{\rm Z}} \leftarrow \mathcal{Q}_{\phi}(\epsilon), \epsilon \leftarrow \mu, \sigma^2, \epsilon$ by Equations (\ref{eqn21}), (\ref{eqn22})
            \STATE Compute the transferred latent variable, i.e. $q(\bm{{\rm Z}}^{\ast}|\bm{{\rm M}},\bm{{\rm C}})$ and $p(\bm{{\hat{\rm Z}}}^{\ast}|\bm{{\rm C}})$ by Equation (\ref{eqn25})
            \STATE $\bm{\hat{{\rm Y}}} \leftarrow \mathcal{F}(\mathcal{T}_d(\bm{\tilde{{\rm X}}}, \bm{\tilde{{\rm C}}}, \bm{{\rm Z}}^{\ast}))$, $\bm{\hat{{\rm X}}} \leftarrow \mathcal{B}(\mathcal{T}_d(\bm{\tilde{{\rm X}}}, \bm{\tilde{{\rm C}}}, \bm{{\rm Z}}^{\ast}))$
            \STATE Compute the stochastic gradient of $\Theta$ based on $\mathcal{L}(\Theta)$ by Equation (\ref{eqn26}), (\ref{eqn27})
            \STATE Update $\Theta$ based on the gradients and learning rate $\gamma$ 
            \STATE $n \leftarrow n+1$
        \ENDWHILE
    \end{algorithmic}
\end{algorithm}

\textbf{Conditional Continuous Normalizing Flow}~The posterior distribution $q(\bm{{\rm Z}}|\bm{{\rm M}},\bm{{\rm C}})$ is intractable but usually approximated by a Gaussian distribution. Nevertheless, it is unrealistic to think of the posterior as a Gaussian distribution. Regardless of the family of distributions we choose to estimate the posterior, it may not fit. Inspired by~\cite{DBLP:conf/aaai/ZhangYXL21}, we use CCNF to invertibly transform the Gaussian distribution into a complex distribution for flexibly inferring the temporal conditional distribution. First, a latent variable obeys a Gaussian distribution $\mathcal{N}(\mu,\sigma)$. A CCNF ${\Phi}$ is then used to convert the sample of this Gaussian distribution. After applying continuous-time dynamics, the latent variable $\bm{{\rm Z}}^{\ast} \in \mathbb{R}^{w \times k}$, where $k$ denotes the representation dimension of the latent variable, is computed as follows:
\begin{equation} \label{eqn23}
    \bm{{\rm Z}}^{\ast}=\Phi(\bm{{\rm Z}}(t_0))=\bm{{\rm Z}}(t_0)+\int_{t_0}^{t_1}\Omega(\bm{{\rm Z}}(t),t)dt 
\end{equation}
\begin{equation} \label{eqn24}
    \bm{{\rm Z}}(t_0) \sim \mathcal{N}(\mu, \sigma)
\end{equation}
where $\bm{{\rm Z}}=\bm{{\rm Z}}(t_0)$, $\bm{{\rm Z}}^{\ast}=\bm{{\rm Z}}(t_1)$, and $\Omega$ is a continuous mapping as $\Omega(\bm{{\rm Z}}(t),t)=\frac{\partial \bm{{\rm Z}}(t)}{\partial t}$ with the initial value $\bm{{\rm Z}}(t_0)=\bm{{\rm Z}}_0$. 
Therefore, the log-density of the transferred latent variable is:
\begin{equation} \label{eqn25}
    \log q(\bm{{\rm Z}}^{\ast}|\bm{{\rm M}},\bm{{\rm C}})=\log q(\bm{{\rm Z}}(t_0)|\bm{{\rm M}},\bm{{\rm C}})-\int_{t_0}^{t_1}{\rm Tr}(\frac{\partial \Omega}{\partial \bm{{\rm Z}}(t)})dt
\end{equation}

By this means, the Gaussian distribution is turned into a form-free posterior. The same operation is also performed on $\bm{{\hat{\rm Z}}}$ to obtain $\bm{{\hat{\rm Z}}}^{\ast}$. This transformation facilitates our model to learn a more flexible distribution $p(\bm{{\hat{\rm Z}}}^{\ast}|\bm{{\rm C}})$ and $q(\bm{{\rm Z}}^{\ast}|\bm{{\rm M}},\bm{{\rm C}})$. CCNF can potentially learn a less entangled internal representation.

\subsection{Learning Objective}
We input the flexible posterior distribution $q(\bm{{\rm Z}}^{\ast}|\bm{{\rm M}},\bm{{\rm C}})$, temporal factors $\tilde{\bm{{\rm C}}}$ and input representation $\tilde{\bm{{\rm X}}}$ together into the decoder $\mathcal{T}_d$  for decoding. The decoding results flow to the backcasting network $\mathcal{B}$ and the forecasting network $\mathcal{F}$ and generate outputs denoted by $\bm{\hat{{\rm X}}}$ and $\bm{\hat{{\rm Y}}}$, respectively. Given the generated waveforms $\bm{\hat{{\rm X}}}$ and $\bm{\hat{{\rm Y}}}$, we thus want to maximize the variational lower bound, also alluded to as the evidence lower bound (ELBO): 
\begin{equation} \label{eqn26}
    \begin{split}
        \max\limits_{\theta,\phi,\psi,\Upsilon}&\mathbb{E}_{q_{\phi}(\bm{{\rm Z}}^{\ast}|\bm{{\rm M}},\bm{{\rm C}})}\log p_{\psi}(\bm{\hat{{\rm X}}}|\bm{{\rm Z}}^{\ast},\tilde{\bm{{\rm C}}},\tilde{\bm{{\rm X}}})+\mathbb{E}_{q_{\phi}(\bm{{\rm Z}}^{\ast}|\bm{{\rm M}},\bm{{\rm C}})} \\
        &\log p_{\Upsilon}(\bm{\hat{{\rm Y}}}|\bm{{\rm Z}}^{\ast},\tilde{\bm{{\rm C}}},\tilde{\bm{{\rm X}}})-\lambda\mathbb{KL}(q_{\phi}(\bm{{\rm Z}}^{\ast}|\bm{{\rm M}},\bm{{\rm C}})||p_{\theta}(\bm{{\hat{\rm Z}}}^{\ast}|\bm{{\rm C}})) 
    \end{split}
\end{equation}
\begin{equation} \label{eqn27}
    \begin{split}
       &\mathbb{KL}(q_{\phi}(\bm{{\rm Z}}^{\ast}|\bm{{\rm M}},\bm{{\rm C}})||p_{\theta}(\bm{{\hat{\rm Z}}}^{\ast}|\bm{{\rm C}})) \\
        &=\mathbb{E}_{q_{\phi}(\bm{{\rm Z}}^{\ast}|\bm{{\rm M}},\bm{{\rm C}})}\log q_{\phi}(\bm{{\rm Z}}^{\ast}|\bm{{\rm M}},\bm{{\rm C}})-\mathbb{E}_{p_{\theta}(\bm{{\hat{\rm Z}}}^{\ast}|\bm{{\rm C}})}\log p_{\theta}(\bm{{\hat{\rm Z}}}^{\ast}|\bm{{\rm C}}) \\
        &=\mathbb{E}_{q_{\phi}(\bm{{\rm Z}}^{\ast}|\bm{{\rm M}},\bm{{\rm C}})}[\frac{{\bm{{\rm Z}}^{\ast}}^2}{2}-\int_{t_0}^{t_1}{\rm Tr}(\frac{\partial \Omega}{\partial \bm{{\rm Z}}(t)})dt]\\
        &\qquad -\mathbb{E}_{p_{\theta}(\bm{{\hat{\rm Z}}}^{\ast}|\bm{{\rm C}})}[\frac{{\bm{{\hat{\rm Z}}}^{\ast}}^2}{2}-\int_{t_0}^{t_1}{\rm Tr}(\frac{\partial \Omega}{\partial \bm{{\hat{\rm Z}}}(t)})dt]
    \end{split}
\end{equation}
where $p_{\theta}(\bm{{\hat{\rm Z}}}^{\ast}|\bm{{\rm C}})$ and $q_{\phi}(\bm{{\rm Z}}^{\ast}|\bm{{\rm M}},\bm{{\rm C}})$ are neural networks implementing Equations (\ref{eqn17})-(\ref{eqn25}). $p(\bm{{\hat{\rm Z}}}^{\ast}|\bm{{\rm C}})$ is the prior distribution given condition $\bm{{\rm C}}$. $p_{\psi}(\bm{\hat{{\rm X}}}|\bm{{\rm Z}}^{\ast},\tilde{\bm{{\rm C}}},\tilde{\bm{{\rm X}}})$ is the integration of decoder $\mathcal{T}_d$ and backcasting network $\mathcal{B}$. $p_{\Upsilon}(\bm{\hat{{\rm Y}}}|\bm{{\rm Z}}^{\ast},\tilde{\bm{{\rm C}}},\tilde{\bm{{\rm X}}})$ is the integration of decoder $\mathcal{T}_d$ and forecasting network $\mathcal{F}$. $\lambda$ is a trade-off parameter. 

The pseudo-code for the overall training process is summarized in Algorithm \ref{alg:1}. We aim at minimizing the training loss $\mathcal{L}(\Theta)$, i.e., the negative ELBO that can be considered as the sum of reconstruction loss $-\log p_{\psi}(\bm{\hat{{\rm X}}}|\bm{{\rm Z}}^{\ast},\tilde{\bm{{\rm C}}},\tilde{\bm{{\rm X}}})$, forecasting loss $-\log p_{\psi}(\bm{\hat{{\rm Y}}}|\bm{{\rm Z}}^{\ast},\tilde{\bm{{\rm C}}},\tilde{\bm{{\rm X}}})$ and Kullback-Leibler $\mathbb{KL}$ divergence $q_{\phi}(\bm{{\rm Z}}^{\ast}|\bm{{\rm M}},\bm{{\rm C}})||p_{\theta}(\bm{{\hat{\rm Z}}}^{\ast}|\bm{{\rm C}})$. $\Theta$ is the model parameter and we set the loss weight for all experiments.

\begin{table*}[!t]
 \centering
 \newcommand{\tabincell}[2]{\begin{tabular}{@{}#1@{}}#2\end{tabular}}
 \caption{Dataset Statistics, Where Larger Average ADF Test Statistic Indicates More Severe Distribution Drifts}
 \label{table 1}  
 \begin{tabular}{c|c|c|c|c|c|c}
  \toprule 
  Datasets & Traffic & Electricity & Solar-Energy & COVID-19 & PeMSD7(M) & METR-LA \\ 
  \midrule 
  Samples & 10,560 & 25,968 & 52,560 & 673 & 11,232 & 34,272 \\
  Variables & 963 & 370 & 137 & 280 & 228 & 207 \\ 
  Sample Interval & 1 hour & 1 hour & 10 minutes & 1 day & 5 minutes & 5 minutes \\ 
  Start Time & 1/1/2015 & 1/1/2011 & 1/1/2006 & 1/22/2020 & 7/1/2016 & 9/1/2018 \\ 
  Average ADF Test Statistic & -9.89 & -7.62 & -37.23 & 0.06 & -14.20 & -16.03 \\
  \bottomrule 
 \end{tabular}
\end{table*}

\section{EXPERIMENTS}

\subsection{Datasets and Metrics}
We select the following six representative real-world multivariate time series datasets from different application scenarios for experimental evaluation and summarize them in Table~\ref{table 1}. Real-world datasets suffer from non-stationarity, i.e., the distribution changes over time. Following~\cite{DBLP:journals/corr/abs-2205-14415}, we especially chose the Augmented Dick-Fuller (ADF) test statistic as the metric to quantitatively measure the degree of distribution drift. A larger average ADF test statistic means a higher level of non-stationarity, i.e., more severe distribution drifts. According to the average ADF test, we can see that our datasets exhibit high levels of non-stationarity derived from the distribution drift in the datasets. For instance, the METR-LA dataset exhibits intense and irregular fluctuations and temporal correlations and trends, as illustrated in Fig.~\ref{figure 1}. These patterns are significantly distinct from normal fluctuation, which typically appears uniform, regular, and unrelated to variables or temporal factors.



\textbf{Traffic\footnote{\url{https://archive.ics.uci.edu/ml/datasets/PEMS-SF}}} It records the hourly occupancy rate of different lanes on San Francisco highway from 2015 to 2016 and contains 10,560 timestamps and 963 sensors.

\textbf{Electricity\footnote{\url{https://archive.ics.uci.edu/ml/datasets/ElectricityLoadDiagrams20112014}}} It collects the hourly electricity consumption of 370 clients and involves 25,968 timestamps.   

\textbf{Solar-Energy\footnote{\url{https://www.nrel.gov/grid/solar-power-data.html}}}~It collects the solar power production records every $10$ minutes from 137 PV plants in Alabama State in 2006 and includes 52,560 timestamps. 

\textbf{COVID-19\footnote{\url{https://github.com/CSSEGISandData/COVID-19/tree/master}}}~The dataset records the daily number of newly confirmed cases from 1/22/2020 to 11/25/2021 collected from the COVID-19 Data Repository. 

\textbf{PeMSD7(M)\footnote{\url{http://pems.dot.ca.gov/?dnode=Clearinghouse&type=station_5min&district_id=7&submit=Submit}}}~It is from the sensors spanning the freeway system of California and consists of 11,232 timestamps and 228 variables at a 5-minute interval.    

\textbf{METR-LA\footnote{\url{https://github.com/liyaguang/DCRNN}}}~The dataset records traffic information from 207 sensors of the Los Angeles County's highway. It contains 34,272 timestamps and 207 variables at a 5-minute interval.

Following previous works~\cite{DBLP:conf/nips/CaoWDZZHTXBTZ20, DBLP:conf/nips/0001YL0020}, we evaluate all comparative methods via three metrics: Mean Absolute Error $MAE=\frac{1}{N}\sum_{i=1}^N\left|\bm{{\rm y}}_i-\hat{\bm{{\rm y}}}_i\right|$, Root Mean Squared Error $RMSE=\sqrt{\frac{1}{N}\sum_{i=1}^N(\bm{{\rm y}}_i-\hat{\bm{{\rm y}}}_i)^2}$ and Mean Absolute Percent Error $MAPE=\frac{1}{N}\sum_{i=1}^N\frac{\left|\bm{{\rm y}}_i-\hat{\bm{{\rm y}}}_i\right|}{\bm{{\rm y}}_i}\mathds{1}\{\left|\bm{{\rm y}}_i\right|>0\}$,
where $N$ denotes the number of samples in the testing set, $\hat{\bm{{\rm y}}}_i$ and $\bm{{\rm y}}_{i}$ denote the prediction and groundtruth respectively.

\subsection{Baselines}
We elaborately compare our proposed TCVAE with the following eleven competitive MTS forecasting baselines from different classes. To be specific, two representative RNN-based methods (LSTNet and AdaRNN), three CNN-based methods (DSANet, STNorm and TS2VEC) and four advanced Transformer-based methods (Informer, Autoformer, FEDformer, and Nstatformer) are good at capturing temporal patterns with different ranges such as short- and long-range, wherein, TS2VEC, AdaRNN and Nstatformer utilize special designs to tackle non-stationary time series. Additionally, StemGNN and AGCRN are based on graph neural networks. All baselines adopt the same normalization method as TCVAE.

(1) \textbf{LSTNet}~\cite{DBLP:conf/sigir/LaiCYL18}:~It employs two key components including a convolutional neural network and an LSTM with recurrent-skip connection in the time dimension;

(2) \textbf{AdaRNN}~\cite{DBLP:conf/cikm/Du0FPQXW21}:~It proposes a temporal distribution characterization module and a temporal distribution matching module for non-stationary MTS forecasting;

(3) \textbf{DSANet}~\cite{DBLP:conf/cikm/HuangWWT19}:~It constructs global and local temporal convolution to extract complicated temporal patterns and employs self-attention to model dependencies;

(4) \textbf{STNorm}~\cite{DBLP:conf/kdd/DengCJST21}:~It employs temporal and spatial normalization modules which separately refine the local component and the high-frequency component underlying the raw data. We use the version with Wavenet as the backbone;


(5) \textbf{StemGNN}~\cite{DBLP:conf/nips/CaoWDZZHTXBTZ20}:~It uses a graph network to capture inter-series and intra-series correlation jointly in the spectral domain;

(6) \textbf{AGCRN}~\cite{DBLP:conf/nips/0001YL0020}:~It has a graph-learning component and a personalized RNN component to extract fine-grained spatial and temporal correlations automatically;

(7) \textbf{TS2VEC}~\cite{DBLP:conf/aaai/YueWDYHTX22}:~It focuses on using timestamp masking and random cropping to learn augmented context views for dealing with level shifts of time series data;

(8) \textbf{Informer}~\cite{DBLP:conf/aaai/ZhouZPZLXZ21}:~It is a Transformer-like architecture with a sparse-attention mechanism to maintain a higher capacity for long sequence prediction;

(9) \textbf{Autoformer}~\cite{DBLP:conf/nips/WuXWL21}:~It designs a decomposition framework with an efficient and accurate auto-correlation mechanism for long-term MTS forecasting;

(10) \textbf{FEDformer}~\cite{DBLP:conf/icml/ZhouMWW0022}:~It introduces a frequency-enhanced attention mechanism with Fourier and Wavelet transform and a mixture of experts decomposition to extract trend components for long-term MTS forecasting;

(11) \textbf{Nstatformer}~\cite{DBLP:journals/corr/abs-2205-14415}:~It introduces a series stationarization module with statistics for better predictability and a destationary attention module to re-integrate the inherent non-stationary information for non-stationary MTS forecasting.

\begin{table*}[!t]\footnotesize
\setlength{\tabcolsep}{2.5pt}
 \centering
 \caption{Overall Comparison of Different Methods on Six Datasets with $w=24, h=24$}
 \label{table 2}
 \newcommand{\tabincell}[2]{\begin{tabular}{@{}#1@{}}#2\end{tabular}}
 \begin{tabular}{c|cccccccccccccccccc}
  \toprule
  Datasets & \multicolumn{3}{c}{Traffic} & \multicolumn{3}{c}{Electricity} & \multicolumn{3}{c}{Solar-Energy} & \multicolumn{3}{c}{COVID-19} & \multicolumn{3}{c}{PeMSD7(M)} & \multicolumn{3}{c}{METR-LA} \\
  \cmidrule(r){2-4} \cmidrule(r){5-7} \cmidrule(r){8-10} \cmidrule(r){11-13} \cmidrule(r){14-16} \cmidrule(r){17-19}
  Metrics & MAE & RMSE & MAPE & MAE & RMSE & MAPE & MAE & RMSE & MAPE & MAE & RMSE & MAPE & MAE & RMSE & MAPE & MAE & RMSE & MAPE     \\ 
  \midrule 
  LSTNet & 0.067 & 0.099 & 1.971 & 0.139 & 0.191 & 0.568 & 0.169 & 0.254 & 4.437 & 0.319 & 0.392 & 0.467 & 0.155 & 0.183 & 0.339 & 0.189 & 0.279 & 1.370 \\
  AdaRNN & 0.043 & 0.070 & 0.894 & 0.108 & 0.162 & 0.601 & 0.106 & 0.123 & \underline{2.564} & 0.423 & 0.482 & 0.500 & 0.093 & 0.133 & 0.272 & 0.163 & 0.227 & \textbf{0.167} \\
  DSANet & 0.059 & 0.089 & 1.236 & 0.084 & 0.128 & 5.948 & 0.153 & 0.230 & 4.247 & 0.164 & 0.230 & 0.720 & 0.154 & 0.198 & 0.337 & 0.168 & 0.316 & 0.228 \\
  STNorm & 0.056 & 0.080 & 1.143 & 0.153 & 0.167 & 2.476 & 0.120 & 0.167 & 3.898 & 0.263 & 0.264 & 0.541 & 0.094 & 0.178 & 0.247 & 0.119 & 0.240 & 0.196 \\
  StemGNN & 0.047 & 0.076 & 1.129 & 0.193 & 0.268 & 0.607 & 0.086 & 0.110 & 3.848 & 0.528 & 0.635 & 0.530 & \underline{0.077} & 0.123 & 0.232 & \underline{0.116} & \underline{0.209} & \textbf{0.167} \\
  AGCRN & 0.041 & 0.079 & \textbf{0.469} & \underline{0.062} & 0.110 & \underline{0.293} & 0.082 & 0.175 & 3.459 & \underline{0.110} & 0.182 & 0.192 & 0.083 & 0.150 & 0.233 & 0.129 & 0.269 & 0.790 \\
  TS2VEC & 0.041 & \underline{0.068} & 0.883 & 0.114 & 0.163 & 0.574 & 0.088 & \underline{0.104} & 2.854 & 0.296 & 0.407 & 0.366 & 0.078 & \underline{0.119} & 0.216 & 0.135 & 0.221 & 0.174 \\
  Informer & 0.045 & 0.073 & 0.876 & 0.070 & 0.105 & 0.330 & 0.078 & 0.109 & 3.747 & 0.318 & 0.391 & 0.419 & 0.079 & 0.125 & 0.215 & 0.133 & 0.242 & 0.219 \\
  Autoformer & 0.048 & 0.077 & 1.065 & 0.065 & 0.092 & 0.388 & 0.150 & 0.205 & 3.620 & 0.116 & 0.161 & 0.177 & 0.124 & 0.179 & 0.307 & 0.159 & 0.265 & 0.224 \\
  FEDformer & 0.045 & 0.072 & 0.979 & \underline{0.062} & 0.089 & 0.380 & 0.091 & 0.131 & 4.205 & 0.117 & 0.155 & 0.212 & 0.104 & 0.160 & 0.271 & 0.139 & 0.246 & 0.205 \\
  Nstatformer & \underline{0.040} & \underline{0.068} & \underline{0.531} & \textbf{0.061} & \underline{0.076} & 0.323 & \underline{0.077} & 0.147 & \textbf{2.329} & \underline{0.110} & \underline{0.152} & \underline{0.170} & 0.079 & 0.136 & \underline{0.209} & 0.119 & 0.240 & \underline{0.171} \\
  TCVAE & \textbf{0.034} & \textbf{0.055} & 0.553 & \textbf{0.061} & \textbf{0.073} & \textbf{0.286} & \textbf{0.075} & \textbf{0.099} & 3.069 & \textbf{0.109} & \textbf{0.145} & \textbf{0.167} & \textbf{0.075} & \textbf{0.111} & \textbf{0.208} & \textbf{0.104} & \textbf{0.156} & 0.467 \\
  \bottomrule 
 \end{tabular}
\end{table*}

\subsection{Implementation Details and Settings}
Following the commonly-used settings in~\cite{DBLP:conf/icde/CirsteaYGKP22,DBLP:conf/nips/0001YL0020}, we predict future $h=24$ timestamps using historical $w=24$ timestamps. Additionally, we vary the prediction timestamps in a range of $h \in \{48,72,96,180,720\}$ on PeMSD7(M) and Solar-Energy respectively to investigate the performance of long-term forecasting. We use an Adam optimizer with a fixed learning rate of 0.001, a batch size of 64, and epochs of 50 to train our model. 
In addition, we elaborately tune the hyperparameters and select the settings with the best performance: specifically, the trade-off parameter $\lambda \in \{1,10^{-1},10^{-2},10^{-3}\}$, the number of heads $n \in \{4,8,16\}$, the representation dimension $d \in \{256,512,1024\}$ and the latent variable dimension $k \in \{256,512,1024\}$. 
Regarding the rolling step, we consider a commonly-used setting~\cite{DBLP:conf/aaai/ZhouZPZLXZ21, DBLP:conf/aaai/NguyenQ21, DBLP:conf/aaai/HeZBYN22,10285474} for a multi-step forecasting setup that the rolling step is fixed to 1. In the parameter sensitivity experiment, the window size $w$ is selected from $\{24,48,96\}$. We implement TCVAE and other baselines on Python 3.6.13 with the package Pytorch 1.8.1. The codes of our model and baselines are carried out on Ubuntu 18.04.6 LTS, with one Inter(R) Xeon(R) CPU @ 2.10GHz and four NVIDIA GeForce 3090 GPU cards.

\subsection{Results}
\subsubsection{Overall Comparison}
We adopt three widely used metrics, MAE, RMSE, MAPE, to measure the performance of our proposed TCVAE and all the comparative models. Table \ref{table 2} shows the overall experimental results for the default forecasting setup of $w = 24$ and $h = 24$, where the best results are highlighted in bold and the suboptimal results are underlined. A smaller value indicates better performance. Remarkably, TCVAE establishes a new state-of-the-art and achieves 0.91$\%$-15$\%$ MAE improvement, 3.95$\%$-25.36$\%$ RMSE improvement and 0.48$\%$-2.39$\%$ MAPE improvement over all baselines on most of the datasets. Especially for some of the datasets, such as Traffic and METR-LA, the improvement is even more significant (over 10$\%$). 
Regarding the temporal-aware sequential models, the Transformer-based models, i.e., Informer, Autoformer and FEDformer, achieve obviously better performance over RNN- and CNN-based models (including LSTNet, DSANet and STNorm), showing the benefit of Transformers or variants combining various attention mechanisms in capturing temporal dependencies in modeling sequential data.
Compared with the temporal-aware models, the GNN-based models, i.e., StemGNN and AGCRN, can capture dynamic correlations explicitly among multiple time series and show better performance on several datasets. For example, StemGNN shows competitive performance on PeMSD7(M) and METR-LA, and AGCRN achieves desirable performance on Traffic and Electricity. The reason lies in that the models use parallel parameter space for time series to model temporal and topology dynamics jointly without the need for a predefined graph structure. However, the GNN-based models, especially AGCRN, require higher memory space compared with other models when dealing with high-dimensional data such as Traffic (963 dimensions).
In addition, the above GNN-based models and temporal-aware sequence models have performance limitations and fall behind Nstatformer and our proposed TCVAE, this is attributed to the fact that the distributions of adjacent time slices are diverse and change over time and the potential of current deep models is still constrained by such non-stationary data. Nstatformer tackles the issue by recovering the intrinsic non-stationary information into temporal dependencies and achieves impressive performance second only to TCVAE. TS2VEC and AdaRNN also deal with the distribution drift problem, but their performance is unsatisfactory on some datasets, which may be due to the limitations of RNN and CNN as backbones for learning temporal dependencies. From all the results, we conclude that, in addition to capturing temporal information and inter-series correlations, using temporal factors to guide distributional drift learning is important and effective.

\begin{table*}[!t]\footnotesize
\setlength{\tabcolsep}{3.5pt}
 \centering
 \caption{Overall Accuracy on PeMSD7(M) and Solar-Energy with $w=24, h=\{48,72,96,180,720\}$}
 \label{table 3}
 \newcommand{\tabincell}[2]{\begin{tabular}{@{}#1@{}}#2\end{tabular}}
 \begin{tabular}{c|c|ccc|ccc|ccc|ccc|ccc}
  \toprule
  \multirow{2}{*}{Datasets} & \multirow{2}{*}{Methods} & \multicolumn{3}{c|}{$h=48$} & \multicolumn{3}{c|}{$h=72$} & \multicolumn{3}{c|}{$h=96$} & \multicolumn{3}{c|}{$h=180$} & \multicolumn{3}{c}{$h=720$} \\
  \cline{3-17}
  & & MAE & RMSE & MAPE & MAE & RMSE & MAPE & MAE & RMSE & MAPE & MAE & RMSE & MAPE & MAE & RMSE & MAPE \\ 
  \midrule 
  & AdaRNN & 0.097 & 0.137 & 0.281 & \underline{0.089} & 0.129 & 0.278 & 0.092 & \underline{0.135} & 0.260 & 0.112 & 0.157 & 0.322 & 0.147 & 0.213 & 0.503 \\
  & StemGNN & 0.098 & 0.150 & 0.320 & 0.106 & 0.162 & 0.364 & 0.117 & 0.175 & 0.410 & 0.140 & 0.205 & 0.507 & 0.142 & 0.204 & 0.501 \\
  PeMSD7(M) & AGCRN & 0.088 & 0.157 & 0.240 & 0.090 & 0.156 & 0.246 & \underline{0.086} & 0.152 & 0.223 & \textbf{0.094} & 0.162 & 0.264 & 0.119 & \underline{0.159} & \underline{0.262} \\
  & TS2VEC & 0.088 & \underline{0.132} & 0.255 & 0.090 & 0.135 & 0.270 & 0.092 & 0.137 & 0.279 & 0.114 & 0.159 & 0.285 & \underline{0.116} & 0.164 & 0.302  \\
  & Informer & \textbf{0.085} & 0.134 & \underline{0.212} & \underline{0.089} & \underline{0.127} & \underline{0.207} & \textbf{0.081} & \textbf{0.128} & \underline{0.218} & 0.111 & \underline{0.154} & \underline{0.252} & 0.124 & 0.178 & 0.401 \\
  & Autoformer & 0.145 & 0.206 & 0.392 & 0.160 & 0.240 & 0.466 & 0.167 & 0.242 & 0.468 & 0.195 & 0.274 & 0.532 & 0.179 & 0.256 & 0.479 \\
  & FEDformer & 0.125 & 0.186 & 0.354 & 0.125 & 0.182 & 0.354 & 0.127 & 0.186 & 0.365 & 0.174 & 0.240 & 0.460 & 0.180 & 0.260 & 0.493 \\
  & Nstatformer & 0.093 & 0.165 & 0.279 & 0.106 & 0.182 & 0.323 & 0.115 & 0.194 & 0.352 & 0.133 & 0.214 & 0.411 & 0.128 & 0.208 & 0.398  \\
  & TCVAE & \underline{0.087} & \textbf{0.126} & \textbf{0.189} & \textbf{0.079} & \textbf{0.124} & \textbf{0.196} & 0.105 & 0.144 & \textbf{0.214} & \underline{0.110} & \textbf{0.150} & \textbf{0.218} & \textbf{0.111} & \textbf{0.154} & \textbf{0.238} \\
  \midrule
  & AdaRNN & 0.121 & 0.217 & \textbf{2.904} & 0.166 & 0.248 & 5.290 & 0.164 & 0.260 & 5.592 & 0.168 & 0.266 & 5.457 & 0.180 & 0.271 & 5.685 \\
  & StemGNN & 0.098 & 0.161 & 4.044 & 0.113 & 0.180 & 6.100 & 0.122 & 0.186 & 6.215 & 0.112 & 0.172 & 5.661 & \textit{OOM} & \textit{OOM} & \textit{OOM}  \\
  Solar-Energy & AGCRN & 0.161 & 0.249 & 3.660 & 0.150 & 0.230 & 3.647 & 0.118 & 0.194 & 3.545 & 0.137 & 0.208 & \underline{3.548} & \textit{OOM} & \textit{OOM} & \textit{OOM} \\ 
  & TS2VEC & 0.111 & 0.153 & 4.587 & 0.130 & 0.172 & 5.647 & 0.128 & 0.169 & 5.540 & 0.114 & 0.159 & 4.894 & 0.125 & 0.173 & 5.506   \\
  & Informer & \underline{0.079} & \underline{0.115} & 4.043 & \textbf{0.081} & \underline{0.123} & 4.284 & \underline{0.099} & \underline{0.134} & 5.571 & \underline{0.089} & \underline{0.137} & 5.612 & \textbf{0.092} & \underline{0.141} & 5.804 \\
  & Autoformer & 0.221 & 0.282 & 5.097 & 0.205 & 0.260 & 5.137 & 0.208 & 0.270 & 3.859 & 0.206 & 0.275 & 4.481 & 0.200 & 0.249 & \underline{4.704} \\
  & FEDformer & 0.147 & 0.207 & 4.923 & 0.141 & 0.183 & 4.464 & 0.129 & 0.174 & 4.728 & 0.122 & 0.180 & 5.064 & 0.135 & 0.190 & 5.643 \\
  & Nstatformer & 0.128 & 0.215 & \underline{3.258} & 0.162 & 0.256 & \underline{3.514} & 0.166 & 0.259 & \underline{3.530} & 0.151 & 0.243 & 4.132 & 0.170 & 0.270 & 4.747 \\
  & TCVAE & \textbf{0.076} & \textbf{0.111} & 3.507 & \underline{0.093} & \textbf{0.121} & \textbf{3.472} & \textbf{0.093} & \textbf{0.126} & \textbf{3.510} & \textbf{0.082} & \textbf{0.124} & \textbf{3.545} & \underline{0.102} & \textbf{0.139} & \textbf{3.534} \\
  \bottomrule 
 \end{tabular}
\end{table*}

To investigate the ability of the comparative models in long sequence forecasting, we further conduct experiments by fixing $w = 24$ and enlarging $h$ to 48, 72, 96, 180 and 720. The experimental results are shown in Table \ref{table 3}, where the best results are highlighted in bold, the suboptimal results are underlined, and \textit{OOM} denotes running out of memory. We compare TCVAE with all baselines on PeMSD7(M) and Solar-Energy except for LSTNet, DSANet, and STNorm, which are the bottom-3 baselines from the above experiments. From the results, we can observe: 
(1) TCVAE significantly outperforms most of the baselines and achieves 3.80$\%$-11.24$\%$ MAE improvement, 1.63$\%$-9.49$\%$ RMSE improvement and 0.08$\%$-24.87$\%$ MAPE improvement over all baselines on most of the datasets, indicating the superiority of TCVAE in long sequence forecasting; 
(2) it is worth noting that AGCRN highly benefits from modeling inter-series correlations within a topological structure and achieves suboptimal and even better performance than TCVAE on PeMSD7(M) with $h \in \{96,180,720\}$. This may be attributed to the fact that long time series facilitates a more accurate learned structure but bring some challenges for learning distributional drifts. However, when the total number of time steps is extremely large, e.g., on Solar-Energy, AGCRN and StemGNN run out of memory (as marked in \textit{OOM} in Table \ref{table 3}) with the prediction step $h = 720$. In contrast, TCVAE functions well on Solar-Energy, further indicating the high scalability of TCVAE, relative to GNN-based models; 
(3) when $h \in \{180,720\}$, the advantage of AdaRNN in characterizing and matching temporal distribution information decreases compared with Nstatformer and our TCVAE, validating an RNN-type backbone is difficult to model the long sequences and easily affected by the cumulative errors; 
(4) on the PeMSD7(M) and Solar-Energy datasets, Informer achieves appreciable performance for long-range forecasting second only to our TCVAE. We attribute this close performance to the fact that the above datasets exhibit relatively weak non-stationary properties, which is consistent with the ADF test statistic in Table \ref{table 1}.

\subsubsection{Ablation Study}
To better understand the effectiveness of different components in TCVAE, we perform additional experiments on Electricity and PeMSD7(M) with ablation consideration by removing different components from TCVAE. In the experiments, we select a setting of $w = 24$ and $h = 24$. Table \ref{table 4} summarizes the results and shows that all the components are integral. 
Specifically, \textbf{w/o THA} is a variant where the Hawkes process is removed from TCVAE, and its temporal factor representation focuses on global-location rather than local-location contribution. Comparing TCVAE and \textbf{w/o THA}, we observe that temporal Hawkes attention significantly improves temporal attention (+3.74$\%$, +12.91$\%$, +3.44$\%$ corresponding to MAE, RMSE and MAPE respectively), verifying our design of using the Hawkes process to model time series as temporal point processes. A possible reason is that THA better captures the excitation induced by influential events while learning the distributed scores of every time step, which could also be observed from the visualization in Fig. \ref{figure 7}. 
The superiority of TCVAE over variant \textbf{w/o GAM} verifies the effect of temporal factors in guiding the generation of information flow. Due to the adaptive control of GAM, the valuable information generated by multiple heads is identified and the noise parts are greatly silenced. 
In addition, TCVAE shows significant improvement (+5.14$\%$, +13.74$\%$, +6.23$\%$) over variant \textbf{w/o FDA} using a standard CVAE and removing CCNF $\Phi$, indicating that breaking the specific distribution form and using temporal factors as distribution guides are of great significance. 
More finely-grained, \textbf{w/o CCNF} excludes the conditional continuous normalizing flow of the FDA module, thereby the TCVAE degenerates to a temporal factor-guided CVAE which directly matches the prior and posterior distributions with $\mathbb{KL}$ divergence. Comparing TCVAE and \textbf{w/o CCNF}, an important conclusion is that the flexible posterior approximated by CCNF surpasses the Gaussian posterior on both datasets. We argue that the intrinsic cause is that the simple Gaussian posterior can not estimate the true posterior well, using CCNF can help learn a more flexible density. 
And the improvements of TCVAE over variants \textbf{w/o Backcasting} and \textbf{w/o $\mathbb{KL}$} prove that both temporal conditions and backcasting designs can introduce additional information and improve sequence representation capacity. 
Back to Table \ref{table 4}, the MAE, RMSE and MAPE of TCVAE are significantly improved (+11.41$\%$, +16.00$\%$, +3.12$\%$) over \textbf{w/o $\mathbb{KL}$}, showing that $\mathbb{KL}$ loss helps emphasize the constraints of temporal factors in the latent space during the training. 
We note that removing backcasting hardly leads to significant degradation in performance, probably because the backcasting (reconstruction) loss has the same or duplicated effect as the forecasting loss.

\begin{table}[!t]\footnotesize
\setlength{\tabcolsep}{4.5pt}
 \centering
 \newcommand{\tabincell}[2]{\begin{tabular}{@{}#1@{}}#2\end{tabular}}
 \caption{Results of Ablation Study on Electricity and PeMSD7(M)}
 \label{table 4}  
 \begin{tabular}{ccccccc}
  \toprule 
  Datasets & \multicolumn{3}{c}{Electricity} & \multicolumn{3}{c}{PeMSD7(M)} \\ 
  \cmidrule(r){2-4} \cmidrule(r){5-7}
  Metrics & MAE & RMSE & MAPE & MAE & RMSE & MAPE \\
  \midrule 
  w/o THA & 0.065 & 0.093 & 0.294 & 0.076 & 0.116 & 0.217 \\ 
  w/o GAM & 0.064 & 0.092 & 0.287 & 0.077 & 0.115 & 0.212 \\ 
  w/o FDA & 0.067 & 0.095 & 0.309 & 0.076 & 0.116 & 0.219  \\ 
  w/o CCNF & 0.064 & 0.092 & 0.289 & 0.076 & 0.112 & 0.213 \\ 
  w/o $\mathbb{KL}$ &  0.065 & 0.093 & 0.287 & 0.090 & 0.124 & 0.221 \\ 
  w/o Backcasting & 0.062 & 0.089 & 0.288 & 0.075 & 0.115 & 0.217 \\ 
  \textbf{TCVAE} & 0.061 & 0.073 & 0.286 & 0.075 & 0.111 & 0.208 \\ 
  \bottomrule 
 \end{tabular}
\end{table}


\begin{figure}[!t]
\centering
\subfigure[Effect of window size $w$]{
    \label{w}
    \includegraphics[width=4cm,height=3cm]{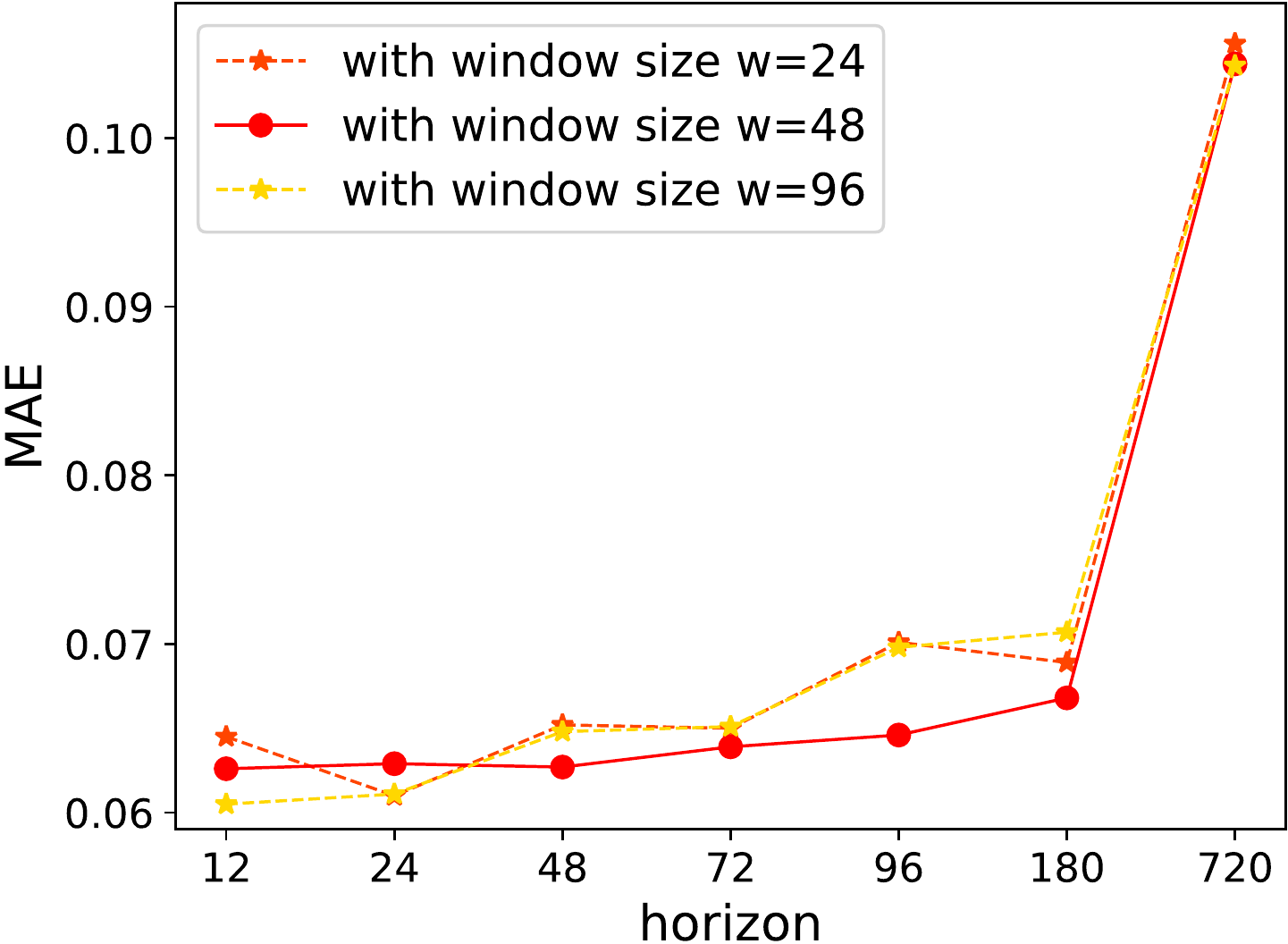}
    \includegraphics[width=4cm,height=3cm]{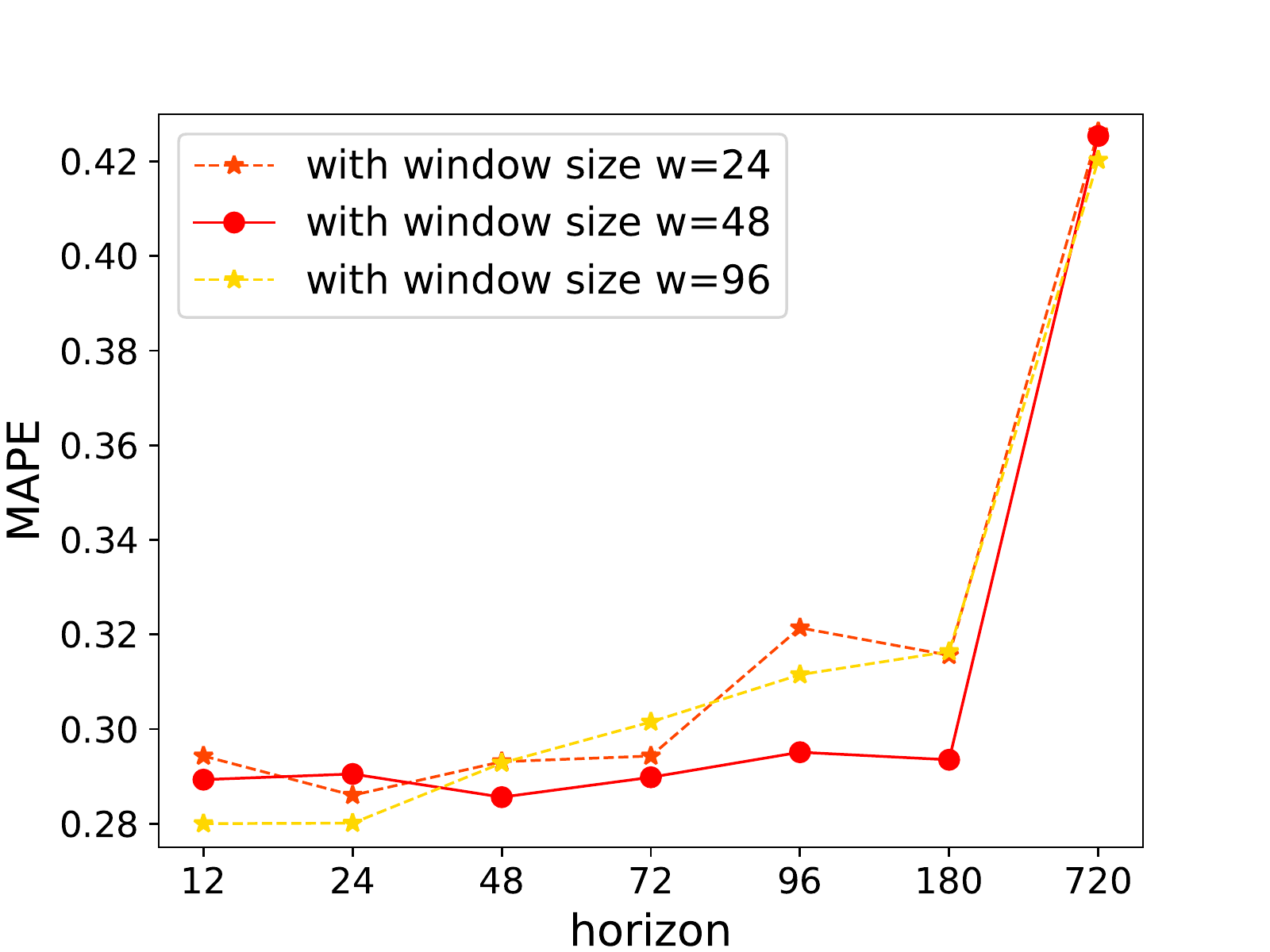}
    }
\subfigure[Effect of trade-off parameter $\lambda$]{
    \label{lambda}
    \includegraphics[width=4cm,height=3cm]{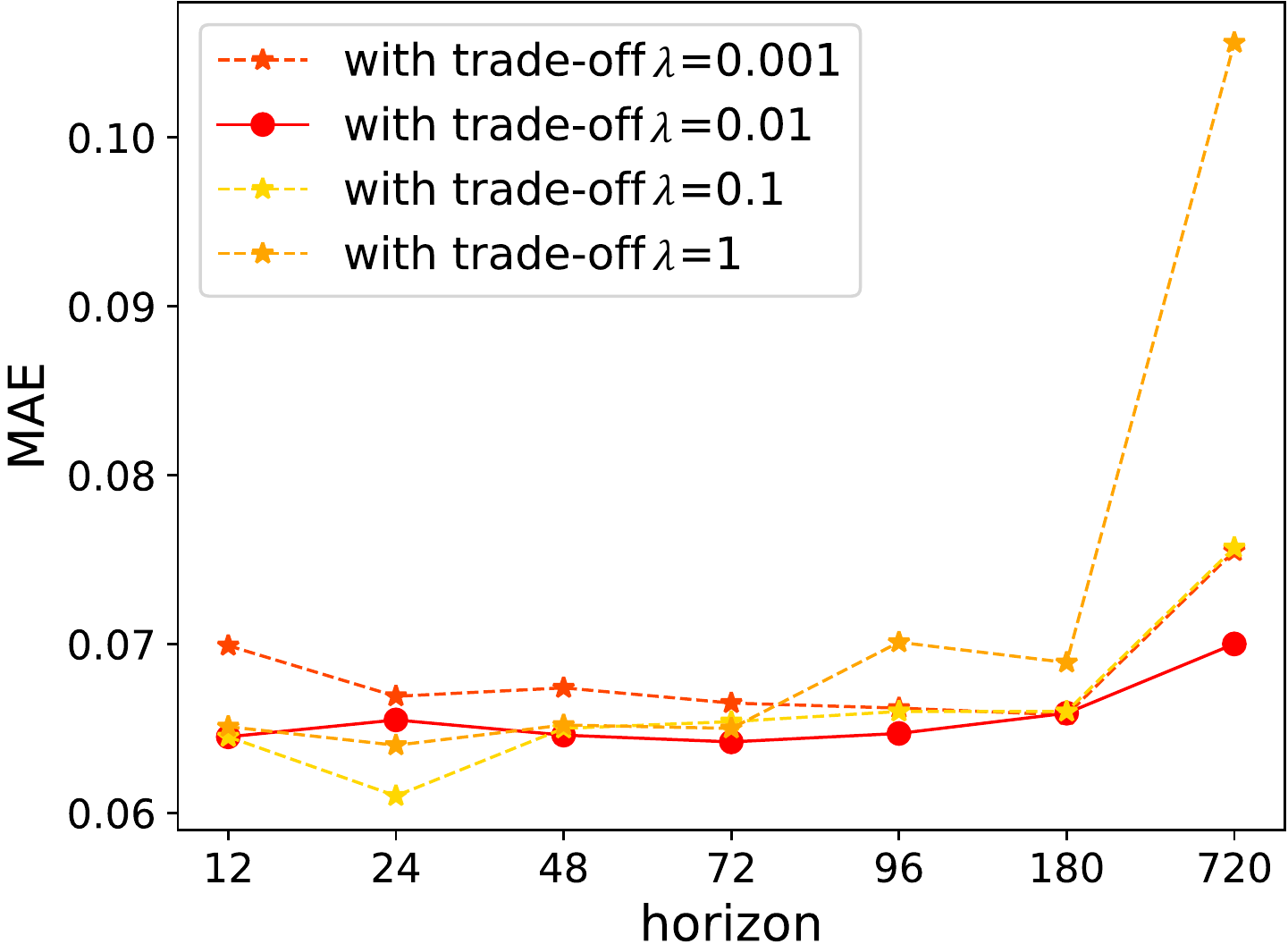}l
    \includegraphics[width=4cm,height=3cm]{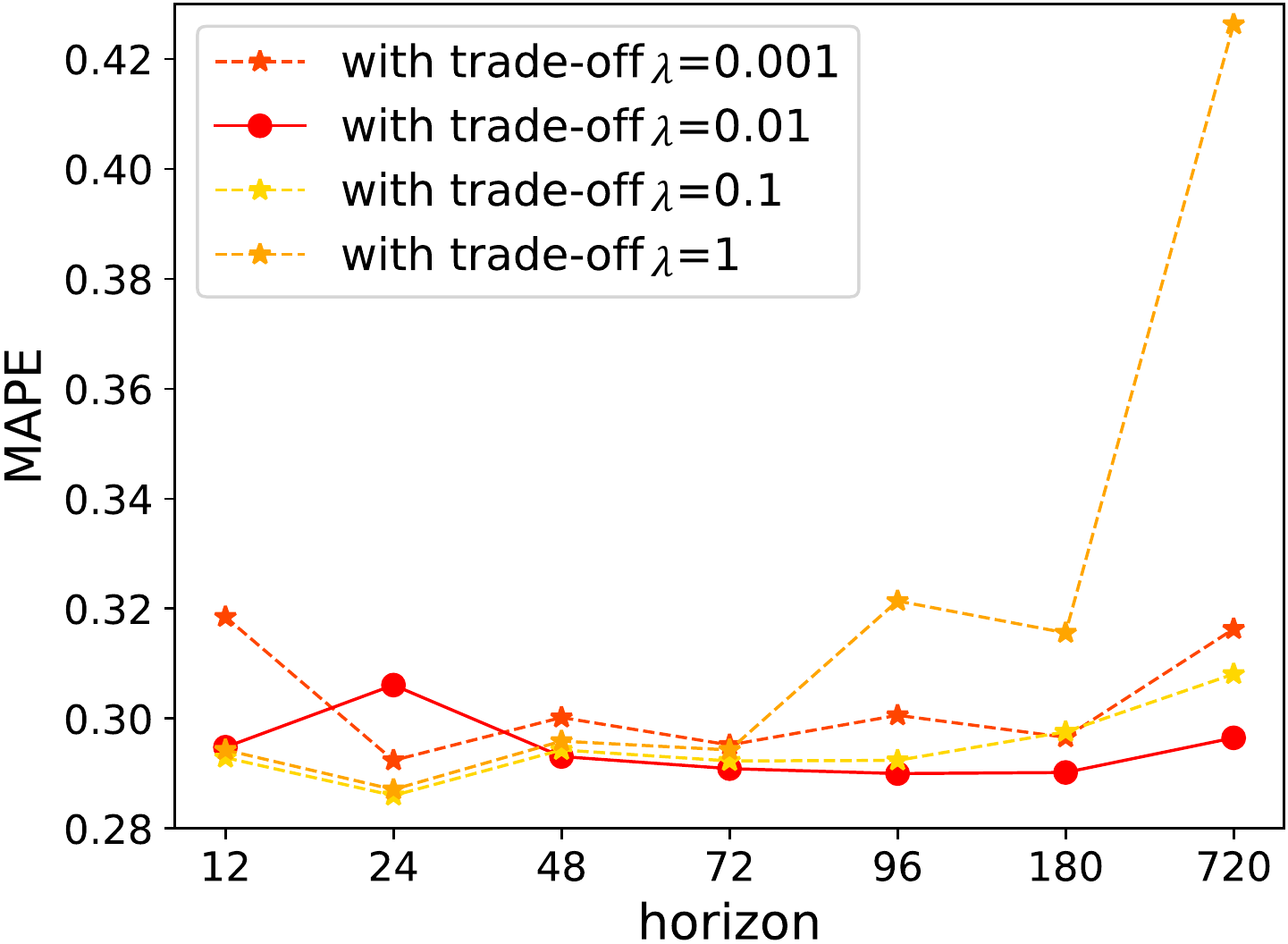}
    }
\subfigure[Effect of number of heads $n$]{
    \label{n}
    \includegraphics[width=4cm,height=3cm]{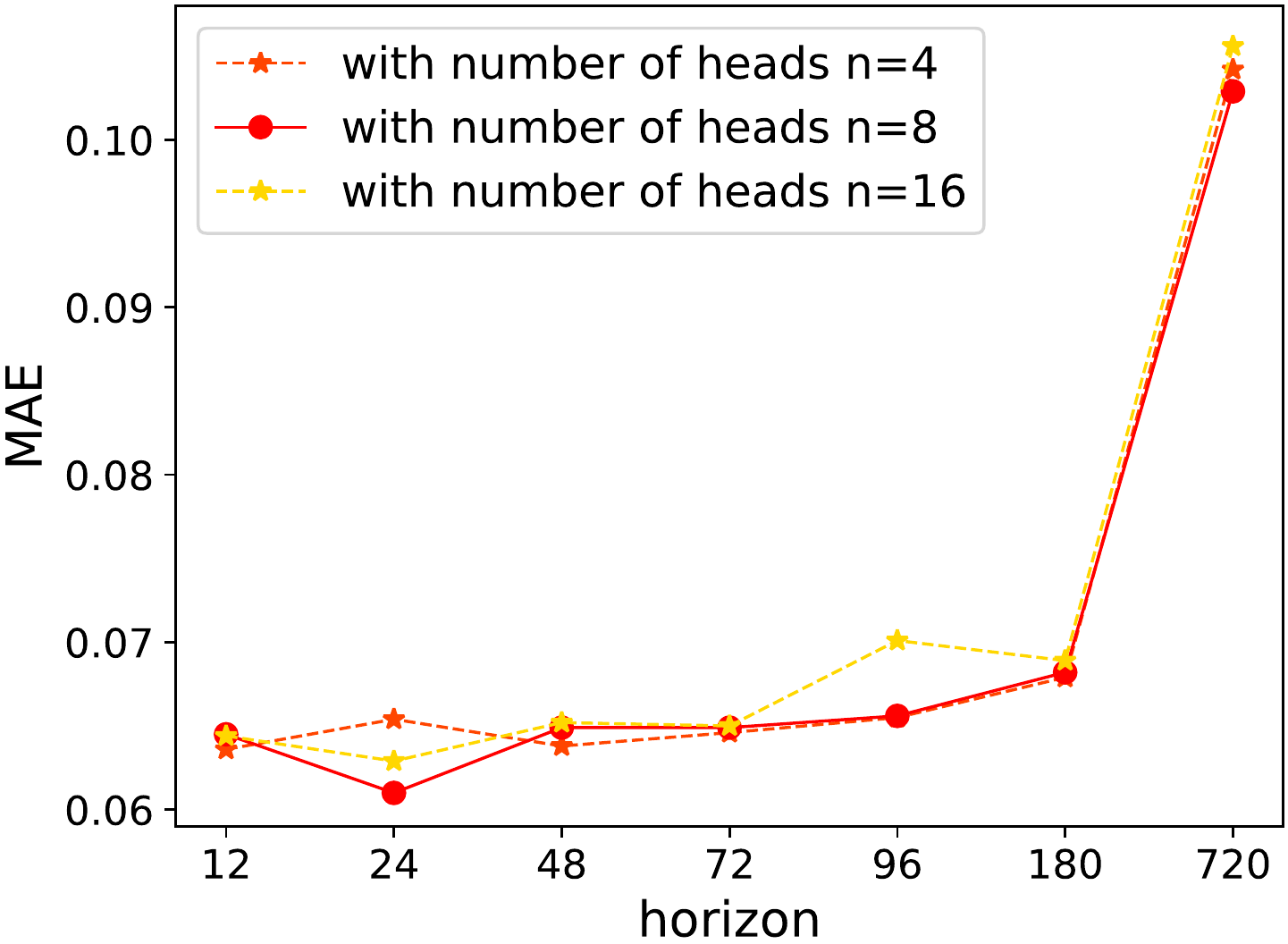}
    \includegraphics[width=4cm,height=3cm]{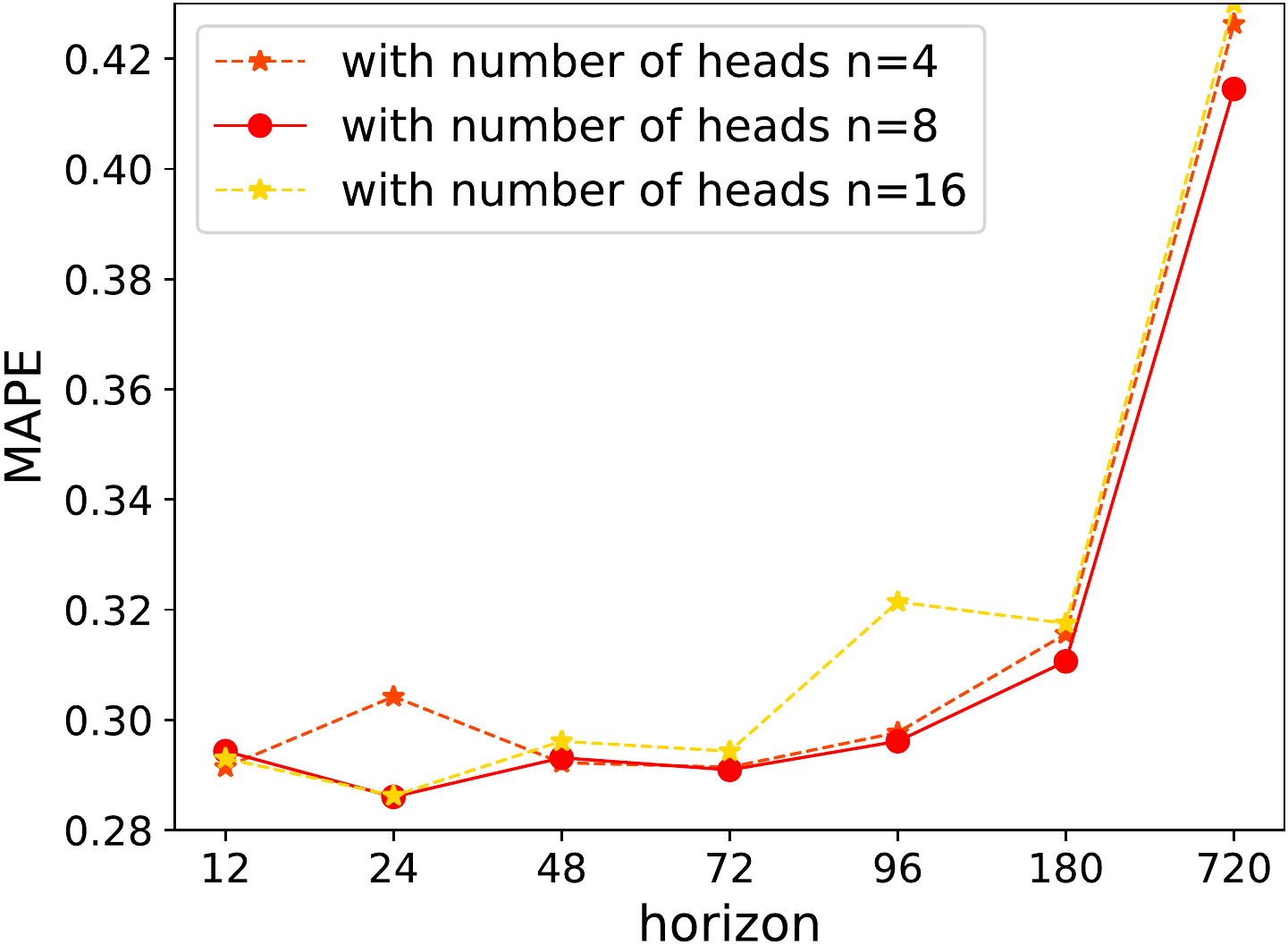}
    }
\subfigure[Effect of representation dimension $d$]{
    \label{d}
    \includegraphics[width=4cm,height=3cm]{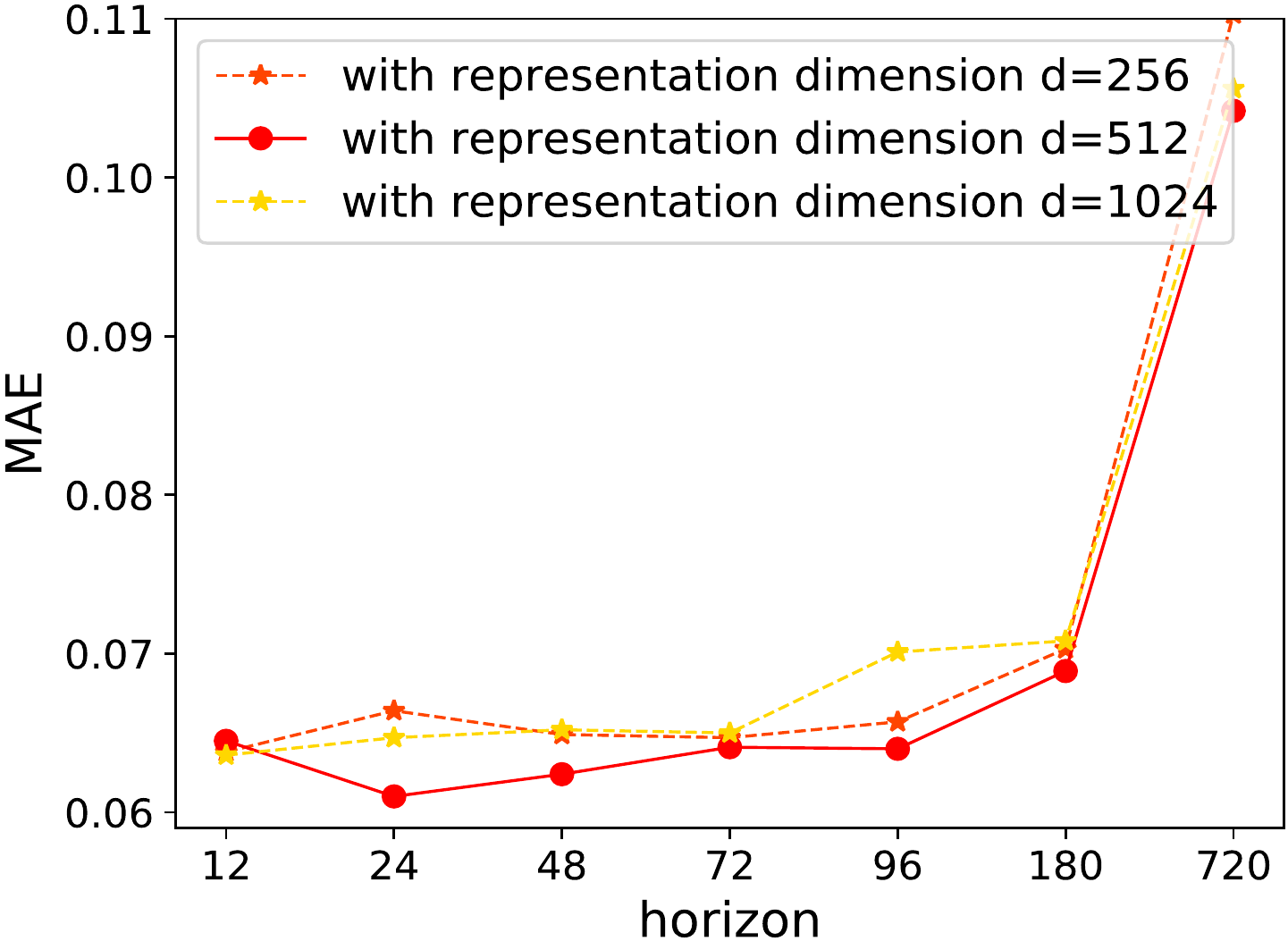}
    \includegraphics[width=4cm,height=3cm]{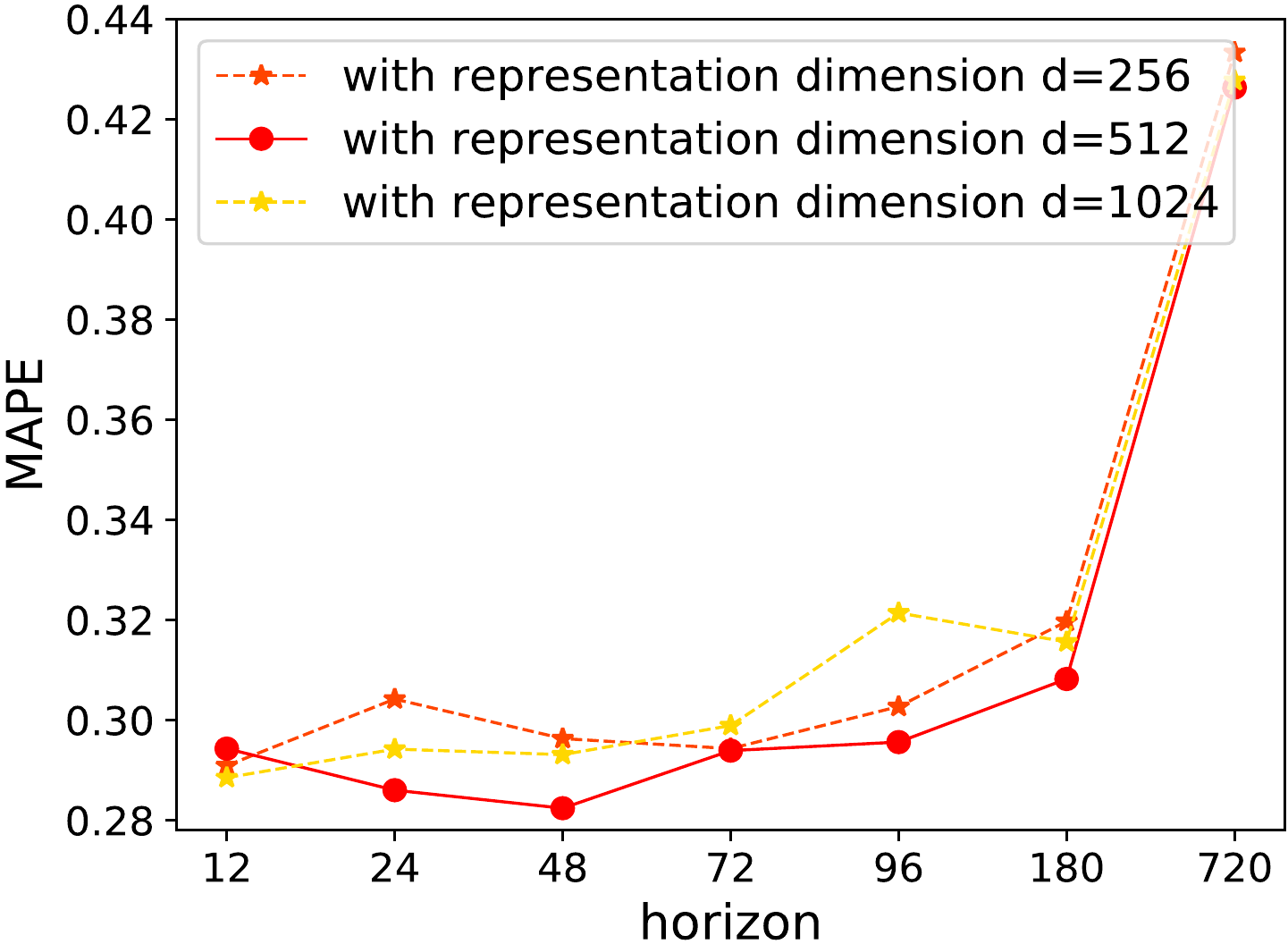}
    }
\caption{Parameters sensitivity analysis of TCVAE on MAE and MAPE.}
\label{figure 3}
\end{figure}

\subsubsection{Parameter Sensitivity}
We investigate the sensitivity of TCVAE in terms of the time window size $w \in \{24,48,96\}$, the trade-off parameter $\lambda \in \{1,10^{-1},10^{-2},10^{-3}\}$, the number of heads $n \in \{4,8,16\}$, the representation dimension $d\in \{256,512,1024\}$ on Electricity. The latent variable dimension $k$ varies in a range of $\{256,512,1024\}$ on Electricity and METR-LA. Fig.~\ref{figure 3} (a) - (d) shows the MAE and MAPE on Electricity by changing one parameter while fixing others. From the results, we have observations as follows. 
\textbf{(1) Window size $w$}:~Window size impacts learning distributional drifts. When predicting a short horizon (such as $h \in \{12,24\}$), $w=96$ achieves the best performance due to the sufficient temporal patterns and dependency information.
When $h \textgreater 48$, $w=48$ achieves the best performance.
The interpretation from a training perspective is that a relatively small value of $w$ may not pass enough information to the latent space $\mathcal{Z}$, while a large value of $w$ brings redundant information leading to losing accuracy.
\textbf{(2) Trade-off parameter $\lambda$}:~The influence of weight $\lambda$ for the $\mathbb{KL}$ loss term (Equation (\ref{eqn26})) is analyzed. The setting $\lambda=0.1$ with $h \in \{12,24\}$ and $\lambda=0.01$ with other prediction lengths give the optimal results. The results confirm that a larger $\lambda$ imposes a stronger emphasis on VAE to find the underlying factors leading to drift; however, a large $\lambda$ (e.g., $\lambda=1$) may hurt the performance. 
\textbf{(3) Number of heads $n$}: The results report that 4-head attention achieves the best performance in predicting 12 and 48 timestamps. The forecasting quality drops off with too many heads $n=16$.
\textbf{(4) Representation dimension $d$}:~The representation dimension of inputs and temporal factors in TCVAE highly determines the parameter effectiveness in temporal factor guidance and the capability of the learned representations. Fig.~\ref{figure 3} (d) shows the impact of various representation dimensions on Electricity. When the representation dimension is set to 512, TCVAE achieves the best performance. An excessively small or large representation dimension will result in poor performance. 
\textbf{(5) Latent variable dimension $k$}:~As shown in Table \ref{table 5}, similar to the representation dimension, $k=512$ results in the best performance of TCVAE, confirming that a larger $k$ improves the representation capability of TCVAE but may easily lead to the overfitting problem.

\begin{table}[!t]\footnotesize
 \centering
 \newcommand{\tabincell}[2]{\begin{tabular}{@{}#1@{}}#2\end{tabular}}
 \caption{Effect of Different Latent Variable Dimensions}
 \label{table 5}  
 \begin{tabular}{ccccccc}
  \toprule 
  Datasets & \multicolumn{3}{c}{Electricity} & \multicolumn{3}{c}{METR-LA} \\ 
  \cmidrule(r){2-4} \cmidrule(r){5-7}
  $k$ & 256 & 512 & 1024 & 256 & 512 & 1024 \\
  \midrule 
  MAE & 0.064 & 0.061 & 0.064 & 0.183 & 0.104 & 0.127 \\ 
  RMSE & 0.091 & 0.073 & 0.090 & 0.229 & 0.156 & 0.213 \\ 
  MAPE & 0.294 & 0.286 & 0.291 & 0.792 & 0.467 & 0.751 \\ 
  \bottomrule 
 \end{tabular}
\end{table}

\subsubsection{Case Study}
We deliver some case examples and analyze the pattern learned by several key components, i.e., FDA, GAM and THA of TCVAE, respectively. Traffic and pandemic data are highly dynamic and volatile, providing excellent scenarios for case studies.

\begin{figure}[!t]
\centering
\subfigure[At 8:00-10:00 A.M. (the 96$th$ window)]{
    \label{figure 4a}
    \includegraphics[width=1.0\linewidth]{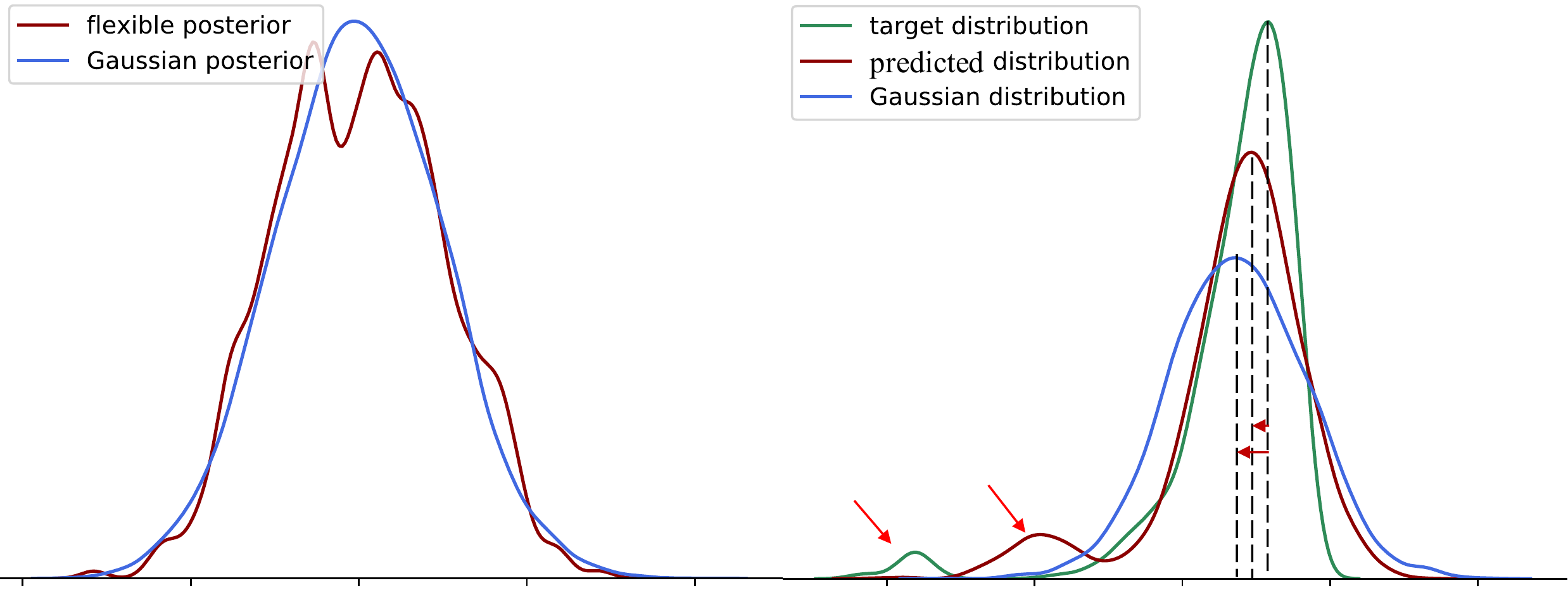}
    }
\subfigure[At 3:00-5:00 P.M. (the 192$nd$ window)]{
    \label{figure 4b}
    \includegraphics[width=1.0\linewidth]{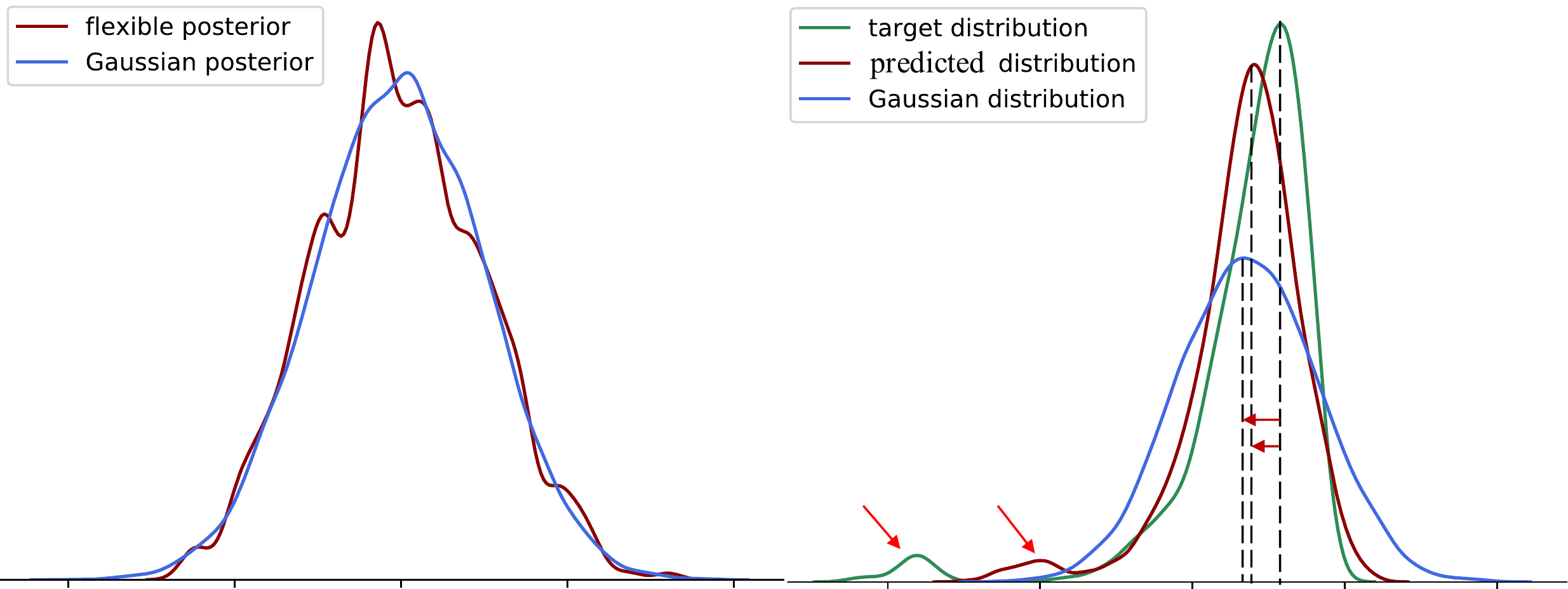}
    }
\subfigure[At 10:00-12:00 P.M. (the 288$th$ window)]{
    \label{figure 4c}
    \includegraphics[width=1.0\linewidth]{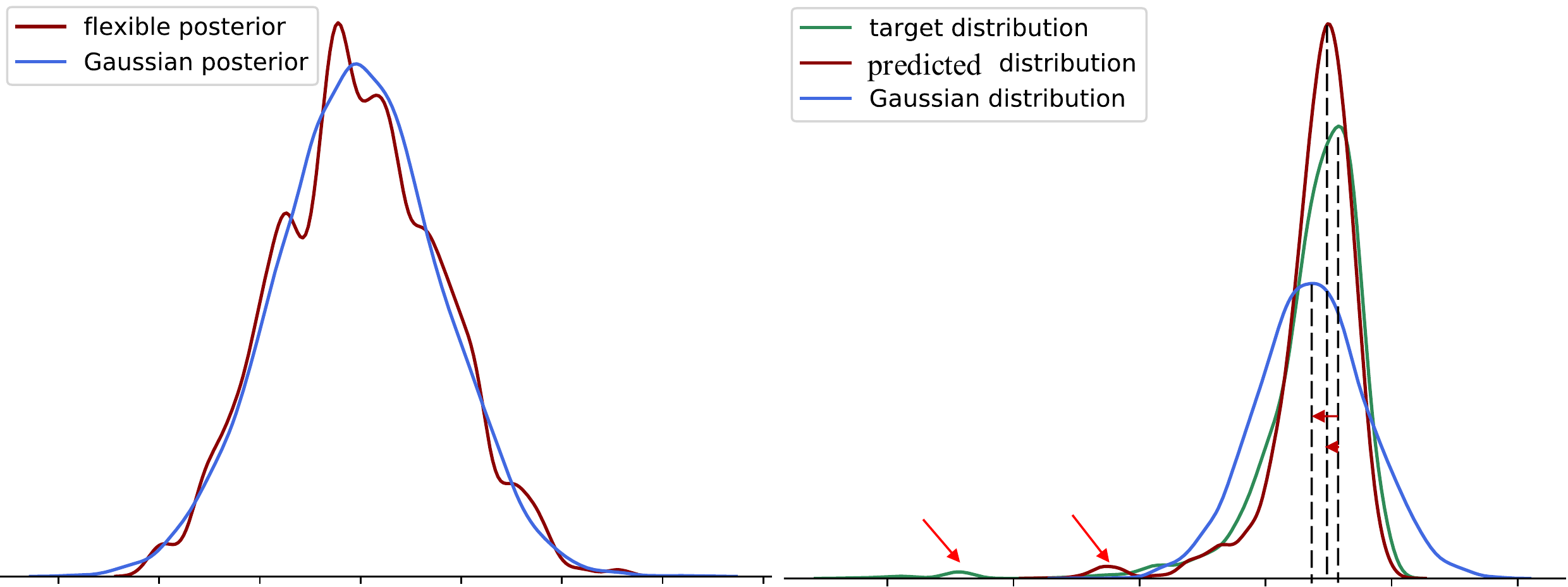}
    }
\caption{Distribution of different time windows, METR-LA. The left part is a comparison of flexible and Gaussian posterior in FDA. Comparisons of the target distribution, predicted distribution, and Gaussian distribution are shown in the right part.}
\label{figure 4}
\end{figure}

\textbf{Flexible Distribution Approximation}~To study the validity of our proposed FDA insightfully, we select three time windows 8:00-10:00 A.M. (the 96$th$ window), 3:00-5:00 P.M. (the 192$nd$ window) and 10:00-12:00 P.M. (the 288$th$ window) at a 7-hour interval. We fit the flexible posterior of FDA and a Gaussian posterior in the left part of Fig.~\ref{figure 4}. The target distribution, predicted distribution, and Gaussian distribution are compared in the right part. We can see that more details are reserved by the flexible posterior. A specific distribution such as a univariate Gaussian cannot approximate a target distribution (e.g. a multivariate distribution) accurately (red arrows in Fig.~\ref{figure 4}). The visualization is consistent with the study~\cite{DBLP:conf/aaai/WuNCZSCLZCD21}: the distribution of time series data that changes over time can be better approximated by a dynamic mixture distribution. Although there is a mean left shift (brown arrows in Fig.~\ref{figure 4}) between the target distribution and the predicted distribution, the deviation is reduced via learning a flexible distributional representation adaptive to temporal factors.

\begin{figure}[!t]
\centering
\includegraphics[width=1\linewidth]{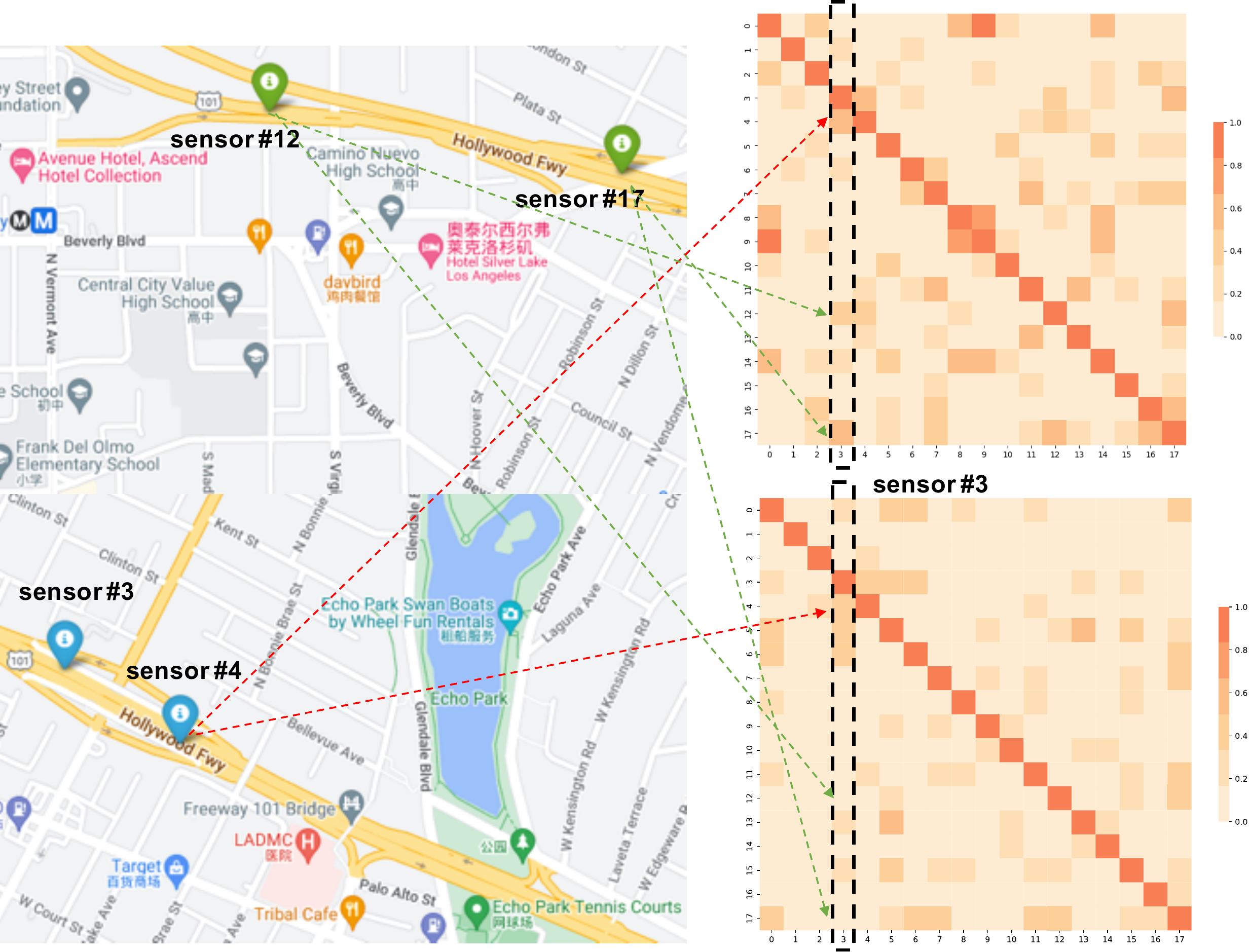}
\caption{Visualization of correlation matrix at 8:00-10:00 A.M. (the 96$th$ window) and 10:00-12:00 P.M. (the 288$th$ window) on 9/1/2018, METR-LA.}
\label{figure 5}
\end{figure}

In addition, we choose four sensors from METR-LA and show their corresponding locations on Google Map (left part in Fig.~\ref{figure 5}). In this experiment, we introduce additional intermediate variables $\bm{{\rm Z}}^{\#}=\mathcal{M}(\bm{{\rm Z}}^{\star})$ which can reflect the patterns learned by $\bm{{\rm Z}}^{\star}$ and $\bm{{\rm X}}^{\#}=\mathcal{M}(\tilde{\bm{{\rm X}}})$, where $\mathcal{M}(\bm{{\rm Z}}^{\star}):\mathbb{R}^k \rightarrow \mathbb{R}^{d_x}$ and $\mathcal{M}(\tilde{\bm{{\rm X}}}):\mathbb{R}^d \rightarrow \mathbb{R}^{d_x}$ are two feed-forward mappers before decoder $\mathcal{T}_d$ and $d_x=207$ denotes the number of sensors. 
Since each latent column vector in $\bm{{\rm Z}}^{\#}$ always corresponds to a unique sensor in raw multivariate time series, the correlation matrix of different sensors is computed by $\bm{{\rm Z}}^{\#}$ on two-time windows 8:00-10:00 A.M. and 10:00-12:00 P.M. (the right part in Fig.~\ref{figure 5}). The $i$th column in the matrix embodies the correlation strength between sensor $\#i$ and other sensors in the real-world environment. 
We can observe that the correlation between sensor $\#$3 and sensor $\#$4 is high and hardly varies, while the low correlation between sensor $\#$3 and sensor $\#$12, $\#$17 varies dramatically over time. The results suggest that some sensors are always closely related to each other because of the fixed spatial relationship, however, other sensors are closely related in one-time slice but `leave apart’ relations in other time slices. This dynamic is reasonable, since sensor $\#$3 and sensor $\#$12, $\#$17 are located on different main roads surrounded by different environments (schools, parks, etc.). Therefore, the module FDA can capture both invariance and dynamics of time series in the process of distributional drift adaptation.

\begin{figure}[!t]
\centering
\subfigure[On COVID-19]{
    \label{figure 6a}
    \includegraphics[width=1.0\linewidth]{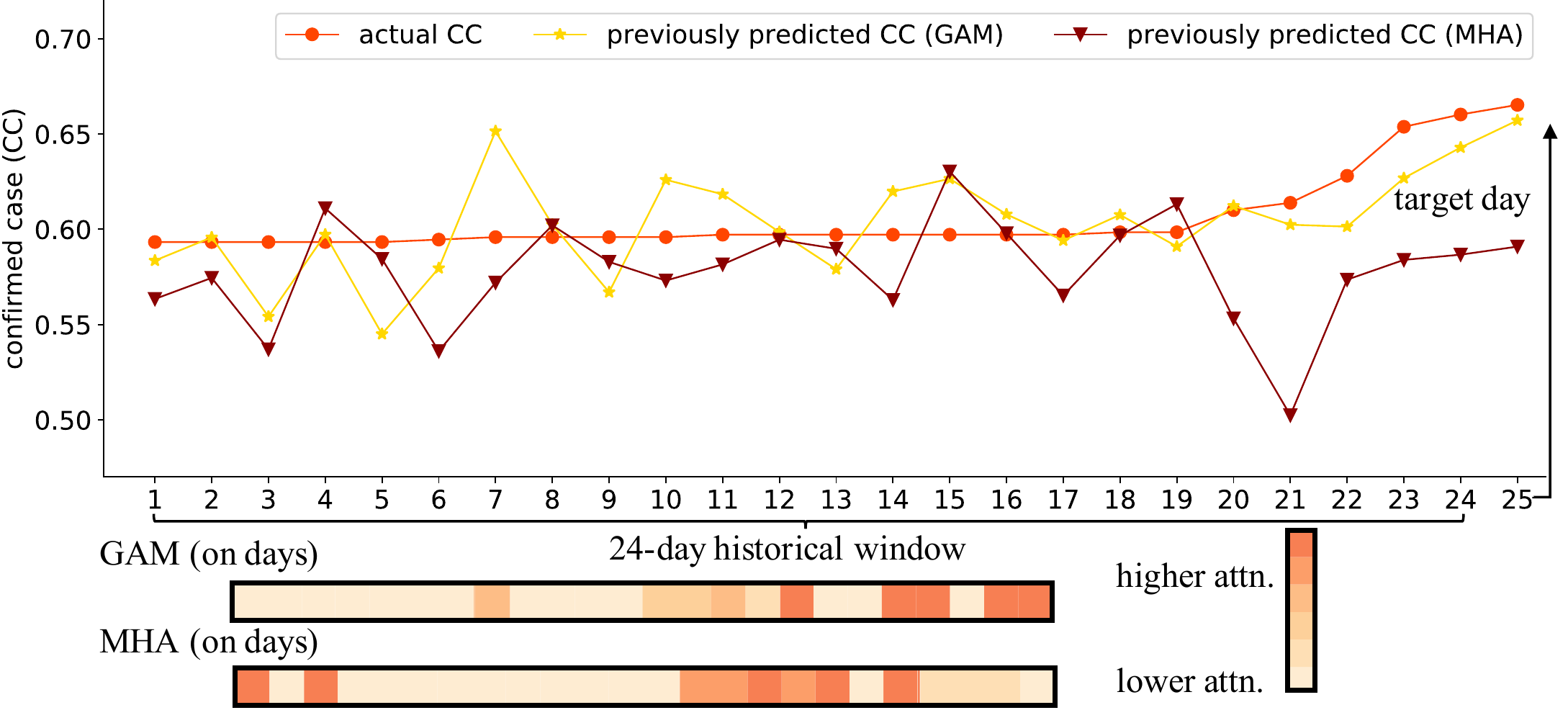}
    }
\subfigure[On Traffic]{
    \label{figure 6b}
    \includegraphics[width=1.0\linewidth]{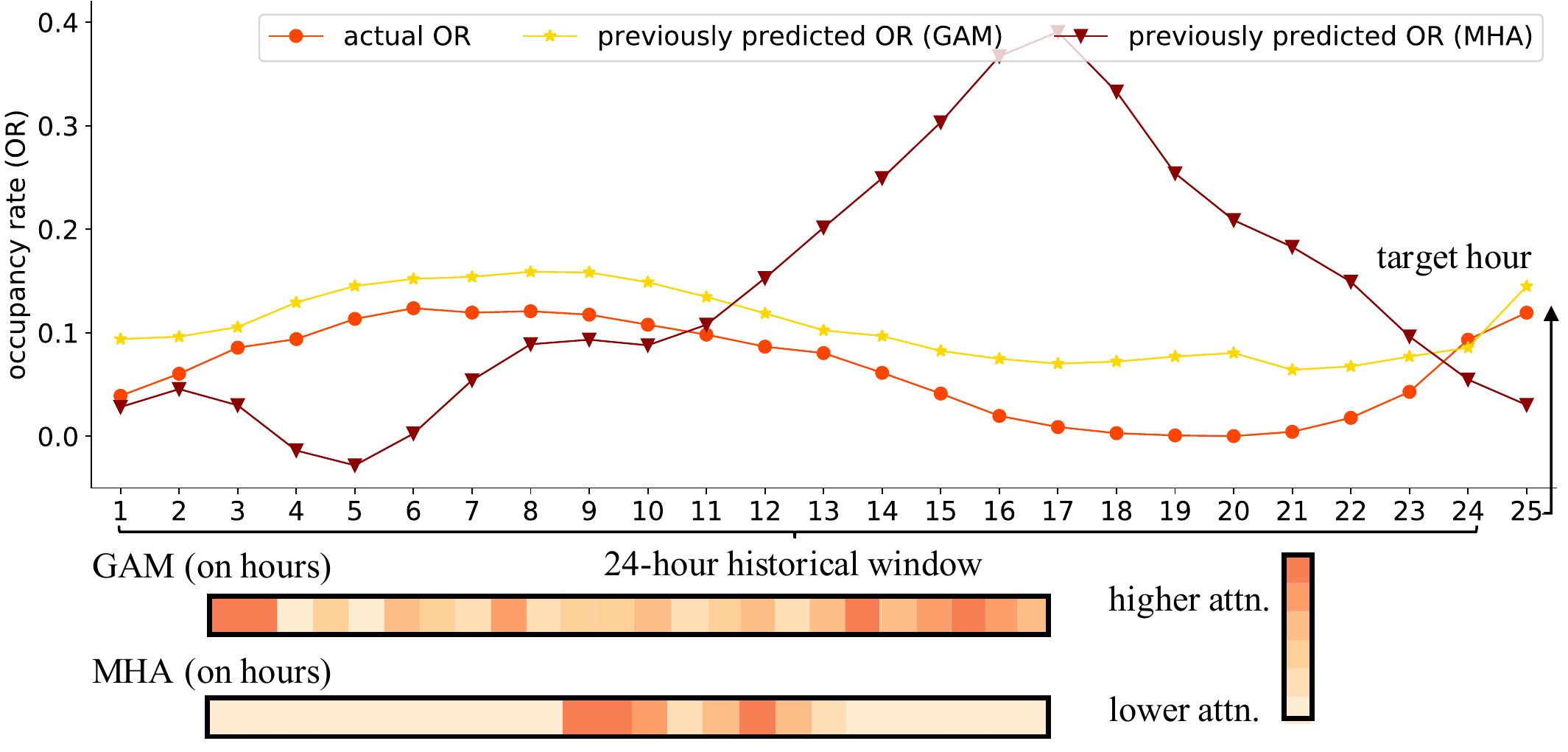}
    }
\caption{Day/hour-level attention visualization and prediction using TCVAE with GAM and MHA methods, where darker colors indicate higher weights.}
\label{figure 6}
\end{figure}

\textbf{Gated Attention Mechanism}~Next, we qualitatively analyze the time-guided component of TCVAE by comparing GAM and traditional multi-head attention (MHA) over a 24-day lookback for the number of Confirmed Cases (CC) in Jiangsu province in China from the COVID-19 test set in August 2021. We visualize day-level attention throughout the historical window and analyze the corresponding predicted CC for the 25$th$ day. For comparison, we also show actual and previously predicted CC across all days. The results are shown in Fig. \ref{figure 6a}. We see that TCVAE using MHA predicts the 25$th$ day CC with a relative error of 11.18$\%$ from the actual value, whereas TCVAE using GAM predicts that closer to the actual value (1.23$\%$). Despite the varying trends throughout the historical window, GAM focuses on the sharp upscent trend (e.g., the 21$st$ day, the 23$rd$ day) towards the end of the window, whereas MHA learns weighted scores and captures a gentle upscent trend (e.g., the 1$st$ day, the 15$th$ day). The results of GAM are consistent with the theory~\cite{DBLP:conf/aaai/SawhneyAWDS21}: the impact of long-distance events gradually decays, potentially because GAM effectively adapts the information flow of multiple heads to temporal factors to decay the impact of long-distance time points enabling TCVAE to focus on more recent salient time points, which better reflects the temporal state of event development. 
In Fig. \ref{figure 6b}, we further look back 24-hour Occupancy Rate (OR) of a randomly selected sensor from the test set of Traffic with obvious periodical patterns~\cite{DBLP:conf/iclr/0010ZYCF0L22}. TCVAE using GAM predicts the 25$th$ hour OR closer to the actual value than using MHA. In terms of hour-level attention, GAM accurately assigns larger attention scores to hours with an upscent trend (e.g., the 1$st$ hour, the 22$nd$ hour), whereas MHA assigns larger weights to hours with a descent trend (e.g., the 11$th$ hour, the 12$th$ hour). On periodic data, we note that GAM hardly reflects the regularity that the impact of long-distance events gradually decays, which is reasonable because periodic data generally exhibits repeated patterns where distant time points may receive high attention scores due to similar patterns.

\begin{figure}[!t]
\centering
\includegraphics[width=1\linewidth]{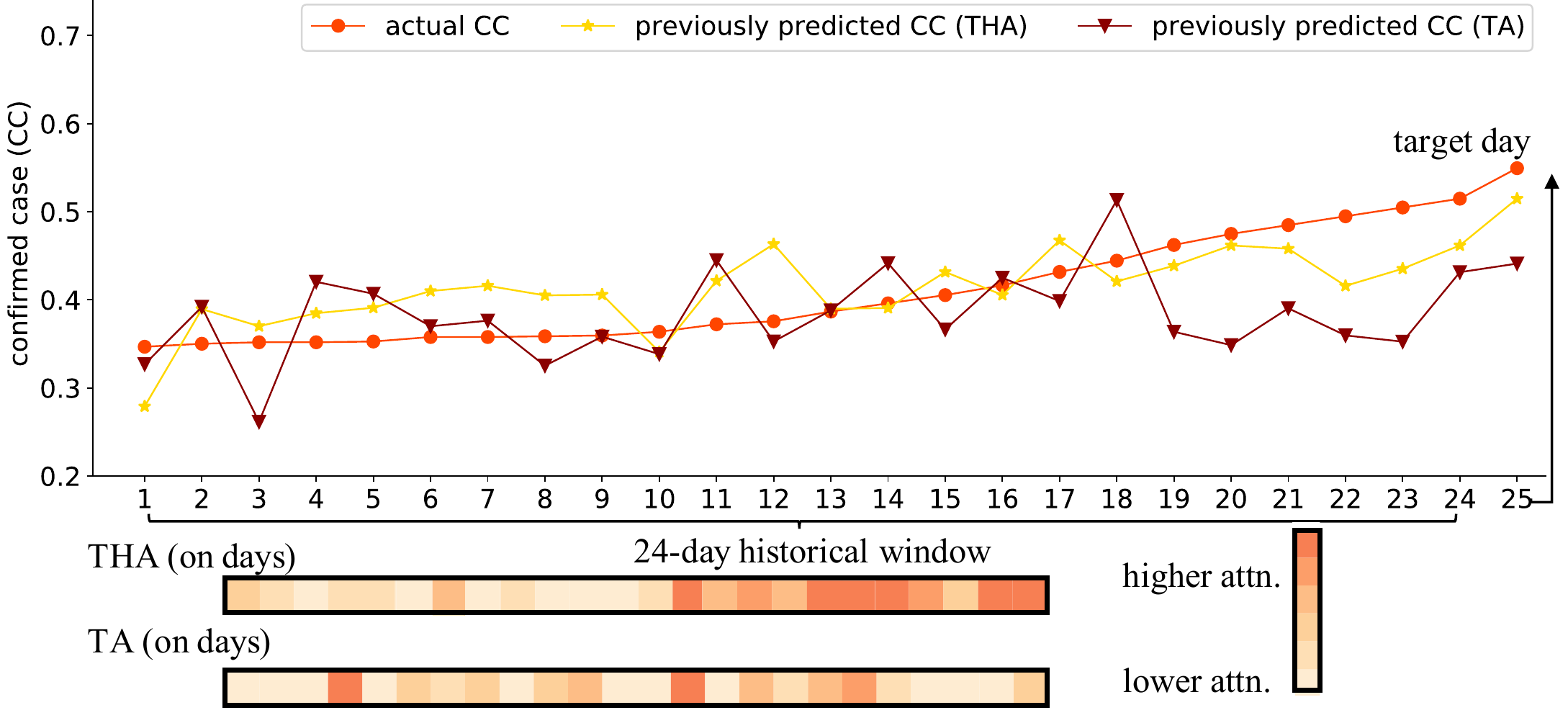}
\caption{Day-level attention visualization and prediction using TCVAE with THA and TA methods on COVID-19, where darker colors indicate higher weights.}
\label{figure 7}
\end{figure}

\textbf{Temporal Hawkes Attention}~To understand what is of importance the temporal Hawkes attention learns, we compare THA and temporal attention (TA) denoted as $\vartheta(\cdot)$ in Section IV.C over a 24-day lookback for the number of CC in Beijing from the COVID-19 training set in June 2020. Accordingly, the visualization of day-level attention and the predicted curves using THA and TA are given in Fig. \ref{figure 7}. TCVAE using TA predicts the 25$th$ day CC with a relative error of 19.72$\%$ from the actual value, whereas using THA, predicts the cases closer to the actual value (6.30$\%$). THA receives more attention scores concentrated in the later part of the window and captures the fast-rising trend, whereas TA learns more scattered weighted scores and captures a gentle uptrend, probably because THA better captures the excitation induced by influential events, i.e., a clustered epidemic outbreak starting in the middle of the window.

\subsubsection{Computational Complexity}
With the multivariate setting and all the methods’ current finest implementation, we perform a statistical comparison of the time costs and parameter volumes on Traffic in Table \ref{table 6}. We investigate the computational complexity of all the comparative baselines, TCVAE and its variant \textbf{w/o CCNF}. From the experimental results, we can observe that: (1) TCVAE requires more parameters than the Transformer- and GNN-based methods, aiming to learn temporal dependencies to address the distribution drift in MTSs; (2) regarding training/test time, TCVAE runs faster than the GNN-based methods but slower than the Transformer-based methods; (3) StemGNN and AGCRN have higher time costs than the SOTA Transformer-based methods despite achieving competitive forecasting performance; (4) compared to the RNN- and CNN-based methods (except for DSANet), the Transformer- and GNN-based methods require much more parameters and training/test time costs but achieve much better multi-step forecasting performance; (5) the variant of TCVAE, \textbf{w/o CCNF} achieves both high efficiency and desirable forecasting performance compared with the baselines.

Although TCVAE needs relatively large parameters and time costs, it has addressed the challenging distribution drift issue instead of the general MTS forecasting issue and achieved a significant performance improvement over the baselines, see Table \ref{table 2}. In addition, since TCVAE shows a considerable improvement over \textbf{w/o CCNF}, TCVAE still maintains the time-consuming CCNF module to achieve higher performance and transform the Gaussian distribution to a complex and form-free distribution, see Table \ref{table 4} and Fig. \ref{figure 4}. 
In summary, considering the significant performance improvement of TCVAE and its effectiveness in tackling the challenging distribution drift in MTSs, the computation and time costs of TCVAE are moderate and acceptable.

\begin{table*}[!t]\footnotesize
\setlength{\tabcolsep}{2pt}
 \centering
 \caption{Statistics of Model Complexity in Terms of Training/Test Time Costs and Model Parameters, Where Sec Denotes Seconds}
 \label{table 6}
 \newcommand{\tabincell}[2]{\begin{tabular}{@{}#1@{}}#2\end{tabular}}
 \begin{tabular}{c|ccccccc|cccc|cc}
  \toprule
  \toprule
  Models & LSTNet & AdaRNN & DSANet & STNorm & StemGNN & AGCRN & TS2VEC & Informer & Autoformer & FEDformer & Nstatformer & TCVAE & w/o CCNF \\
  \midrule 
  Parameters (M) & 0.55 & 0.44 & 39.75 & 0.14 & 7.08 & 0.76 & 0.70 & 14.83 & 15.43 & 17.00 & 14.50 & 18.34 & 15.21 \\
  \midrule 
  \makecell{Training Time \\ (Sec/Epoch)} & 9.54 & 15.52 & 90.99 & 9.26 & 67.16 & 71.24 & 8.77 & 26.83 & 32.88 & 35.45 & 34.43 & 65.4 & 37.29 \\
  \midrule 
  Test Time (Sec) & 4.94 & 13.60 & 25.60 & 1.06 & 24.31 & 17.89 & 4.06 & 3.96 & 11.21 & 8.78 & 10.39 & 16.04 & 8.65 \\
  \bottomrule 
  \bottomrule 
 \end{tabular}
\end{table*}

\section{Conclusion and Future Work}
In this work, a novel variational autoencoder architecture TCVAE for distributional drift adaptation is introduced to model the dynamic dependencies between historical observations and future data varying with time in MTS. Specifically, we design a temporal Hawkes attention mechanism to represent temporal factors that estimate the temporal Gaussian and a gated attention mechanism to dynamically adapt the network structure of the Transformer-based encoder and decoder. We propose to take advantage of transforming the temporal Gaussian into a flexible distribution that breaks the limitations of distributional form for inferring the temporal conditional distribution. In a variety of MTS forecasting applications, TCVAE based on more robust representation regularly outperforms previous techniques.

We will consider future work from two perspectives. First, TCVAE learns the distributional drift adaptation in estimating the conditional distribution of future data based on the assumption that the distributional drift frequencies of the training  and testing sets are the same. The cases of different distribution frequencies often exist in some scenarios. We will consider the dynamics to improve our model in the future. Second, we will explore its application in other real distribution-drift scenarios, such as research hot spots, stock trend forecasting, and product demand analysis.


\section*{Acknowledgments}
This work is supported by the Engineering Research Center of Integration and Application of Digital Learning Technology, Ministry of Education (1331001), National Natural Science Foundation of China (62272048), and National Key Research and Development Program of China (2019YFB1406302).


 

\bibliographystyle{IEEEtran}
\bibliography{TNNLS-2022-P-24163.R3}

\begin{thebibliography}{10}
\providecommand{\url}[1]{#1}
\csname url@samestyle\endcsname
\providecommand{\newblock}{\relax}
\providecommand{\bibinfo}[2]{#2}
\providecommand{\BIBentrySTDinterwordspacing}{\spaceskip=0pt\relax}
\providecommand{\BIBentryALTinterwordstretchfactor}{4}
\providecommand{\BIBentryALTinterwordspacing}{\spaceskip=\fontdimen2\font plus
\BIBentryALTinterwordstretchfactor\fontdimen3\font minus
  \fontdimen4\font\relax}
\providecommand{\BIBforeignlanguage}[2]{{%
\expandafter\ifx\csname l@#1\endcsname\relax
\typeout{** WARNING: IEEEtran.bst: No hyphenation pattern has been}%
\typeout{** loaded for the language `#1'. Using the pattern for}%
\typeout{** the default language instead.}%
\else
\language=\csname l@#1\endcsname
\fi
#2}}
\providecommand{\BIBdecl}{\relax}
\BIBdecl

\bibitem{DBLP:conf/nips/0001YL0020}
L.~Bai, L.~Yao, C.~Li, X.~Wang, and C.~Wang, ``Adaptive graph convolutional
  recurrent network for traffic forecasting,'' in \emph{Proc. NeurIPS}, 2020.

\bibitem{DBLP:conf/icde/CirsteaYGKP22}
R.~Cirstea, B.~Yang, C.~Guo, T.~Kieu, and S.~Pan, ``Towards spatio- temporal
  aware traffic time series forecasting,'' in \emph{Proc. {ICDE}}, 2022, pp.
  2900--2913.

\bibitem{DBLP:conf/kdd/ZhangAQ17}
L.~Zhang, C.~C. Aggarwal, and G.~Qi, ``Stock price prediction via discovering
  multi-frequency trading patterns,'' in \emph{Proc. {KDD}}, 2017, pp.
  2141--2149.

\bibitem{DBLP:journals/eswa/LeeK20}
S.~W. Lee and H.~Y. Kim, ``Stock market forecasting with super-high dimensional
  time-series data using convlstm, trend sampling, and specialized data
  augmentation,'' \emph{Expert Syst. Appl.}, vol. 161, p. 113704, 2020.

\bibitem{DBLP:journals/tnn/TranIKG19}
D.~T. Tran, A.~Iosifidis, J.~Kanniainen, and M.~Gabbouj, ``Temporal
  attention-augmented bilinear network for financial time-series data
  analysis,'' \emph{{IEEE} Trans. Neural Networks Learn. Syst.}, vol.~30,
  no.~5, pp. 1407--1418, 2019.

\bibitem{DBLP:conf/aaai/XuCZSNYLC020}
D.~Xu, W.~Cheng, B.~Zong, D.~Song, J.~Ni, W.~Yu, Y.~Liu, H.~Chen, and X.~Zhang,
  ``Tensorized {LSTM} with adaptive shared memory for learning trends in
  multivariate time series,'' in \emph{Proc. {AAAI}}, 2020, pp. 1395--1402.

\bibitem{DBLP:journals/air/TarusNM18}
J.~K. Tarus, Z.~Niu, and G.~Mustafa, ``Knowledge-based recommendation: a review
  of ontology-based recommender systems for e-learning,'' \emph{Artif. Intell.
  Rev.}, vol.~50, no.~1, pp. 21--48, 2018.

\bibitem{DBLP:journals/tnn/ZhangCSN22}
Q.~Zhang, L.~Cao, C.~Shi, and Z.~Niu, ``Neural time-aware sequential
  recommendation by jointly modeling preference dynamics and explicit feature
  couplings,'' \emph{{IEEE} Trans. Neural Networks Learn. Syst.}, vol.~33,
  no.~10, pp. 5125--5137, 2022.

\bibitem{DBLP:journals/pami/SpadonHBMRS22}
G.~Spadon, S.~Hong, B.~Brandoli, S.~Matwin, J.~F.~R. Jr., and J.~Sun, ``Pay
  attention to evolution: Time series forecasting with deep graph-evolution
  learning,'' \emph{{IEEE} Trans. Pattern Anal. Mach. Intell.}, vol.~44, no.~9,
  pp. 5368--5384, 2022.

\bibitem{DBLP:conf/nips/CaoWDZZHTXBTZ20}
D.~Cao, Y.~Wang, J.~Duan, C.~Zhang, X.~Zhu, C.~Huang, Y.~Tong, B.~Xu, J.~Bai,
  J.~Tong, and Q.~Zhang, ``Spectral temporal graph neural network for
  multivariate time-series forecasting,'' in \emph{Proc. NeurIPS}, 2020.

\bibitem{Cao-covid21}
L.~Cao, Q.~Liu, and W.~Hou, ``{COVID-19} modeling: {A} review,'' \emph{CoRR},
  vol. abs/2104.12556, 2021.

\bibitem{Vector1993}
M.~W. Watson, ``Vector autoregressions and cointegration,'' \emph{Working Paper
  Series, Macroeconomic Issues}, vol.~4, 1993.

\bibitem{SeegerSF16}
M.~W. Seeger, D.~Salinas, and V.~Flunkert, ``Bayesian intermittent demand
  forecasting for large inventories,'' in \emph{Proc. {NIPS}}, 2016, pp.
  4646--4654.

\bibitem{9380704}
X.~Chen and L.~Sun, ``Bayesian temporal factorization for multidimensional time
  series prediction,'' \emph{{IEEE} Trans. Pattern Anal. Mach. Intell.}, pp.
  1--1, 2021.

\bibitem{DBLP:conf/sigir/LaiCYL18}
G.~Lai, W.~Chang, Y.~Yang, and H.~Liu, ``Modeling long- and short-term temporal
  patterns with deep neural networks,'' in \emph{Proc. {SIGIR}}, 2018, pp.
  95--104.

\bibitem{DBLP:journals/ml/ShihSL19}
S.~Shih, F.~Sun, and H.~Lee, ``Temporal pattern attention for multivariate time
  series forecasting,'' \emph{Mach. Learn.}, vol. 108, no. 8-9, pp. 1421--1441,
  2019.

\bibitem{DBLP:journals/tnn/BandaraBH21}
K.~Bandara, C.~Bergmeir, and H.~Hewamalage, ``Lstm-msnet: Leveraging forecasts
  on sets of related time series with multiple seasonal patterns,''
  \emph{{IEEE} Trans. Neural Networks Learn. Syst.}, vol.~32, no.~4, pp.
  1586--1599, 2021.

\bibitem{9669023}
W.~Zheng and J.~Hu, ``Multivariate time series prediction based on temporal
  change information learning method,'' \emph{{IEEE} Trans. Neural Networks
  Learn. Syst.}, pp. 1--15, 2022.

\bibitem{10197239}
J.~Xu and L.~Cao, ``Copula variational lstm for high-dimensional cross-market
  multivariate dependence modeling,'' \emph{IEEE Trans. Neural Networks Learn.
  Syst.}, pp. 1--15, 2023.

\bibitem{DBLP:conf/nips/SenYD19}
R.~Sen, H.~Yu, and I.~S. Dhillon, ``Think globally, act locally: {A} deep
  neural network approach to high-dimensional time series forecasting,'' in
  \emph{Proc. NeurIPS}, 2019, pp. 4838--4847.

\bibitem{DBLP:conf/aaai/ZhouZPZLXZ21}
H.~Zhou, S.~Zhang, J.~Peng, S.~Zhang, J.~Li, H.~Xiong, and W.~Zhang,
  ``Informer: Beyond efficient transformer for long sequence time-series
  forecasting,'' in \emph{Proc. {AAAI}}, 2021, pp. 11\,106--11\,115.

\bibitem{DBLP:conf/iclr/LiuYLLLLD22}
S.~Liu, H.~Yu, C.~Liao, J.~Li, W.~Lin, A.~X. Liu, and S.~Dustdar, ``Pyraformer:
  Low-complexity pyramidal attention for long-range time series modeling and
  forecasting,'' in \emph{Proc. {ICLR}}, 2022.

\bibitem{DBLP:conf/nips/WuXWL21}
H.~Wu, J.~Xu, J.~Wang, and M.~Long, ``Autoformer: Decomposition transformers
  with auto-correlation for long-term series forecasting,'' in \emph{Proc.
  {NeurIPS}}, 2021, pp. 22\,419--22\,430.

\bibitem{DBLP:conf/icml/ZhouMWW0022}
T.~Zhou, Z.~Ma, Q.~Wen, X.~Wang, L.~Sun, and R.~Jin, ``Fedformer: Frequency
  enhanced decomposed transformer for long-term series forecasting,'' in
  \emph{Proc. {ICML}}, vol. 162, 2022, pp. 27\,268--27\,286.

\bibitem{DBLP:conf/iclr/ZhangY23}
Y.~Zhang and J.~Yan, ``Crossformer: Transformer utilizing cross-dimension
  dependency for multivariate time series forecasting,'' in \emph{Proc.
  {ICLR}}, 2023.

\bibitem{10308867}
S.~Zhu, J.~Zheng, and Q.~Ma, ``Mr-transformer: Multiresolution transformer for
  multivariate time series prediction,'' \emph{{IEEE} Trans. Neural Networks
  Learn. Syst.}, pp. 1--13, 2023.

\bibitem{cj_Cao14}
L.~Cao, ``Non-iidness learning in behavioral and social data,'' \emph{Comput.
  J.}, vol.~57, no.~9, pp. 1358--1370, 2014.

\bibitem{DBLP:conf/aaai/LiYLX022}
W.~Li, X.~Yang, W.~Liu, Y.~Xia, and J.~Bian, ``{DDG-DA:} data distribution
  generation for predictable concept drift adaptation,'' in \emph{Proc.
  {AAAI}}, 2022, pp. 4092--4100.

\bibitem{9783029}
W.~Zheng, P.~Zhao, G.~Chen, H.~Zhou, and Y.~Tian, ``A hybrid spiking neurons
  embedded lstm network for multivariate time series learning under
  concept-drift environment,'' \emph{{IEEE} Trans. Knowl. Data Eng.}, pp. 1--1,
  2022.

\bibitem{DBLP:conf/aaai/ZhouLLYZJ19}
Y.~Zhou, C.~Liang, N.~Li, C.~Yang, S.~Zhu, and R.~Jin, ``Robust online matching
  with user arrival distribution drift,'' in \emph{Proc. {AAAI}}, 2019, pp.
  459--466.

\bibitem{DBLP:journals/corr/abs-2204-05101}
C.~Liu, Y.~Li, X.~Zhao, C.~Peng, Z.~Lin, and J.~Shao, ``Concept drift
  adaptation for {CTR} prediction in online advertising systems,'' \emph{CoRR},
  vol. abs/2204.05101, 2022.

\bibitem{DBLP:journals/corr/abs-2205-14415}
Y.~Liu, H.~Wu, J.~Wang, and M.~Long, ``Non-stationary transformers: Rethinking
  the stationarity in time series forecasting,'' in \emph{Proc. NeurIPS}, 2022.

\bibitem{DBLP:journals/tnn/PassalisTKGI20}
N.~Passalis, A.~Tefas, J.~Kanniainen, M.~Gabbouj, and A.~Iosifidis, ``Deep
  adaptive input normalization for time series forecasting,'' \emph{{IEEE}
  Trans. Neural Networks Learn. Syst.}, vol.~31, no.~9, pp. 3760--3765, 2020.

\bibitem{9509335}
F.~Ilhan, O.~Karaahmetoglu, I.~Balaban, and S.~S. Kozat, ``Markovian rnn: An
  adaptive time series prediction network with hmm-based switching for
  nonstationary environments,'' \emph{{IEEE} Trans. Neural Networks Learn.
  Syst.}, pp. 1--14, 2021.

\bibitem{DBLP:journals/corr/abs-2302-14829}
W.~Fan, P.~Wang, D.~Wang, D.~Wang, Y.~Zhou, and Y.~Fu, ``Dish-ts: {A} general
  paradigm for alleviating distribution shift in time series forecasting,'' in
  \emph{Proc. {AAAI}}, 2023, pp. 7522--7529.

\bibitem{1611835114}
J.~Kirkpatrick, R.~Pascanu, N.~Rabinowitz, and et~al., ``Overcoming
  catastrophic forgetting in neural networks,'' \emph{Proceedings of the
  National Academy of Sciences}, vol. 114, no.~13, pp. 3521--3526, 2017.

\bibitem{DBLP:conf/aaai/YueWDYHTX22}
Z.~Yue, Y.~Wang, J.~Duan, T.~Yang, C.~Huang, Y.~Tong, and B.~Xu, ``Ts2vec:
  Towards universal representation of time series,'' in \emph{Proc. {AAAI}},
  2022, pp. 8980--8987.

\bibitem{DBLP:conf/cikm/Du0FPQXW21}
Y.~Du, J.~Wang, W.~Feng, S.~J. Pan, T.~Qin, R.~Xu, and C.~Wang, ``Adarnn:
  Adaptive learning and forecasting of time series,'' in \emph{Proc. {CIKM}},
  2021, pp. 402--411.

\bibitem{DBLP:conf/cikm/YouZDFH21}
X.~You, M.~Zhang, D.~Ding, F.~Feng, and Y.~Huang, ``Learning to learn the
  future: Modeling concept drifts in time series prediction,'' in \emph{Proc.
  {CIKM}}, 2021, pp. 2434--2443.

\bibitem{DBLP:journals/tnn/SongLLZ23}
Y.~Song, J.~Lu, H.~Lu, and G.~Zhang, ``Learning data streams with changing
  distributions and temporal dependency,'' \emph{{IEEE} Trans. Neural Networks
  Learn. Syst.}, vol.~34, no.~8, pp. 3952--3965, 2023.

\bibitem{10.1145/3580305.3599315}
L.~Zhao, S.~Kong, and Y.~Shen, ``Doubleadapt: A meta-learning approach to
  incremental learning for stock trend forecasting,'' in \emph{Proc. {KDD}},
  2023, pp. 3492--3503.

\bibitem{DBLP:conf/acl/ZhaoZE17}
T.~Zhao, R.~Zhao, and M.~Esk{\'{e}}nazi, ``Learning discourse-level diversity
  for neural dialog models using conditional variational autoencoders,'' in
  \emph{Proc. {ACL} {(1)}}, 2017, pp. 654--664.

\bibitem{DBLP:conf/iclr/GuCHK19}
X.~Gu, K.~Cho, J.~Ha, and S.~Kim, ``Dialogwae: Multimodal response generation
  with conditional wasserstein auto-encoder,'' in \emph{Proc. {ICLR} (Poster)},
  2019.

\bibitem{DBLP:conf/ijcai/YangYDSZZ21}
H.~Yang, X.~Yao, Y.~Duan, J.~Shen, J.~Zhong, and K.~Zhang, ``Progressive
  open-domain response generation with multiple controllable attributes,'' in
  \emph{Proc. {IJCAI}}, 2021, pp. 3279--3285.

\bibitem{DBLP:journals/corr/abs-1910-11800}
R.~Jankovic, I.~Mihajlovic, and A.~Amelio, ``Time series vector autoregression
  prediction of the ecological footprint based on energy parameters,''
  \emph{CoRR}, vol. abs/1910.11800, 2019.

\bibitem{DBLP:conf/nips/SalinasBCMG19}
D.~Salinas, M.~Bohlke{-}Schneider, L.~Callot, R.~Medico, and J.~Gasthaus,
  ``High-dimensional multivariate forecasting with low-rank gaussian copula
  processes,'' in \emph{Proc. NeurIPS}, 2019, pp. 6824--6834.

\bibitem{WangRXC21}
Q.~Wang, S.~Ren, Y.~Xia, and L.~Cao, ``{BiCMTS}: Bidirectional coupled
  multivariate learning of irregular time series with missing values,'' in
  \emph{Proc. {CIKM}}, 2021, pp. 3493--3497.

\bibitem{DBLP:journals/tnn/MohajerinW19}
N.~Mohajerin and S.~L. Waslander, ``Multistep prediction of dynamic systems
  with recurrent neural networks,'' \emph{{IEEE} Trans. Neural Networks Learn.
  Syst.}, vol.~30, no.~11, pp. 3370--3383, 2019.

\bibitem{DBLP:conf/cikm/HuangWWT19}
S.~Huang, D.~Wang, X.~Wu, and A.~Tang, ``Dsanet: Dual self-attention network
  for multivariate time series forecasting,'' in \emph{Proc. {CIKM}}, 2019, pp.
  2129--2132.

\bibitem{DBLP:conf/kdd/DengCJST21}
J.~Deng, X.~Chen, R.~Jiang, X.~Song, and I.~W. Tsang, ``St-norm: Spatial and
  temporal normalization for multi-variate time series forecasting,'' in
  \emph{Proc. {KDD}}, 2021, pp. 269--278.

\bibitem{DBLP:conf/nips/YiZFWWHALCN23}
K.~Yi, Q.~Zhang, W.~Fan, S.~Wang, P.~Wang, H.~He, N.~An, D.~Lian, L.~Cao, and
  Z.~Niu, ``Frequency-domain mlps are more effective learners in time series
  forecasting,'' in \emph{Proc. {NeurIPS}}, 2023.

\bibitem{DBLP:conf/ijcai/WuPLJZ19}
Z.~Wu, S.~Pan, G.~Long, J.~Jiang, and C.~Zhang, ``Graph wavenet for deep
  spatial-temporal graph modeling,'' in \emph{Proc. {IJCAI}}, 2019, pp.
  1907--1913.

\bibitem{DBLP:conf/ijcai/YuYZ18}
B.~Yu, H.~Yin, and Z.~Zhu, ``Spatio-temporal graph convolutional networks: {A}
  deep learning framework for traffic forecasting,'' in \emph{Proc. {IJCAI}},
  2018, pp. 3634--3640.

\bibitem{DBLP:conf/kdd/WuPL0CZ20}
Z.~Wu, S.~Pan, G.~Long, J.~Jiang, X.~Chang, and C.~Zhang, ``Connecting the
  dots: Multivariate time series forecasting with graph neural networks,'' in
  \emph{Proc. {KDD}}, 2020, pp. 753--763.

\bibitem{DBLP:conf/nips/YiZFHHWACN23}
K.~Yi, Q.~Zhang, W.~Fan, H.~He, L.~Hu, P.~Wang, N.~An, L.~Cao, and Z.~Niu,
  ``Fouriergnn: Rethinking multivariate time series forecasting from a pure
  graph perspective,'' in \emph{Proc. {NeurIPS}}, 2023.

\bibitem{DBLP:conf/aaai/HeZBYN22}
H.~He, Q.~Zhang, S.~Bai, K.~Yi, and Z.~Niu, ``{CATN:} cross attentive
  tree-aware network for multivariate time series forecasting,'' in \emph{Proc.
  {AAAI}}, 2022, pp. 4030--4038.

\bibitem{DBLP:conf/icpr/LyuL0GSZ20}
Y.~Lyu, M.~Li, X.~Huang, U.~Guler, P.~Schaumont, and Z.~Zhang, ``Treernn:
  Topology-preserving deep graph embedding and learning,'' in \emph{Proc.
  {ICPR}}, 2020, pp. 7493--7499.

\bibitem{DBLP:conf/aaai/FangRSSL00023}
Y.~Fang, K.~Ren, C.~Shan, Y.~Shen, Y.~Li, W.~Zhang, Y.~Yu, and D.~Li,
  ``Learning decomposed spatial relations for multi-variate time-series
  modeling,'' in \emph{Proc. {AAAI}}, 2023, pp. 7530--7538.

\bibitem{Yangdsaa23}
Y.~Yang, Z.~Zhao, and L.~Cao, ``Deep spectral copula mechanisms modeling
  coupled and volatile multivariate time series,'' in \emph{Proc. {DSAA}},
  2023.

\bibitem{10.1145/3653447}
K.~Yi, Q.~Zhang, H.~He, K.~Shi, L.~Hu, N.~An, and Z.~Niu, ``Deep coupling
  network for multivariate time series forecasting,'' \emph{ACM Trans. Inf.
  Syst.}, pp. 1--29, Mar. 2024.

\bibitem{DBLP:conf/aaai/NguyenQ21}
N.~Nguyen and B.~Quanz, ``Temporal latent auto-encoder: {A} method for
  probabilistic multivariate time series forecasting,'' in \emph{Proc. {AAAI}},
  2021, pp. 9117--9125.

\bibitem{DBLP:conf/iclr/FortuinHLSR19}
V.~Fortuin, M.~H{\"{u}}ser, F.~Locatello, H.~Strathmann, and G.~R{\"{a}}tsch,
  ``{SOM-VAE:} interpretable discrete representation learning on time series,''
  in \emph{Proc. {ICLR} (Poster)}, 2019.

\bibitem{DBLP:journals/corr/abs-2111-08095}
A.~Desai, C.~Freeman, Z.~Wang, and I.~Beaver, ``Timevae: {A} variational
  auto-encoder for multivariate time series generation,'' \emph{CoRR}, vol.
  abs/2111.08095, 2021.

\bibitem{DBLP:conf/icml/KimKS21}
J.~Kim, J.~Kong, and J.~Son, ``Conditional variational autoencoder with
  adversarial learning for end-to-end text-to-speech,'' in \emph{Proc. {ICML}},
  vol. 139, 2021, pp. 5530--5540.

\bibitem{DBLP:conf/sigir/AskariSS21}
B.~Askari, J.~Szlichta, and A.~Salehi{-}Abari, ``Variational autoencoders for
  top-k recommendation with implicit feedback,'' in \emph{Proc. {SIGIR}}, 2021,
  pp. 2061--2065.

\bibitem{DBLP:conf/iclr/PrakashKJ21}
M.~Prakash, A.~Krull, and F.~Jug, ``Fully unsupervised diversity denoising with
  convolutional variational autoencoders,'' in \emph{Proc. {ICLR}}, 2021.

\bibitem{DBLP:journals/tnn/HeCLWYCLZ23}
Z.~He, P.~Chen, X.~Li, Y.~Wang, G.~Yu, C.~Chen, X.~Li, and Z.~Zheng, ``A
  spatiotemporal deep learning approach for unsupervised anomaly detection in
  cloud systems,'' \emph{{IEEE} Trans. Neural Networks Learn. Syst.}, vol.~34,
  no.~4, pp. 1705--1719, 2023.

\bibitem{DBLP:conf/nips/MaCX18}
T.~Ma, J.~Chen, and C.~Xiao, ``Constrained generation of semantically valid
  graphs via regularizing variational autoencoders,'' in \emph{Proc. NeurIPS},
  2018, pp. 7113--7124.

\bibitem{DBLP:journals/tog/AbdalZMW21}
R.~Abdal, P.~Zhu, N.~J. Mitra, and P.~Wonka, ``Styleflow: Attribute-conditioned
  exploration of stylegan-generated images using conditional continuous
  normalizing flows,'' \emph{{ACM} Trans. Graph.}, vol.~40, no.~3, pp.
  21:1--21:21, 2021.

\bibitem{DBLP:conf/iclr/KumarBEFLDK20}
M.~Kumar, M.~Babaeizadeh, D.~Erhan, C.~Finn, S.~Levine, L.~Dinh, and D.~Kingma,
  ``Videoflow: {A} conditional flow-based model for stochastic video
  generation,'' in \emph{Proc. {ICLR}}, 2020.

\bibitem{DBLP:conf/icml/LuoYJ21}
Y.~Luo, K.~Yan, and S.~Ji, ``Graphdf: {A} discrete flow model for molecular
  graph generation,'' in \emph{Proc. {ICML}}, vol. 139, 2021, pp. 7192--7203.

\bibitem{DBLP:conf/aaai/ZhangYXL21}
Z.~Zhang, C.~Yu, S.~Xu, and H.~Li, ``Learning flexibly distributional
  representation for low-quality 3d face recognition,'' in \emph{Proc. {AAAI}},
  2021, pp. 3465--3473.

\bibitem{DBLP:conf/iclr/RasulSSBV21}
K.~Rasul, A.~Sheikh, I.~Schuster, U.~M. Bergmann, and R.~Vollgraf,
  ``Multivariate probabilistic time series forecasting via conditioned
  normalizing flows,'' in \emph{Proc. {ICLR}}, 2021.

\bibitem{DBLP:conf/aaai/SawhneyAWDS21}
R.~Sawhney, S.~Agarwal, A.~Wadhwa, T.~Derr, and R.~R. Shah, ``Stock selection
  via spatiotemporal hypergraph attention network: {A} learning to rank
  approach,'' in \emph{Proc. {AAAI}}, 2021, pp. 497--504.

\bibitem{DBLP:conf/emnlp/LaiTBK19}
T.~M. Lai, Q.~H. Tran, T.~Bui, and D.~Kihara, ``A gated self-attention memory
  network for answer selection,'' in \emph{Proc. {EMNLP/IJCNLP} {(1)}}, 2019,
  pp. 5952--5958.

\bibitem{DBLP:conf/emnlp/DingL21}
H.~Ding and X.~Luo, ``Attentionrank: Unsupervised keyphrase extraction using
  self and cross attentions,'' in \emph{Proc. {EMNLP} {(1)}}, 2021, pp.
  1919--1928.

\bibitem{10285474}
H.~He, Q.~Zhang, S.~Wang, K.~Yi, Z.~Niu, and L.~Cao, ``Learning informative
  representation for fairness-aware multivariate time-series forecasting: A
  group-based perspective,'' \emph{IEEE Trans. Knowl. Data Eng.}, pp. 1--13,
  2023.

\bibitem{DBLP:conf/aaai/WuNCZSCLZCD21}
Y.~Wu, J.~Ni, W.~Cheng, B.~Zong, D.~Song, Z.~Chen, Y.~Liu, X.~Zhang, H.~Chen,
  and S.~B. Davidson, ``Dynamic gaussian mixture based deep generative model
  for robust forecasting on sparse multivariate time series,'' in \emph{Proc.
  {AAAI}}, 2021, pp. 651--659.

\bibitem{DBLP:conf/iclr/0010ZYCF0L22}
W.~Fan, S.~Zheng, X.~Yi, W.~Cao, Y.~Fu, J.~Bian, and T.~Liu, ``{DEPTS:} deep
  expansion learning for periodic time series forecasting,'' in \emph{Proc.
  {ICLR}}, 2022.

\end{thebibliography}

\vspace{-1.05cm}
\begin{IEEEbiography}[{\includegraphics[width=1in,height=1.25in,clip,keepaspectratio]{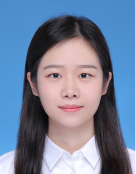}}]{Hui He} received the ME degree from University of Shanghai for Science and Technology, Shanghai, China in 2020. She is currently working toward the PhD degree with the School of Medical Technology, Beijing Institute of Technology, Beijing, China. Her current research interests focus on multivariate time series analysis and knowledge services.
\end{IEEEbiography}

\begin{IEEEbiography}[{\includegraphics[width=1in,height=1.25in,clip,keepaspectratio]{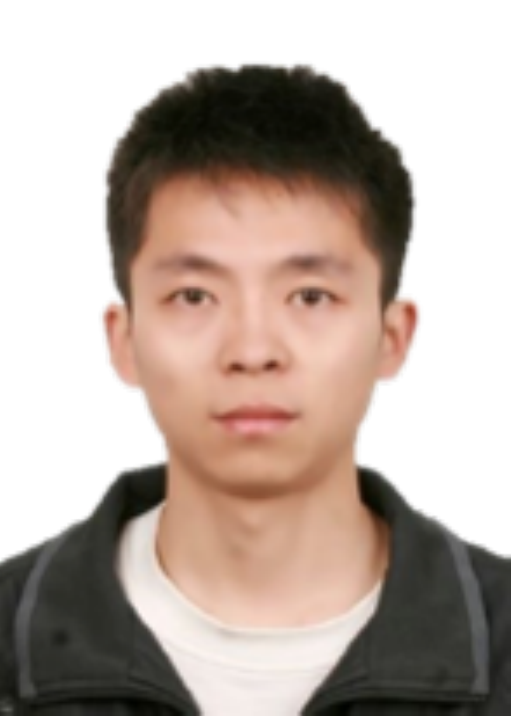}}]{Qi Zhang}
received the PhD degree from the Beijing Institute of Technology, Beijing, China and the University of Technology Sydney, Sydney, NSW, Australia, in 2020, under the dual PhD program. 
He is currently a Research Fellow with Tongji University, Shanghai, China. He has authored high-quality papers in premier conferences and journals, including AAAI Conference on Artificial Intelligence (AAAI), International Joint Conferences on Artificial Intelligence (IJCAI), 
International World Wide Web Conference (TheWebConf), IEEE Transactions on Knowledge and Data Engineering (TKDE), IEEE Transactions on Neural Networks and Learning Systems (TNNLS), and ACM Transactions on Information Systems (TOIS). His primary research interests include collaborative filtering, recommendation, learning to hash, and MTS analysis.
\end{IEEEbiography}

\begin{IEEEbiography}[{\includegraphics[width=1in,height=1.25in,clip,keepaspectratio]{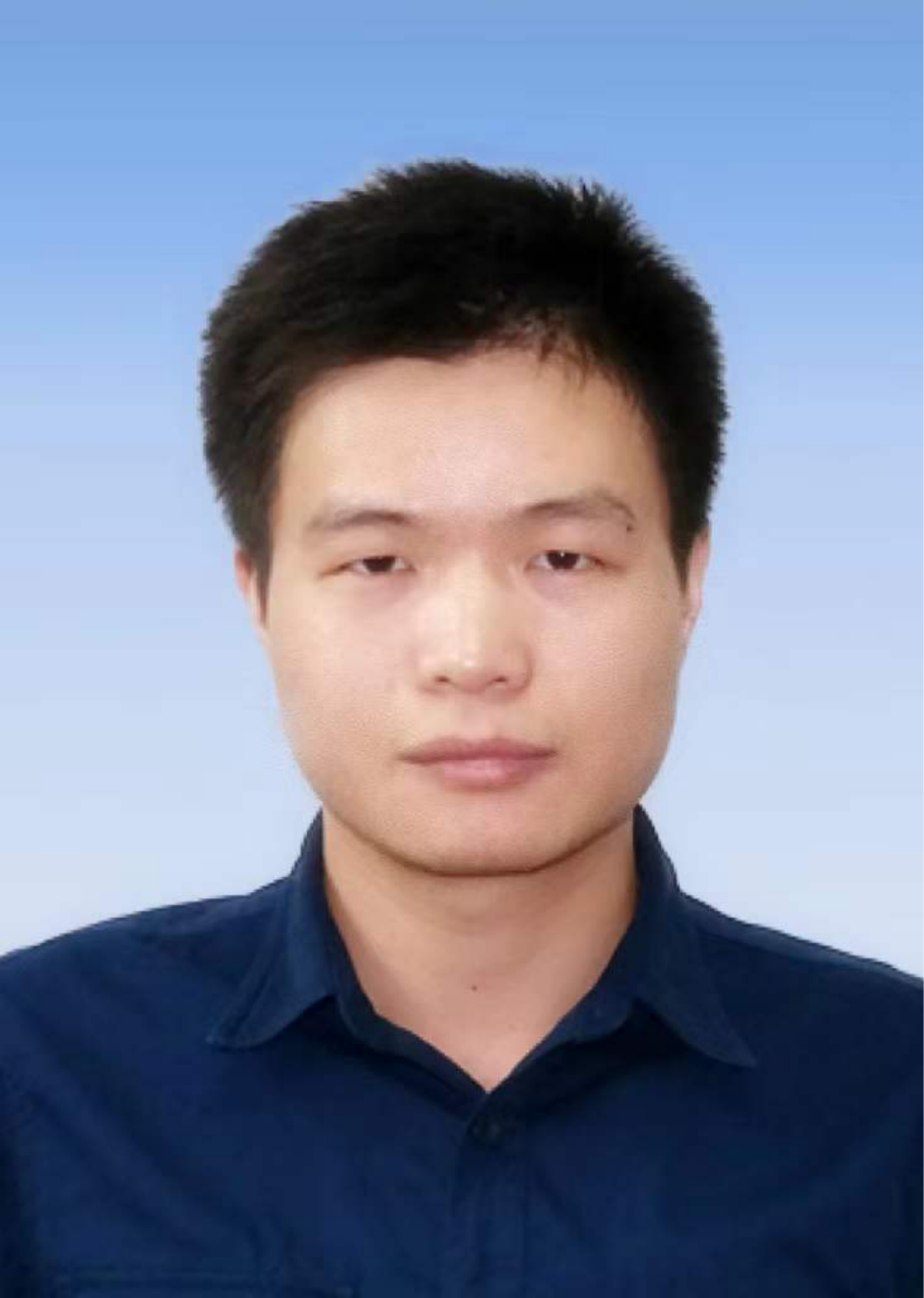}}]{Kun Yi}
is currently working toward PhD degree with the Beijing Institute of Technology, China. His current research interests include multivariate time series forecasting, data science, and knowledge discovery.
\end{IEEEbiography}

\begin{IEEEbiography}[{\includegraphics[width=1in,height=1.25in,clip,keepaspectratio]{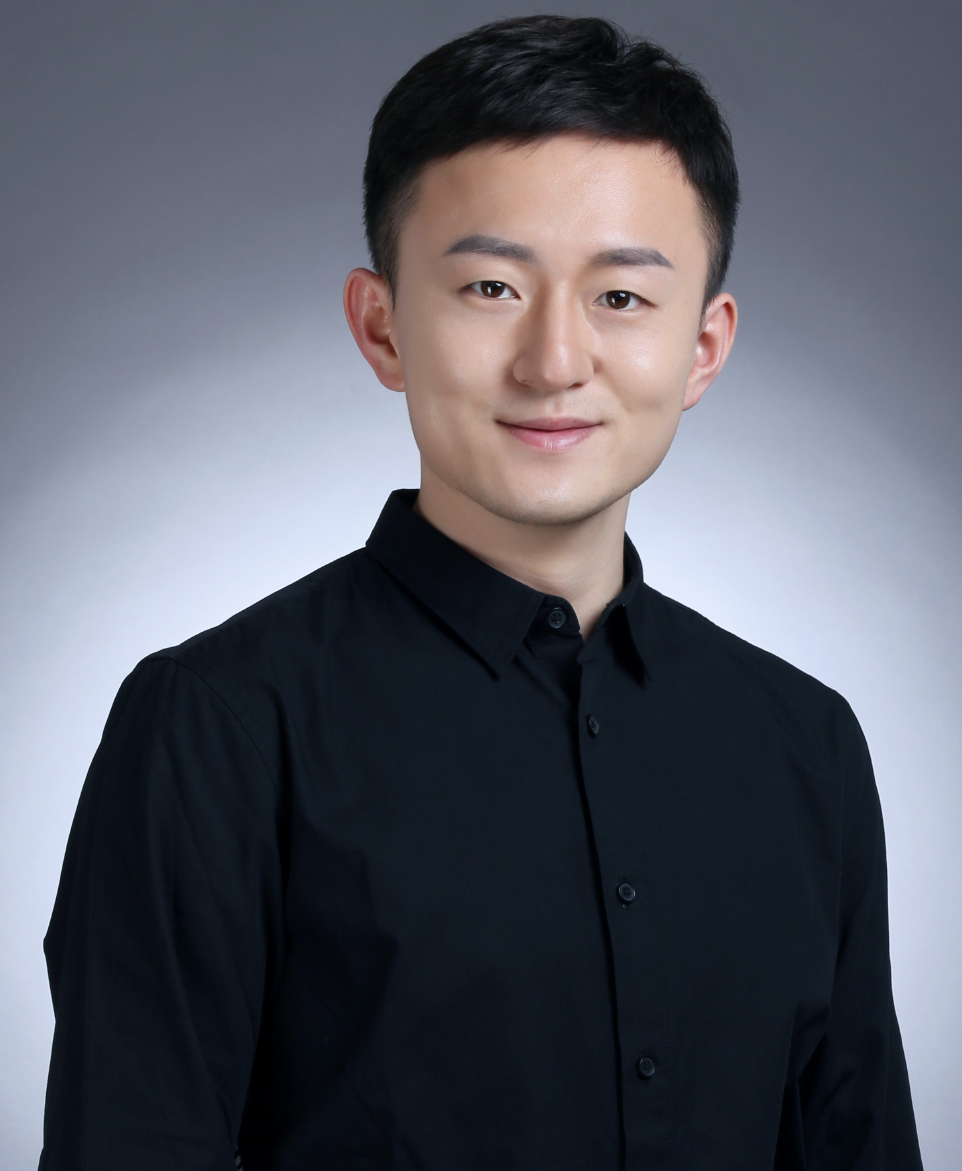}}]{Kaize Shi} 
(S'20-M'21) is with the Data Science and Machine Intelligence Lab, University of Technology Sydney. He has PhD degrees in computer science and computer systems, which are from the Beijing Institute of Technology, China, and the University of Technology Sydney, Australia. His research interests include natural language generation, social computing, cyber-physical-social systems, meteorological knowledge services, intelligent transportation, and artificial intelligence technology. 
He is the associate editor of IEEE Transactions on Computational Social Systems (IEEE TCSS) and academic editor of PeerJ Computer Science and Wireless Communications and Mobile Computing. He also served as a guest editor for the International Journal of Distributed Sensor Networks, and as a reviewer for the IEEE Transactions on Intelligent Transportation Systems, IEEE Internet of Things Journal, etc. 
\end{IEEEbiography}


\begin{IEEEbiography}[{\includegraphics[width=1in,height=1.25in,clip,keepaspectratio]{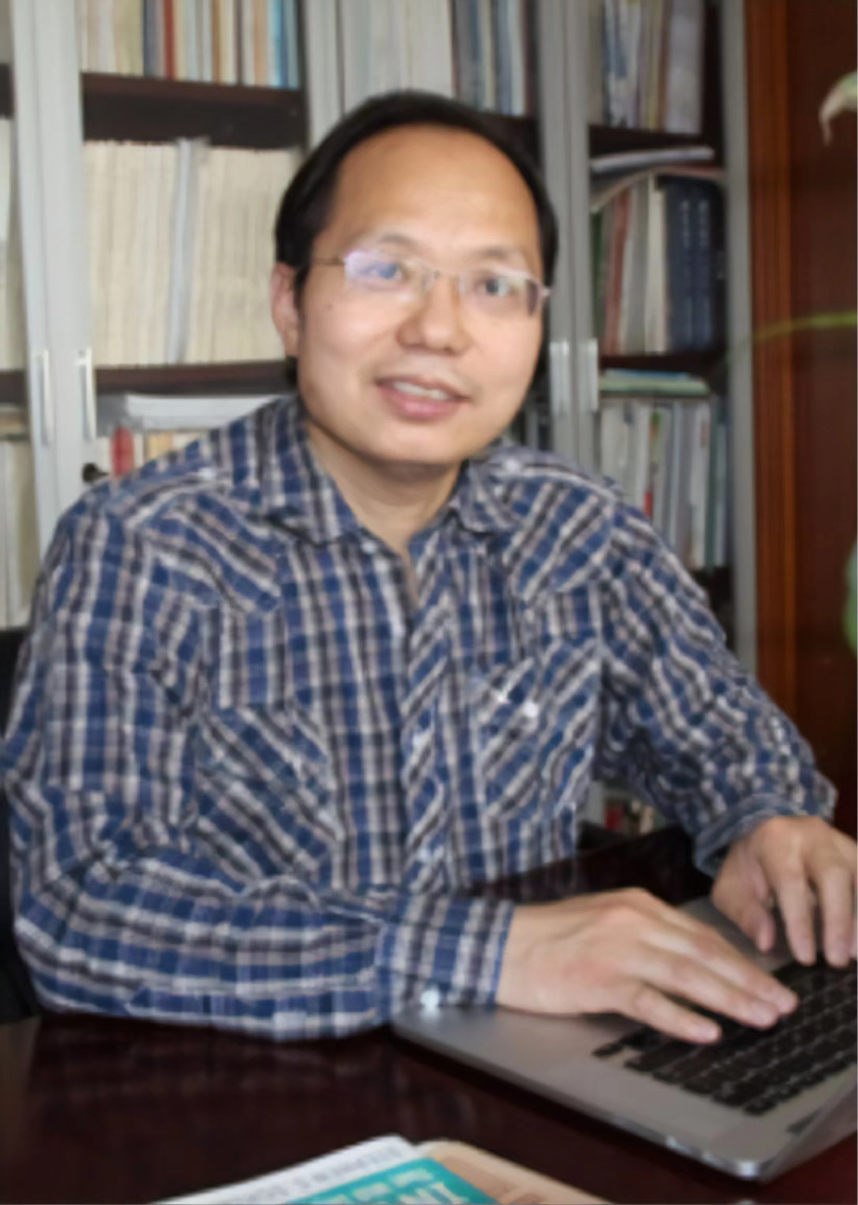}}]{Zhendong Niu}
received the PhD degree in computer science from the Beijing Institute of Technology in 1995. 
He was a Post-Doctoral Researcher with the University of Pittsburgh, Pittsburgh, PA, USA, from 1996 to 1998, a Researcher/Adjunct Faculty Member with Carnegie Mellon University, Pittsburgh, from 1999 to 2004, and a Joint Research Professor with the School of Computing and Information, University of Pittsburgh, in 2006. 
He is a Professor with the School of Computer Science and Technology, Beijing Institute of Technology, Beijing. 
His research interests include informational retrieval, software architecture, digital libraries, and web-based learning. 
He received the International Business Machines Corporation (IBM) Faculty Innovation Award in 2005 and the New Century Excellent Talents in University of the Ministry of Education of China in 2006.  
\end{IEEEbiography}

\begin{IEEEbiography}[{\includegraphics[width=1in,height=1.25in,clip,keepaspectratio]{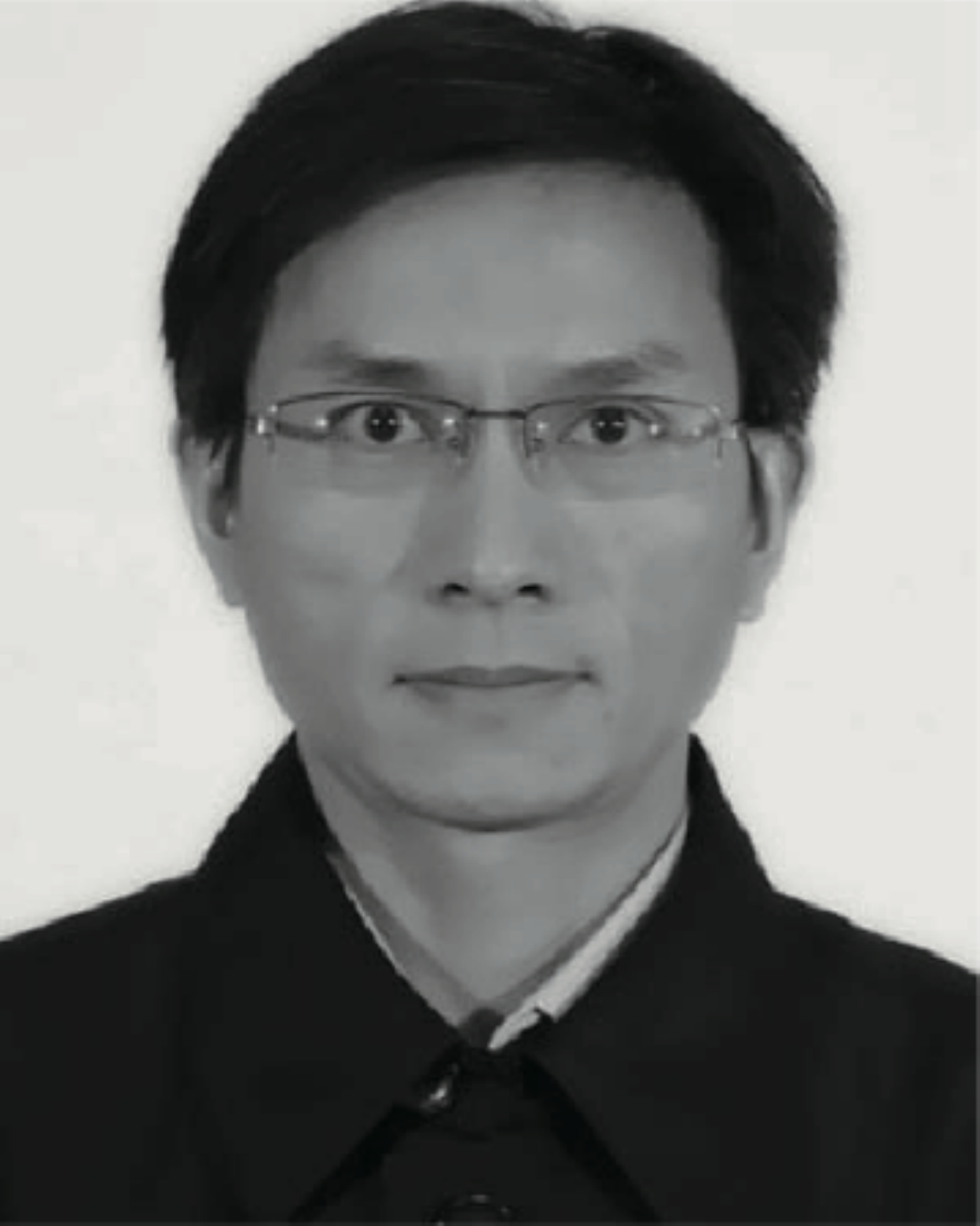}}]{Longbing Cao} (@SM in 2006) received PhD degree in pattern recognition and intelligent systems from the Chinese Academy of Science, China, and another PhD degree in computing sciences from the University of Technology Sydney, Australia. 
He is a Distinguished Chair Professor with Macquarie University, an ARC Future Fellow (professorial level), and the EiCs of IEEE Intelligent Systems and Journal of Data Science and Analytics. His research interests include artificial intelligence, data science, machine learning, behavior informatics, and their enterprise applications.
\end{IEEEbiography}

\vfill

\end{document}